\newif\ifarxiv
\arxivtrue 
\ifarxiv
\documentclass[utf8]{article} 
\usepackage{arxiv}
\usepackage[authoryear,round,longnamesfirst]{natbib}
\usepackage[dvipsnames]{xcolor}
\usepackage{fancyhdr}
\fancyheadoffset{0pt}
\else
\documentclass[utf8]{frontiersSCNS} 
\fi

\usepackage{url,hyperref,lineno,microtype,subcaption}
\usepackage[onehalfspacing]{setspace}
\usepackage{comment}
\usepackage{todonotes}
\usepackage{amsmath,amssymb}
\usepackage{commath}  
\usepackage[group-separator={,}, group-four-digits]{siunitx}
\usepackage[utf8]{inputenc}
\usepackage[english]{babel}
\usepackage{url}
\usepackage[T1]{fontenc}\usepackage[ps]{skak}\usepackage{chessboard}
\usepackage{bigdelim} 
\usepackage{amssymb}
\usepackage{pifont}
\usepackage[justification=centering]{caption}
\usepackage{subcaption}
\usepackage{float}
\usepackage{pdflscape}
\usepackage{units} 
\usepackage{cleveref} 
\usepackage{placeins} 
\usepackage{tikz}
\usepackage{pgfplots}

\usepackage{pgfplots}
\pgfplotsset{compat=newest}
\usepgfplotslibrary{groupplots}
\usepgfplotslibrary{dateplot}
\usepackage{booktabs} 
\usepackage{csquotes}
\usepackage{color}
\usepackage{xcolor}

\def\keyFont{\fontsize{8}{11}\helveticabold }
\def\firstAuthorLast{Czech {et~al.}} 
\def\Authors{
\ifarxiv
Johannes Czech\,$^{1,}$\textsuperscript{*},
\else
Johannes Czech\,$^{1,*}$,
\fi
Moritz Willig\,$^{1}$, Alena Beyer\,$^{1}$,
\ifarxiv
\else
\\
\fi
Kristian Kersting\,$^{1,2}$, Johannes Fürnkranz\,$^{1}$}

\def\RunningTitle{Deep Learning for Crazyhouse Chess}
\def\TitleStr{Learning to play the Chess Variant Crazyhouse above World Champion Level with Deep Neural Networks and Human Data}

\def\Address{$^{1}$\,Department of Computer Science, TU Darmstadt, Germany\\ $^{2}$\,Centre for Cognitive Science, TU Darmstadt, Germany} 

\def\corrEmail{johannes.czech@stud.tu-darmstadt.de}

\newcommand{\alg}[1]{\textsl{#1}}
\newcommand{\CrazyAra}{\alg{CrazyAra} }
\newcommand{\model}[1]{\textsl{#1}}
\ifarxiv

\newcommand{\ccsmall}[1]{\small{#1}}

\newcommand{\ctexttt}[1]{\texttt{#1}}
\newcommand{\cstexttt}[1]{\texttt{#1}}
\def \bracketscale {2.0}
\else

\newcommand{\ccsmall}[1]{\footnotesize{#1}}

\newcommand{\ctexttt}[1]{\small{\texttt{#1}}}
\newcommand{\cstexttt}[1]{\footnotesize{\texttt{#1}}}
\def \bracketscale {1.7}
\fi
\newcommand{\cfootnote}[1]{\,\footnote{#1}}
\newcommand{\cnt}[1]{\hspace{-0.38cm}\vspace{-0.8cm}\fontsize{1.0ex}{1.0ex}\selectfont{
\setlength{\fboxsep}{1.7pt}
\colorbox{gray}{\white{#1}}}}
\def \chessboardscale {0.88} 
\def\crule{\noindent\makebox[\linewidth]{\rule{\textwidth}{0.4pt}}\newline}

\makeatletter
\newcommand\footnoteref[1]{\protected@xdef\@thefnmark{\ref{#1}}\@footnotemark}
\makeatother
\begin{document}
%


\ifarxiv
\title{\TitleStr}
\author{\Authors\vspace{0.2cm}\\
        \Address\vspace{0.2cm}\\
        Correspondence:\textsuperscript{*} \corrEmail
}
\fi

\maketitle


\begin{abstract}

Deep neural networks have been successfully applied in learning the board games Go, chess and shogi without prior knowledge by making use of reinforcement learning.
Although starting from zero knowledge has been shown to yield impressive results, it is associated with high computationally costs especially for complex games.
With this paper, we present
\alg{CrazyAra} which is a neural network based engine solely trained in supervised manner for the chess variant crazyhouse.
Crazyhouse is a game with a higher branching factor than chess and there is only limited data of  lower quality available compared to
\alg{AlphaGo}.
Therefore, we focus on improving efficiency in multiple aspects
while relying on low computational resources.
These improvements include modifications in the neural network design and training configuration, the introduction of a data normalization step and a more sample efficient  Monte-Carlo tree search which has a lower chance to blunder.
After training on \num{569537} human games for \num{1.5} days
we achieve a move
prediction accuracy of $\num{60.4}\,\%$.
During development,
versions of \alg{CrazyAra} played professional human players. Most notably,
\alg{CrazyAra} achieved a four to one win over
2017 crazyhouse world champion
Justin Tan (aka \textit{LM Jann Lee}) who is more than \num{400} Elo higher rated compared to the average player in our training set.
Furthermore, we test the playing strength of \alg{CrazyAra} on CPU
against all participants of the second Crazyhouse Computer Championships 2017, winning against twelve of the thirteen participants. 
Finally, for 
\alg{CrazyAraFish} we continue training our model on 
generated engine games. 
In ten long-time control matches playing \alg{\mbox{Stockfish 10}}, \alg{CrazyAraFish} wins three games and draws one out of ten matches.
%
\ifarxiv
\providecommand{\keywords}[1]
{
  \small	
  \textbf{\textit{Keywords---}} #1
}
\fi
\end{abstract}
\ifarxiv
\newpage
\fi
\section{Introduction}
The project \alg{AlphaZero} \citep{silver2017mastering} with its predecessors \alg{AlphaGoZero} \citep{silverd2016masteringthe} and \alg{AlphaGo} \citep{silver2016mastering} marks a milestone in the field of artificial intelligence, demonstrating that the board games Go, chess and shogi can be learned from zero human knowledge.
%
In this article, we extend the this family of games. We present 
the neural network based engine \alg{CrazyAra}
which learned to play the chess variant crazyhouse soley in a supervised fashion.
Crazyhouse, also known as drop chess, is the single-player version of the game bughouse
and introduces the ability to re-introduce pieces that have been captured. The captured piece switches its color, and is henceforth held in the capturing so-called pocket of the respective play.
Crazyhouse incorporates all classical chess rules including castling, en passant capture and draw by three-fold repetition. In addition, instead of playing a classical move, the player has the option to drop any pocket piece
onto an empty square of the board, with the exception that pawns cannot be dropped on the first or eighth rank. The element of dropping captured pieces is similar to the game of shogi, with the difference that in crazyhouse pawns can also be dropped to deliver immediate checkmate.
The fact that pieces never fully leave the game makes crazyhouse a highly tactical game with a considerably larger branching factor than conventional chess.
The chance of drawing and the average game length are significantly reduced because games almost always end in checkmate, and the element of the chess endgame is missing.
Moreover, the ability to drop pieces to set your opponent in check often enables long forced checkmating sequences.
Furthermore, crazyhouse is commonly played in short time controls, and is increasing in popularity particularly for the online community. 

We approach the problem by training a deep neural network from human game data, in a similar fashion as \alg{AlphaGo} \citep{silver2016mastering}. Unfortunately, \alg{AlphaGo} is difficult, if not impossible, to directly apply to crazyhouse. First of all, there is only a significantly smaller data set of lower quality available compared to Go or chess. Second, because of the higher move complexity and the more dynamic nature of the game, several challenges had to be overcome when adapting the
neural network architecture
and applying Monte-Carlo Tree Search (MCTS).
Specifically, our contributions are as follows:

\begin{itemize}
\item First, we introduce a more compact input board presentation by making the state fully Markovian and removing the history component.

\item Second, we highlight the importance of input preprocessing in form of rescaling or normalization for significant better performance.

\item Third, we present a new more powerful and more efficient neural network architecture based on advancements in computer vision such as grouped depthwise convolutions, pre-activation resnets and squeeze-excitation layers.

\item Fourth, we investigate several techniques to make the Monte Carlo tree search (MCTS) more sample efficient. This includes the usage of Q-Values for move selection, a transposition table which allows sharing evaluations across multiple nodes, and ways to decrease the chances of missing critical moves.

\item Finally, we evaluate the strength of a neural network
in combination with MCTS with expert human players as well as the most common crazyhouse chess engines.

\end{itemize}

We proceed as follows. We start off, in Section~\ref{sec:related}, by briefly reviewing prior work in computer crazyhouse and machine learning in games. 
Section~\ref{sec:overview} then goes through the general scheme on how the neural network is trained and integrated with MCTS to be used as an engine.
Our input representation for encoding the board state is introduced in Section~\ref{sec:input_representation}, and our
output representation in Section~\ref{sec:output_representation}.
Section~\ref{sec:model_arch} goes over the \alg{AlphaZero} network architecture and introduces different convolutional neural network designs, which make use of pre-activation residual blocks \citep{he2016identity}, depthwise separable convolutions \citep{howard2017mobilenets} and Squeeze Excitation Layers (SE; \citealt{hu2018squeeze}).
Next, in Section~\ref{sec:training_data}, we describe the training data in more detail, including statistics of the most frequent players, the win and draw percentages as well as the occurrence of different time controls.
We also summarize how a computer generated data set based on \alg{Stockfish} self play games was created.
Then, the configuration of the supervised training is provided in Section~\ref{sec:supervised_training} and the performance and progress for different network architectures is visualized.
Section~\ref{sec:mcts} outlines the formulas for the MCTS algorithm including its hyperparameter settings.
We also introduce several changes and extensions such as including Q-values for final move selection, a conversion of Q-Values to Centi-Pawn (CP), the usage of a transposition table, a parameterization for calculating the U-Values and a possible integration of domain knowledge to make the search more robust and sample efficient. We continue with a discussion in Section~\ref{sec:discussion} highlighting the benefits and disadvantages of MCTS compared to Alpha-Beta minimax search.
Before concluding, we summarize the match results with human professional players and other crazyhouse engines.

\section{Related Work on Computer Crazyhouse and ML in Board Games}
\label{sec:related}

Crazyhouse is a fairly recent chess variant, which primarily enjoys popularity in on-line chess servers such as
\ifarxiv
\url{lichess.org}.
\fi
Despite its youth, there are already more than a dozen engines available which are able to play this chess variant (cf.~also
Section \ref{sec:strength_eval}).
The first two of these engines are \alg{Sjeng}\cfootnote{\url{https://www.sjeng.org/indexold.html}, accessed 2019-07-30}, written by Gian-Carlo Pascutto released in 1999, and \alg{Sunsetter}\cfootnote{\label{foot:sunsetter}\url{http://sunsetter.sourceforge.net/}, accessed 2019-07-30}, developed by Georg v. Zimmermann and Ben Dean-Kawamura in 2001.
Later the strong open-source chess engine \alg{Stockfish}\cfootnote{\url{https://github.com/ddugovic/Stockfish}, accessed 2019-07-30}
has been adapted to play crazyhouse by Daniel Dugovic, Fabian Fichter and Niklas Fiekas.
\alg{Stockfish} won the first Crazyhouse Computer Championships 2016 and also the second Crazyhouse Computer Championships 2017 \citep{mosca_2nd_2017}.
All these engines have in common that they follow a minimax search regime with alpha-beta pruning, as has been popularized by successful chess engines, most notably \alg{DeepBlue} \citep{lig*Campbell02}.
These engines are often described as having a large number of node evaluations and being strong at tactics in open positions, while sometimes having trouble in generating strategical attacks\,\footnoteref{foot:sunsetter}. 
Due to the higher branching factor in crazyhouse, engines commonly reach a significantly lower search depth compared to classical chess.

Generally, machine learning in computer chess 
\citep{lig*Skiena86,jf:ICCA} and in computer game playing in general has a long history \citep{jf:ML-games}, dating back to Samuel's checkers player \citep{Samuel:59}, which already pioneered many components of modern systems, including linear evaluation functions and reinforcement learning.
The original example that succeeded in tuning a neural network-based evaluation function
to expert strength by playing millions of games against itself is the
backgammon program \alg{TD-Gammon} \citep{TDGammon}. Similar ideas have been carried over to other board games  
\cite[e.g., the chess program \alg{KnightCap};][]{lig*Baxter00}, but the results were not as striking. 

Monte-Carlo Tree Search (MCTS) brought a substantial break-through in the game of Go \citep{gellyMCTSGo}, featuring the idea that instead of using an exhaustive search to a limited depth, as was common in chess-like games, samples of full-depth games can be used to dynamically approximate the game outcome. 
While MCTS works well for Go, in games like chess, where often narrow tactical variations have to be found, MCTS is prone to fall into shallow traps \citep{MCTSTraps}. For such games, it has thus been considered to be considerably weaker than minimax-based approaches, so that hybrid algorithms have been investigated \citep{MCTS-Minimax-Hybrids}.

Recently, \alg{AlphaGo} has brought yet another quantum leap in performance by combining MCTS with deep neural networks, which were trained from human games and improved via self play \citep{silver2016mastering}. \alg{AlphaZero} improved the architecture of \alg{AlphaGo} to a single neural network which can be trained without prior game knowledge. \citet{silver2017mastering} demonstrated the generality of this approach by successfully applying it not only to Go, but also to games like chess and shogi, which have previously been dominated by minimax-based algorithms.
While \alg{AlphaZero} relied
on the computational power of large Google computation servers,
for the games of Go\cfootnote{\url{https://github.com/leela-zero/leela-zero}, accessed 2019-06-10}
and chess\cfootnote{\url{https://github.com/LeelaChessZero/lc0}, accessed 2019-06-10}
the \alg{Leela} project was started with the goal to replicate 
\textit{AlphaZero} 
in a collaborative effort using distributed computing from the crowd.
Several other engines\cfootnote{See e.g.~\url{https://github.com/manyoso/allie}, \url{https://github.com/Cscuile/BetaOne}, accessed 2019-06-10}
built up on \alg{Leela} or are partly based on the source code of the \alg{Leela} project.
Our work on crazyhouse started as an independent project.
Generally, only little work exists on 
machine learning for crazyhouse chess. One exception is the work of 
\cite{droste2008learning}, who used reinforcement learning to learn piece values and piece-square values for three chess variants including crazyhouse.
In parallel to our work, 
\cite{gordon_deep_2019}
also started to develop a neural network to learn crazyhouse.





\section{Overview of the \alg{CrazyAra} Engine}
\begin{figure}[]
\centering
\includegraphics[width=\textwidth]{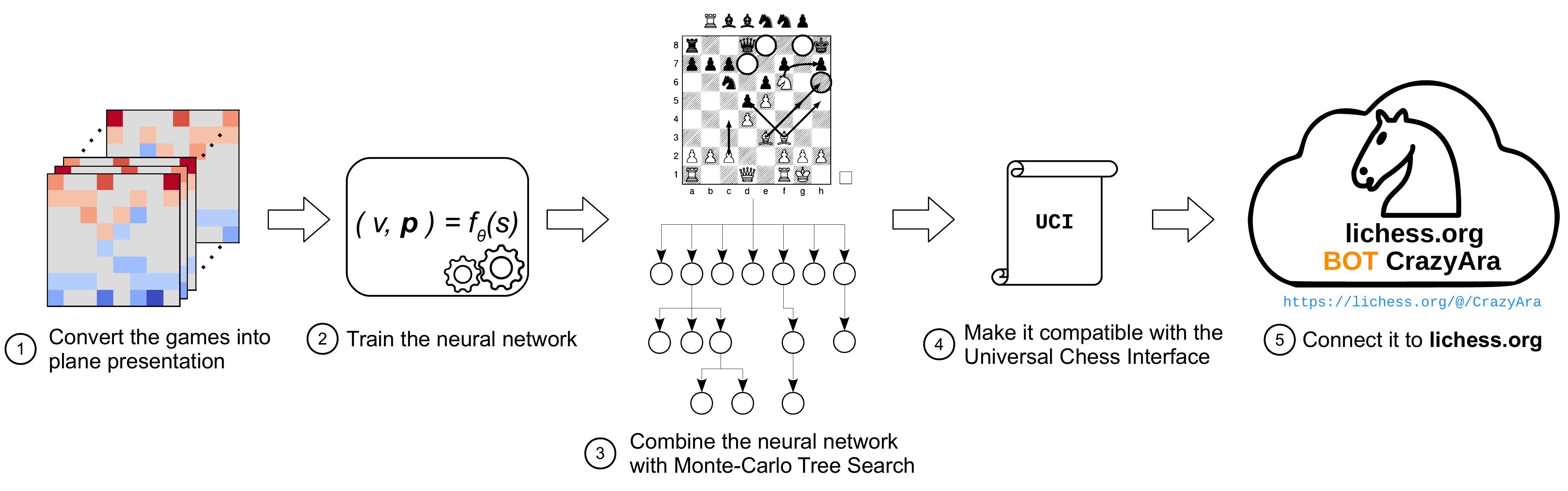}
\caption{Compilation pipeline of the \alg{CrazyAra} engine}
\label{fig:scheme}
\end{figure}
\label{sec:overview}
Our crazyhouse bot \alg{CrazyAra} 
is based on a (deep) neural network that has been first trained on human games and is then optionally refined on computer-generated games. The network is used in a MCTS-based search module, which performs a variable-depth search. In the following, we first briefly sketch all components, see Figure~\ref{fig:scheme}. The details will then be described in subsequent sections. 

\subsection{Deep neural networks for evaluating moves}
\label{sec:DNN}
\alg{CrazyAra} trains 
a (deep) neural network model $f_{\theta}(s)$ to predict the value $v \in [-1, 1]$ of a board state $s$ and its corresponding stochastic policy distribution $\mathbf{p}$ over all available moves (Section~\ref{sec:output_representation}). Since we are learning the network based on preexisting matches, we encode the ground truth label as a one-hot vector. 
Specifically, depending on the final outcome of the game, 
each board state is assigned one of three possible outcomes \{$-1$: lost, $0$: draw, $1$: win\} from the point of view of the player to move. Note that the number of draws is below $\num{1}\,\%$ in human crazhouse games and thus considerably lower than in classical chess.
These assignments are based on the assumption that given a considerable advantage in a particular position for a player, then it is highly unlikely that the respective player will lose the game. This assumption is, however, heavily violated in our data set partly due to the popularity of low time control games, see Section~\ref{sec:training_data}. 

Actually, the neural network is a shared network, which both predicts the value and policy (Section~\ref{sec:model_arch}) for a given board state in a single forward pass.
The loss function is defines as follows
\begin{equation}
    \label{eq:combined_loss}
    l = \alpha (z - v)^2 - \mathbf{\pi}^T \log \mathbf{p} + c \Vert \theta \Vert^2
\end{equation}
where $z$ is the true value, $v$ the predicted value, $\mathbf{\pi}$ the true policy, $\mathbf{p}$ the predicted policy and $c$ the $L_2$ regularization constant for the network weights $\theta$ respectively.
We set $\alpha$ to be $\num{0.01}$ in order to avoid overfitting to the training values as suggested by \cite{silver2016mastering}.
The weights of the neural network are adjusted to reduce the loss by Stochastic Gradient Descent with Neterov's Momentum (NAG;  \citealt{botev2017nesterov}), see also Section~\ref{sec:supervised_training}.
We keep the the neural network weights fixed after training and do not apply any reinforcement learning using self-play yet.

\subsection{Monte-Carlo Tree Search for improving performance}
The trained (deep) neural network of \alg{CrazyAra} has a move prediction accuracy of about 60\%, i.\,e., covers most of the play-style of the average playing strength in the training set.
Its performance is then improved using a variant of the Upper Confidence Bounds for Trees algorithm \citep{UCT}, which integrates sample-based evaluations into a selective tree search. Like with all Monte-Carlo tree search algorithms \citep{MCTSSurvey}, the key idea is to evaluate moves by drawing samples, where good moves are sampled more frequently, and less promising moves are sampled less frequently, thereby trading off exploration and exploitation. 
The key component is the (deep) neural network $f_\theta$ which guides the search in each step.
For each iteration $t$ at state $s_t$, the following UCT-like formula (Equation \ref{eq:node_selection}) is used for selecting the next action $a_t$ leading to a new state $s_{t+1}$.
\begin{equation}
    \label{eq:node_selection}
    a_t = \text{argmax}_a (\text{Q}(s_t,a) + U(s_t, a)) \quad \text{where} \quad
    U(s,a) = c_{\text{puct}} P(s,a) \frac{\sqrt{\sum_b N(s,b)}}{1 + N(s,a)}
\end{equation}
This move selection continues until a new previously unexplored state $s^*$ or a terminal node $s_T$ is reached.
The state~$s^*$ is then expanded, the corresponding position is evaluated using the learned neural network, and the received value evaluation is propagated through all nodes on the path beginning from the current position $s^*$ back to the root position~$s_0$.
In case of a terminal node $s_T$ the static evaluation according to the game rules being either $-1$, $0$ or $+1$ is used instead.
After enough samples have been taken, the most promising move, i.\,e. the node which has been visited most often, is played on the board (Section~\ref{sec:mcts}).


\subsection{Availability of \alg{CrazyAra}}
\alg{CrazyAra} is
compatible with the Universal Chess Interface (UCI; 
\citealt{kahlen_uci_2004}) and uses the BOT API of lichess.org\cfootnote{\url{https://github.com/careless25/lichess-bot}, accessed 2019-06-05}
to provide a public BOT account\cfootnote{\url{https://lichess.org/@/CrazyAra}, accessed 2019-06-05}, which can be challenged to a match when online.
Furthermore, the executable and full source code, including the data preprocessing, architecture definitions, training scripts\cfootnote{\url{https://github.com/QueensGambit/CrazyAra},  accessed 2019-06-05} and MCTS search\cfootnote{\url{https://github.com/QueensGambit/CrazyAra-Engine}, accessed 2019-07-30} is available under the terms of the GNU General Public License v3.0 (GPL-3.0; \citealt{free_gnu.org_2017}).\\
%



Let us now dive into the details of each component of \alg{CrazyAra}. 
\section{Input Representation of \alg{CrazyAra}}
\label{sec:input_representation}
The input to \alg{CrazyAra} is a stack of $8\times8$ planes where each channel describes one of the feature of the current board state described in Table \ref{tab:input_representation}.
The main extension compared to encodings used for classical chess is the addition of pocket pieces accounting for ten additional channels.
Furthermore, we remove the seven step history of previous board states and only consider the current board position as input. 
This decision has several motivations. Crazyhouse, as well as chess is a full information game. In theory, the history is not needed to find the 
best move.
Dropping the game history also allows better compatibility to use the engine in analysis mode. 
Otherwise one would have to add dummy layers, which can distort the network predictions.
Avoiding the history also 
reduces the amount of model parameters, the storage for saving 
and preprocessing the data set. It also allows one to have fewer channels early in the network, as shown in Figure \ref{fig:act_maps}, as well as having higher batch-sizes during training an inference.
Furthermore, it is more consistent in combination with a transposition table, see Section \ref{sec:trans_table}, 
reducing actually the variance of neural network evaluations when the same position is reached on different paths.
Moreover, having a history of past moves increases the chance of overfitting 
of the value prediction on small data sets because it is the same for all past seven board positions.
Finally, the policy can adapt itself to a similar playstyle to the board history including its and the opponents move history.
As a downside, however, we lose the attention mechanism due to the move history.
\renewcommand\arraystretch{1.25}
\begin{table}[]
\centering
\caption{Plane representation for crazyhouse. The features are encoded as a binary maps and features with $\ast$ are single values set over the entire $8\times 8$ plane.}
\vspace{.2cm}
\begin{tabular}{lccl}
\toprule
%
\label{tab:input_representation}
\textbf{Feature} & \textbf{Planes} & \textbf{Type}                         & \textbf{Comment}                                                            \\ \midrule
P1 piece                          & 6                                & bool                      & order: \{\texttt{PAWN}, \texttt{KNIGHT}, \texttt{BISHOP}, \texttt{ROOK}, \texttt{QUEEN}, \texttt{KING}\}                                     \\
P2 piece                          & 6                                & bool                      & order: \{\texttt{PAWN}, \texttt{KNIGHT}, \texttt{BISHOP}, \texttt{ROOK}, \texttt{QUEEN}, \texttt{KING}\}                                     \\
Repetitions\textsuperscript{*}                      & 2                                & bool                      & \multicolumn{1}{l}{indicates how often the board positions has occurred} \\
P1 pocket count\textsuperscript{*}                  & 5                                & \multicolumn{1}{c}{int} & order: \{\texttt{PAWN}, \texttt{KNIGHT}, \texttt{BISHOP}, \texttt{ROOK}, \texttt{QUEEN}\}                                           \\
P2 pocket count\textsuperscript{*}                  & 5                                & \multicolumn{1}{c}{int} & order: \{\texttt{PAWN}, \texttt{KNIGHT}, \texttt{BISHOP}, \texttt{ROOK}, \texttt{QUEEN}\}                                           \\
P1 Promoted Pawns                 & 1                                & bool                      & indicates pieces which have been promoted                                    \\
P2 Promoted Pawns                 & 1                                & bool                      & indicates pieces which have been promoted                                    \\
En-passant square                 & 1                                & bool                      & indicates the square where en-passant capture is possible                        \\
Colour\textsuperscript{*}                           & 1                                & bool                      & all zeros for black and all ones for white                                                   \\
Total move count\textsuperscript{*}                 & 1                                & int                      & sets the full move count (FEN notation)                                         \\
P1 castling\textsuperscript{*}                      & 2                                & bool                      & binary plane, order: \{\texttt{KING\_SIDE}, \texttt{QUEEN\_SIDE}\}                                      \\
P2 castling\textsuperscript{*}                      & 2                                & bool                      & binary plane, order: \{\texttt{KING\_SIDE}, \texttt{QUEEN\_SIDE}\}                                      \\
No-progress count\textsuperscript{*}                & 1                                & int                      & sets the no progress counter (FEN halfmove clock)                                \\ \midrule
Total                             & 34                               &                              &                                                                                           \\
\bottomrule
\end{tabular}
\end{table}

To make the board presentation fully Markovian, we add an additional feature channel, which highlights the square for an en-passent capture if possible.
In contrast to standard chess, piece promotions are a lot more common and often occur multiple times in a game. If a promoted piece gets captured, it will be demoted to a pawn again and added to the pocket of the other player.
To encode this behaviour, we highlight each square of a piece that has been promoted using a binary map for each player.
Overall, the representation to \alg{CrazyAra} is fully compatible with the standard Forsyth–Edwards Notation (FEN) for describing a particular board position.
However, the information of how often a particular position has already occurred gets lost when converting our representation into FEN.

In over-the-board (OTB) games, the players see their pieces usually in the first rank.
We make use of symmetries by flipping the board representation on the turn of the second player, so that the starting square of the queen is always to the left of the king for both players.

\ifarxiv
\begin{figure}[]
\ifarxiv
\else
\subsection{Supplementary Figures and Tables}
\vspace{\floatsep}
\vspace{\floatsep}
\fi
\centering
\begin{minipage}{0.49\textwidth}
\ifarxiv
\subcaptionbox{Combined loss as described in Equation \ref{eq:combined_loss}}{
  \includegraphics[width=\textwidth]{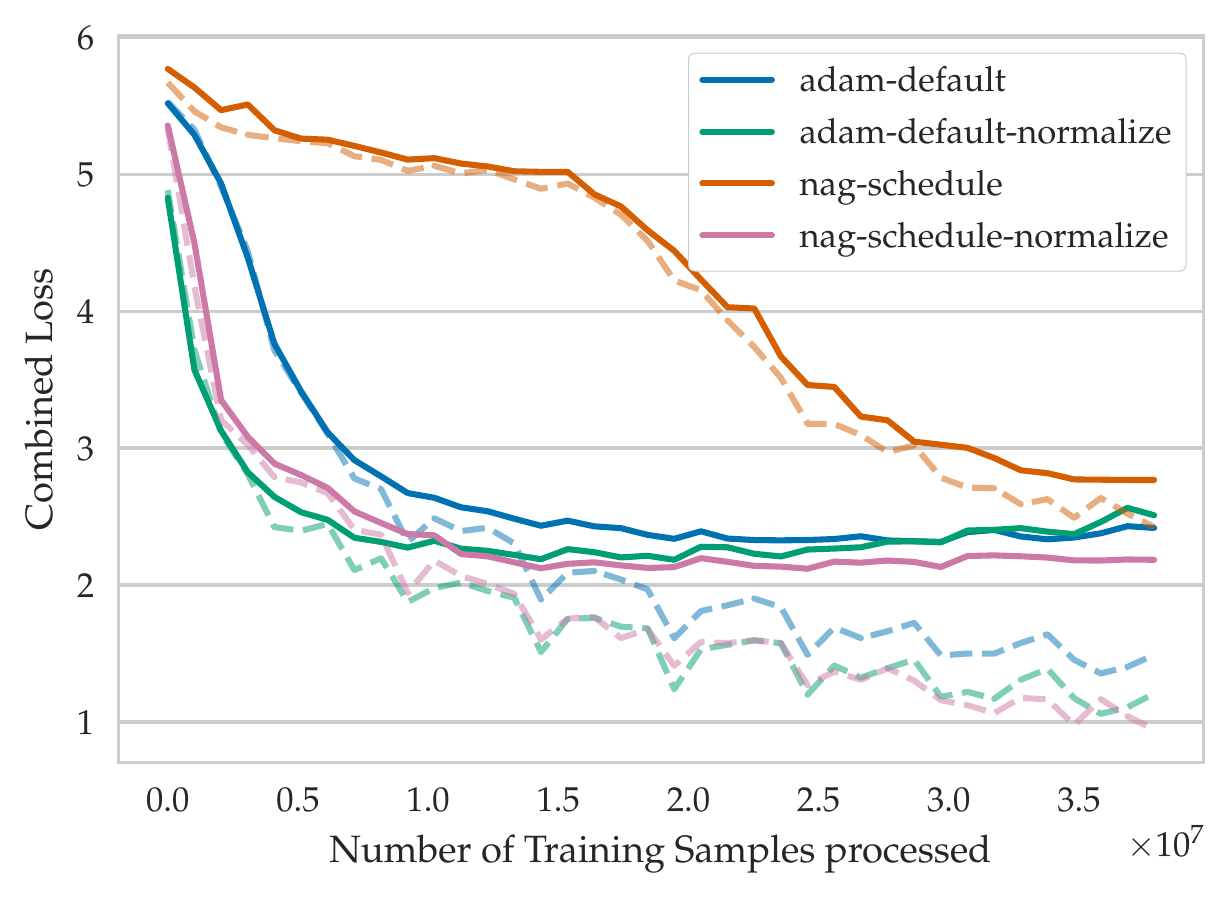}
  }
\else
\subcaptionbox{Combined loss}{
  \includegraphics[width=\textwidth]{figures/supervised_learning/normalize_exp/loss.pdf}
  }  
\fi
\end{minipage}
\begin{minipage}{0.49\textwidth}
    \subcaptionbox{Policy accuracy}{
  \includegraphics[width=\textwidth]{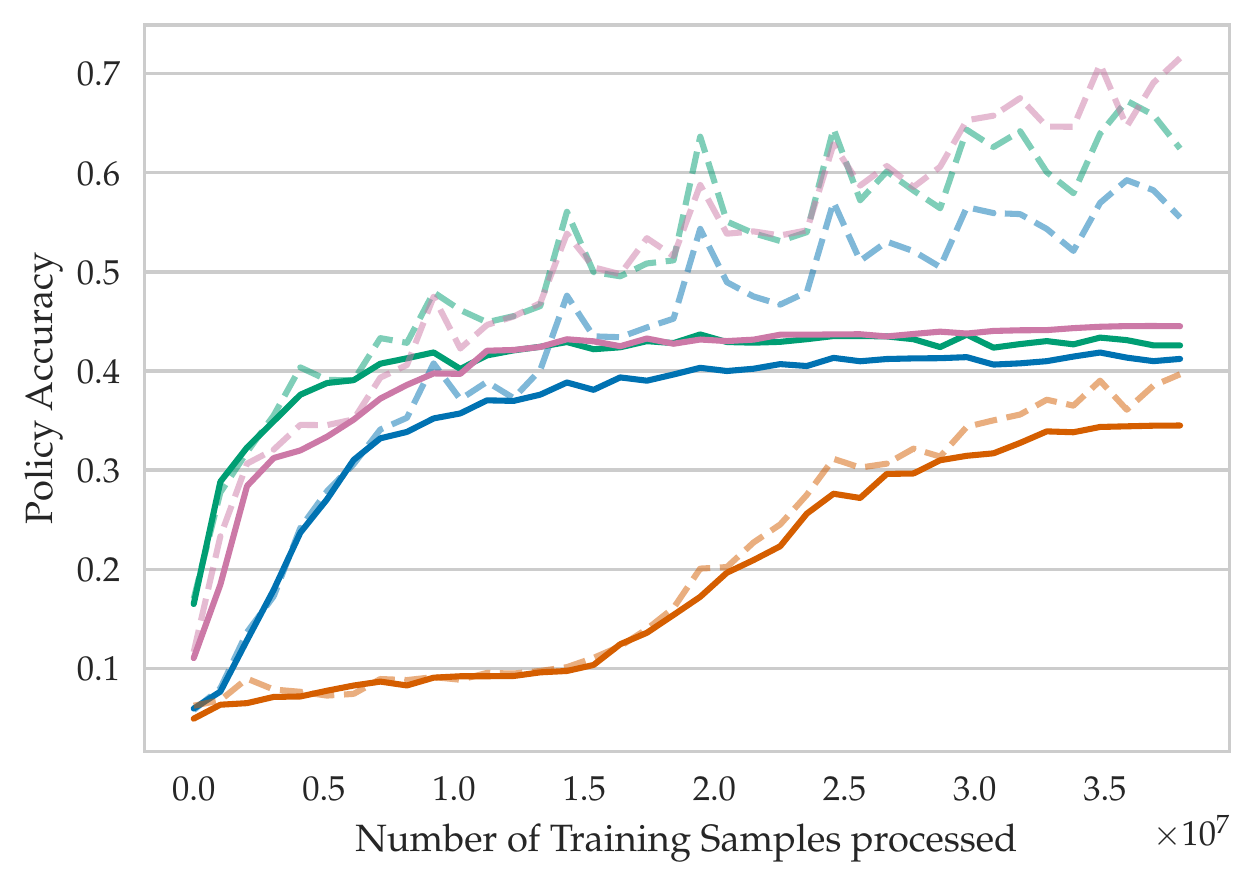}
  }
\end{minipage}
\caption{Learning progress when training on a subset of the lichess.org data set using \num{10000} training games with minimum Elo of \num{1600} for both players. The model used was \model{4-value-8-policy}, $7\times256$ (Table~\ref{tab:A0_arch}, Table~\ref{tab:value_head}). \textit{NAG} stands for the Nesterov Accelerated Gradients. Solid lines and dashed lines describes the score on the validation data set and training set respectively.}
\label{fig:normalize_exp}
\end{figure}
\addtocounter{subfigure}{-2}
\fi
\subsection{Input normalization}
All input planes are scaled to the linear range of $[0,1]$. For most computer vision tasks, each channel of an image is encoded as $8$ bit per pixel resulting in a discrete value between $[0,255]$.
Commonly, these input images are divided by \num{255} to bring each pixel in the $[0,1]$ range.
For ImageNet \citep{russakovsky2015imagenet}, 
it is also frequent practice to subtract the mean image from the train data set for every input.
In the case of crazyhouse chess, we define the following constants, which act as a maximum value for each non-binary feature, see Table~\ref{tab:input_representation}. We set the maximum number of pocket pieces for each piece type to \num{32}, the maximum number of total moves to \num{500}, and the maximum value of the no progress counter to \num{40}\cfootnote{According to the fifty-move rule a value of \num{50} is recommended instead.}.
Next we divide each correspond feature of an input sample with these maximum values.
Note that these maximum values only describe a soft boundary and can be violated without breaking the neural network predictions.
Some values such as the maximum number of pocket pieces and the maximum number of moves could have also been set to a different similar value.

To illustrate the benefits of including this step, we conducted the following experiments, learning curves shown in
\ifarxiv
Figure~\ref{fig:normalize_exp}.
\fi
We trained a small AlphaZero like network
\ifarxiv
, see Table~\ref{tab:A0_arch},
\fi
with seven residual blocks on a subset of our training data using \num{10000} games.
For the optimizer we used ADAM \citep{kingma_adam:_2014} with its default parameters: Learning-rate \num{=0.001}, $\beta_1 = 0.9$,  $\beta_2 = 0.999$, $\epsilon = 10{-8}$ and Stochastic Gradient Descent with Neterov's Momentum (NAG;  \citealt{botev2017nesterov}) using our learning rate and momentum schedule, cf.~Section \ref{sec:supervised_training}.
When training both optimizer with and without normalization for seven epochs with a weight decay of $10^{-4}$ and a batch-size of \num{1024}, one can make the following observations. Both optimizers highly benefit in terms of convergence speed and final convergence when using our input pre-processing step. ADAM gains $+2,3\,\%$ whereas NAG gains $+9,2\,\%$ move prediction accuracy.
The ADAM optimizer is much more robust when dealing with the unnormalized feature set due to its internal automatic feature rescaling, but is outperformed in terms of generalization ability when using NAG with our defined learning and momentum schedule. This agrees with research on different optimizer \citep{keskar2017improving}.

\subsection{Illustrative example for predictions}

\begin{figure}[tp]
\begin{minipage}[tp]{.49\textwidth}
 %
 \center
\vspace{-0.35cm}
\subcaptionbox{Classical board representation}{
\resizebox{0.725\textwidth}{!}{
\chessboard[
pgfstyle={text},
text=\;\qquad\WhiteRookOnWhite,
markfield={h8},
text=\;\qquad\BlackBishopOnWhite \cnt{2}, 
markfield={h7},
text=\;\qquad\BlackKnightOnWhite \cnt{2},
markfield={h6},
text=\;\qquad\BlackPawnOnWhite,
markfield={h5},
marginwidth=0.4cm,
setfen=r2q3k/ppp2p1p/2n1pN2/3pP3/3P4/4BB2/PPP2PPP/R2Q1RK1 w - - 4 22]}
}
\end{minipage}%
\hfill
\begin{minipage}[tp]{.49\textwidth}
    \subcaptionbox{Pseudo color visualization}{
  \includegraphics[width=0.835\linewidth]{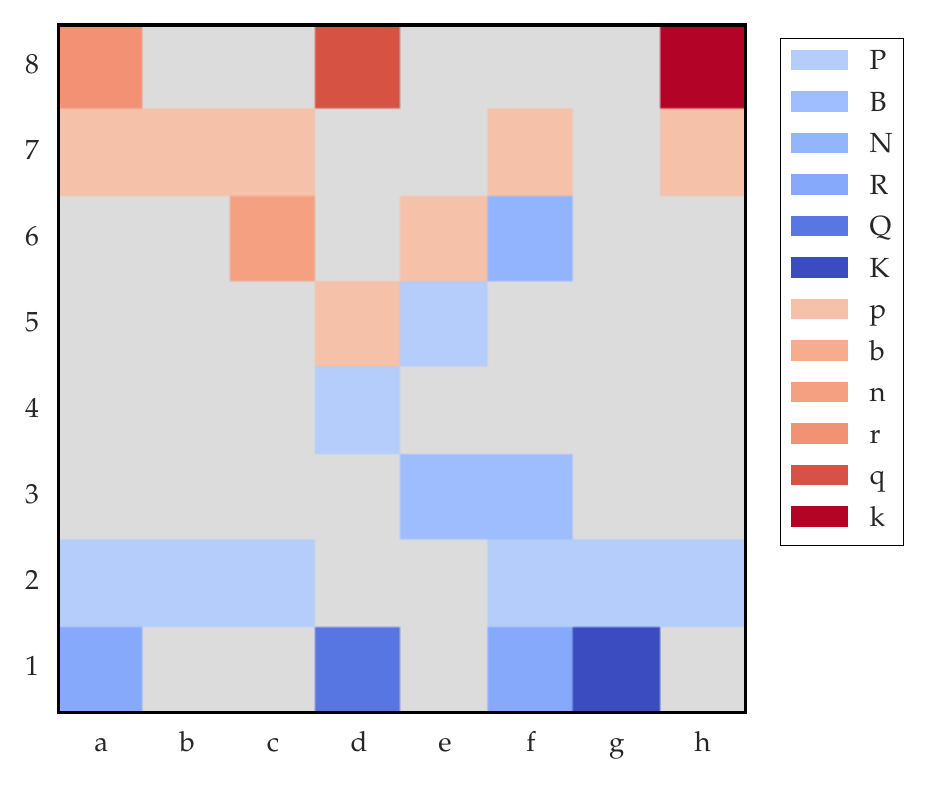}
  }
\end{minipage}
\centering
\ifarxiv
\def\sizeplanerepr{1.06}
\fi
\begin{minipage}{\sizeplanerepr\textwidth}
\centering
\subcaptionbox{Plane representation}{
\includegraphics[width=15cm]{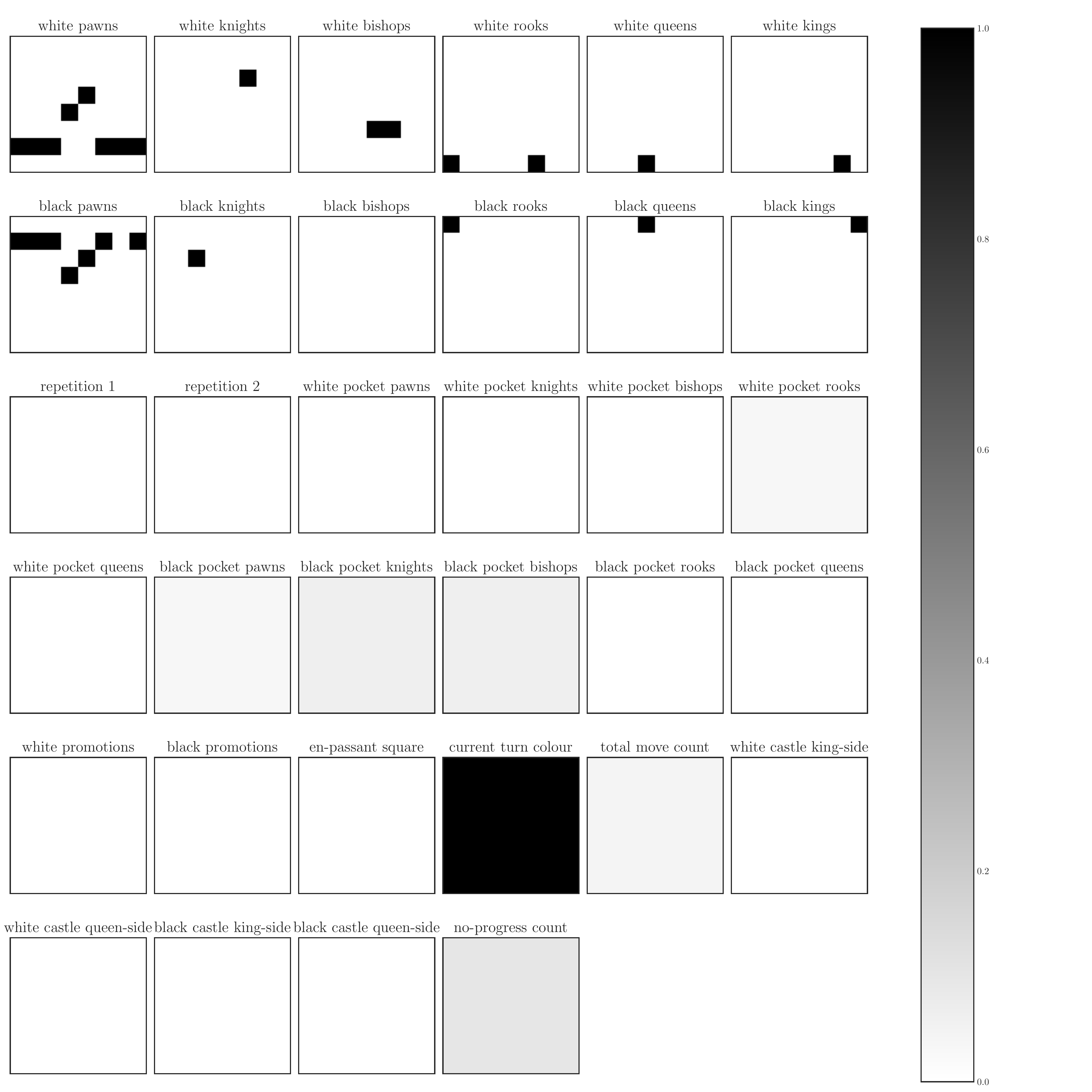}
}
\caption{Plane representation of an exemplary game position of our test data set, \\FEN:
\ifarxiv
\normalsize{\texttt{r2q3k/ppp2p1p/2n1pN2/3pP3/3P4/4BB2/PPP2PPP/R2Q1RK1[Rbbnnp] w - - 4 22}}
\fi
}
\label{fig:test_chessboard}
\end{minipage}
\end{figure}
\addtocounter{subfigure}{-3}
\begin{figure}
\centering
\begin{minipage}{0.49\textwidth} 
\centering
\subcaptionbox{Features after conv0-batchnorm0-relu0\label{fig:first_act_maps}}{
\ifarxiv
  \includegraphics[width=0.85\textwidth]{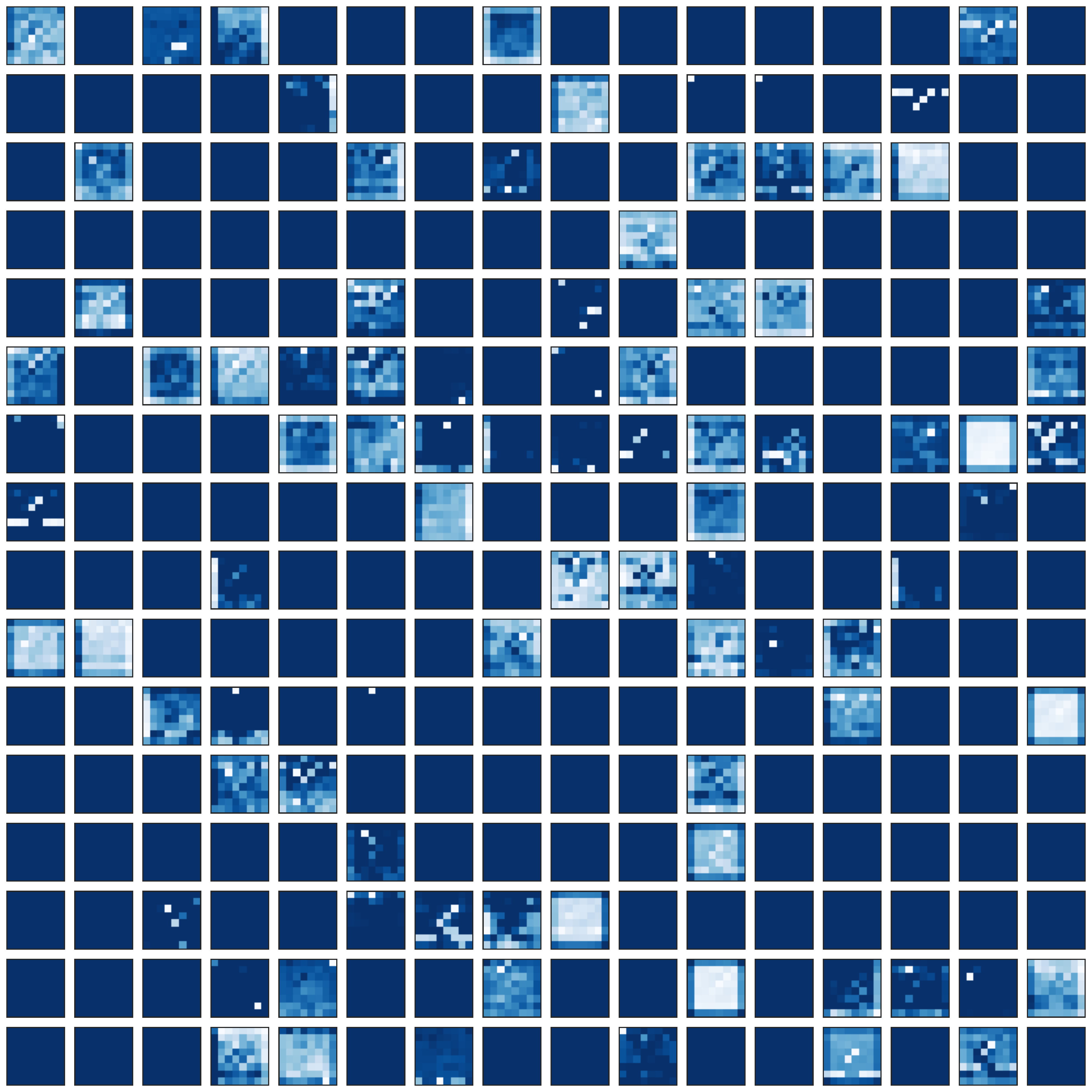}
  \fi
}
\end{minipage}
\hfill
\begin{minipage}{0.49\textwidth}
\centering
\subcaptionbox{Features after the full residual tower\label{fig:final_act_maps}}{
\ifarxiv
      \includegraphics[width=0.85\textwidth]{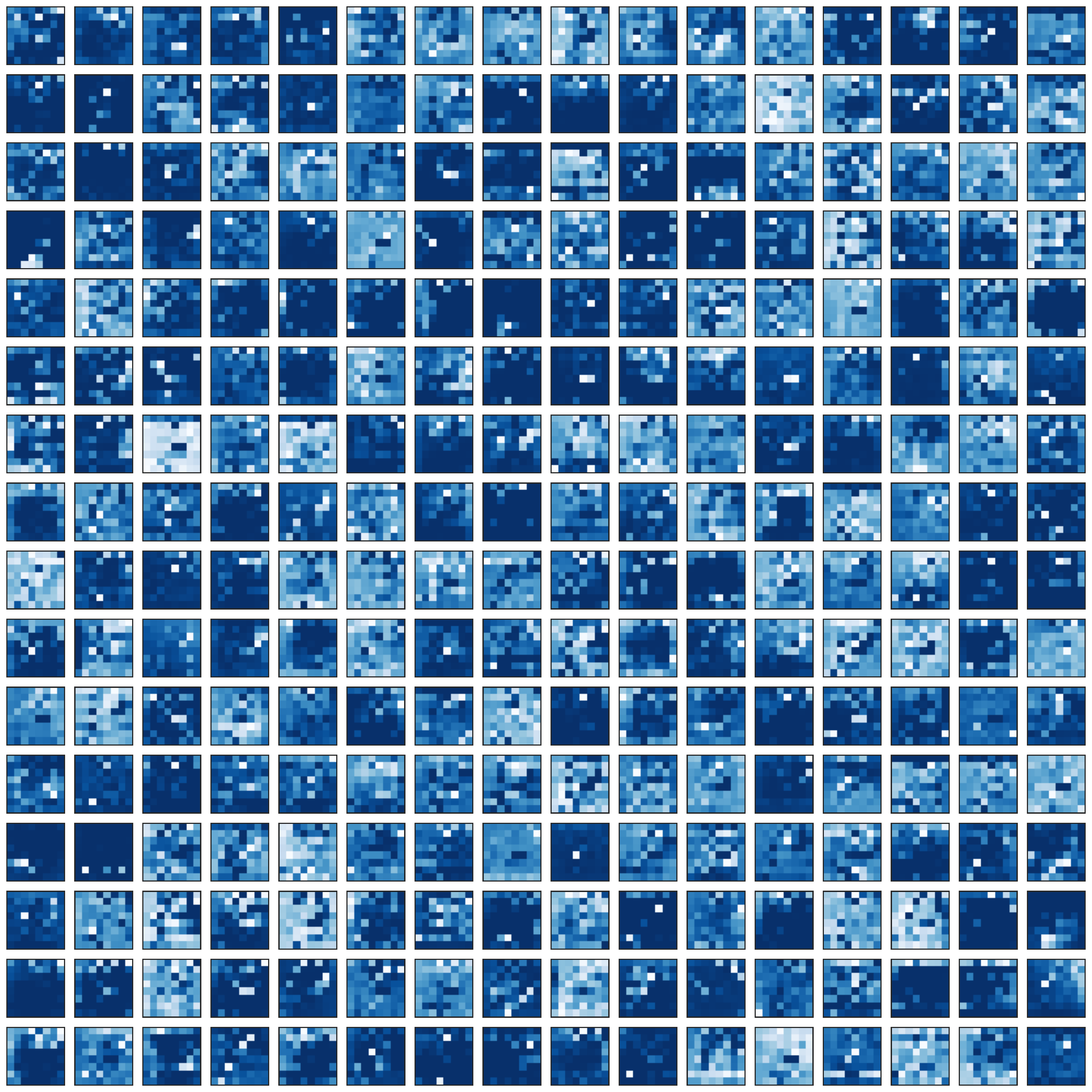}
\fi
}
\end{minipage}
\input{chessboards/value_policy_activations.tex}
%
%
\ifarxiv
\input{chessboards/candidate_moves.tex}
\fi
\caption{Activation maps of model \model{4-value-8-policy} when processing input sample (Figure~\ref{fig:test_chessboard}) from the test data set}
\label{fig:act_maps}
\end{figure}
\ifarxiv
\addtocounter{subfigure}{-6}
\fi
%

Essentially, we treat learning to play crazyhouse as modified computer vision problem where the neural network~$f_{\theta}$ conducts a classification for the policy prediction combined with a regression task for the value evaluation.

Figure~\ref{fig:test_chessboard} visualizes our input representation using an exemplary board position of the first game in our test set.
The pseudo color visualisation, which neglects the pocket information of both players for better visualization purposes, shows how the neural network receives the board as a $8\times8$ multi-channel image.

Next, the activation maps of a fully trained network (Section~\ref{sec:model_arch}) for a single prediction are shown in Figure~\ref{fig:act_maps}.
The activation maps in Figure~\ref{fig:first_act_maps}
are the extracted features after the first convolutional layer and in Figure~\ref{fig:final_act_maps}
after the full residual tower.
In the initial features, the pawn formations for black and white are still recognizable and the features are sparse due to the usage of ReLU activation and our compact input representation. The final features visually exhibit a high contrast and are fed to both the value and policy evaluation head.

The final spatial activation map of the value head are summed along the first axis and overlayed on top of the given board position in Figure~\ref{fig:value_act_maps}. 
Dark orange squares represent high activation regions and unsaturated those of low importance.
The value evaluation fundamentally differs from those used in other crazyhouse engines and fully relies on the estimated winning chances by the neural network for a particular position.
Formulating a handcrafted value evaluation for crazyhouse is challenging because game positions often require high depth in order to reach a relatively stable state and the element of initiative and king safety is intensified and hard to encode as a linear function.
In the case of \textit{Stockfish}, its specialist list of chess hyper parameters  including the values for pieces on the board and in the pocket, \ctexttt{KingAttackWeights}, \ctexttt{MobilityBonus} and \ctexttt{QueenThreats} has been carefully fine-tuned for the crazyhouse variant with the help of a SPSA Tuner \citep{kiiski_spsa_2019, fichter_spsa_2018}.

Based on the games of \CrazyAra with other crazyhouse engines and human players, it seems to attach high importance to initiative and often seemingly, intuitively sacrifices pieces in tactical or strategic positions in order to increase the activity of its pieces. This often leads to strong diverging position evaluations compared to other engines (Section~\ref{sec:strength_eval}) because the value evaluation of most engines is fundamentally based on material.
\CrazyAra seems to have a higher risks of choosing a sacrifice which was not sufficiently explored or missing a tactical sequence rather than slowly losing due to material disadvantage.
The tendency for an active play-style is primarily due to a prominent proportion of human crazyhouse players in our data set which are known to play aggressively.
Moreover, most players are more prone to make a mistake when being under attack and in time trouble which influences the value evaluation.
On the other hand, there are also players which prefer consolidation moves instead of risking a premature attack. During the course of training, the network converges to the best fit limited by its model expressiveness which captures the play style of the majority of the games in our data set.
\ifarxiv
In Figure~\ref{fig:policy_act_maps}, the final spatial features of the policy head are shown and are summed along the first axis.
In this case, the network had the flexibility to generate its own policy maps and primarily highlights squares where it either suggests to drop or move a piece.
Figure~\ref{fig:example_candidate_moves} shows the first ten candidate moves, and their respective softmax activation values are presented in Figure~\ref{fig:sorted_policy_prediction}.
In this instance the move with the highest activation, \ctexttt{e3h6}, coincides with the choice of the respective professional human player\cfootnote{The position occurred in a game between \textit{FM Lastro} and \textit{mahdiafshari63} on move 22 with white to play.\\\textit{FM Lastro} played the move \ctexttt{Bh6} ({\url{https://lichess.org/jsPkVxaU\#42})}, accessed 2019-08-19} and the network returns a position evaluation of $+0.81$ from which the first player managed to win the game.
\fi
%
%

\section{Output Representation of \alg{CrazyAra}}
\label{sec:output_representation}
The value output, which represents the winning chances for a game position, represents as a single floating point value in the range $[-1,+1]$ as described in Section~\ref{sec:DNN}.

For the policy output we investigate two conceptually different representations.
First, the policy head consists of a single convolutional layer with a variable amount of feature channels followed by a fully connected layer storing all $\num{2272}$ theoretical plausible moves as described in UCI-notation.
Each entry in this vector stands for a particular predefined move and queen promoting moves are treated as individual moves, denoted with a suffix \ctexttt{q}, e.g. \ctexttt{e7e8q}.
This form of policy representation was also used by\cfootnote{\url{https://github.com/Zeta36/chess-alpha-zero}, accessed 2019-06-08}.

\ifarxiv
\begin{table}[]
\centering
\caption{\normalsize{Policy map representation for crazyhouse}}
\vspace{.2cm}
\begin{tabular}{lcl}
    \toprule
    \textbf{Feature} &	\textbf{Planes} & \textbf{Comment}  \\
    \midrule
    Queen moves            & 56 & direction order: \{\texttt{N}, \texttt{NE}, \texttt{E}, \texttt{SE}, \texttt{S}, \texttt{SW}, \texttt{W}, \texttt{NW}\} with 7 lengths per direction  \\
    Knight moves           & 8 & move order: \{\texttt{2N1E}, \texttt{1N2E}, \texttt{1S2E}, \texttt{2S1E}, \texttt{2S1W}, \texttt{1S2W}, \texttt{1N2W}, \texttt{2N1W}\}  \\
    Promotions             & 12 & piece order: \{\texttt{KNIGHT}, \texttt{BISHOP}, \texttt{ROOK}, \texttt{QUEEN}\} \\
    Drop moves                 & 5 & piece order: \{\texttt{PAWN}, \texttt{KNIGHT}, \texttt{BISHOP}, \texttt{ROOK}, \texttt{QUEEN}\} \\
    \midrule
    Total                  & 81 & \\
    \bottomrule
\end{tabular}
\label{tab:policy_output}
\end{table}
\fi

In the second version, the policy head representation directly encodes each of the $\num{2272}$ moves on a particular square on $\num{81}$ channels of $8\times8$
\ifarxiv
 planes as shown in Table \ref{tab:policy_output}:
\fi
first come the queen moves in all compass direction, followed by all knight moves for each square, promoting moves and finally all dropping moves.
Queen moves also include all pawn, bishop, rook and king moves.
We spent three additional channels for queen promotion moves, to be consistent with the
\ifarxiv
UCI-move-representation.
\fi
However, please keep in mind that most squares of the $\num{5184}$ values describe illegal moves. This is due to the fact that the corresponding move would lead outside of the board and that promoting moves are only legal from the second last row on.

In the UCI-move-representation, en-passant moves and castling moves do not have separate vector indices and king side and queen side castling is denoted as the move \ctexttt{e1g1} and \ctexttt{e1c1} respectively.
Treating these as special move or as \textit{king captures rook}, which would ensure a better compatibility with the chess960 --- also known as Fischer random variant, is a valuable alternative for the network.

\section{Deep Network Architecture of \alg{CrazyAra}}
\label{sec:model_arch}

Finding a (deep) neural network for playing board games covers three criteria, which determine the playing strength when using the learnt model in MCTS.
First, the performance, i.\,e., the policy and value loss on the test data set is the main component for predicting the best moves at each roll-out.
Second, a better performance is often associated with a longer training and inference time leading to less evaluations per second during search.
Third, the memory consumption per prediction specifies the maximum achievable batch-size. 





To address these points, \alg{CrazyAra} makes use of a dual architecture design with a tower of residual blocks followed by a value and policy head as
\ifarxiv
described in Table \ref{tab:A0_arch} and
\fi
recommended by~\cite{silver2017mastering}.
The originally proposed \alg{AlphaZero} architecture differs from most common computer vision architectures in several ways: there are no down-sampling operation such as max-pooling, average pooling or strided convolution to preserve the spatial size and no increase in feature channels per layer. Consquently, the number of parameters is rather high and most comparable with WideResnet  \citep{zagoruyko2016wide}.
Moreover, in the final reduction step for selecting the classes, convolutional layers are used instead of a global average pooling layer due to the spatial importance of defining moves.

\ifarxiv
%
%
\begin{table}[tp]
\centering
\caption{\normalsize{\alg{AlphaZero's} network architecture: $19 \times 256$}}
\vspace{.2cm}
\begin{tabular}{cccccc}
\toprule
\multicolumn{2}{c}{\textbf{Layer Name}}                                                                  & \multicolumn{2}{c}{\textbf{Output Size}} & \multicolumn{2}{c}{\begin{tabular}[c]{@{}c@{}}\textbf{AlphaZero Resnet}\\ \textbf{39-layer}\end{tabular}}                                  \\ \midrule
\multicolumn{2}{c}{\begin{tabular}[c]{@{}c@{}}\ctexttt{conv0}\\ \ctexttt{batchnorm0}\\ \ctexttt{relu0}\end{tabular}}          & \multicolumn{2}{c}{$256\times8\times8$}     & \multicolumn{2}{c}{conv $3\times3,\:256$}                                              \\ \midrule 
\multicolumn{2}{c}{\begin{tabular}[c]{@{}c@{}}\ctexttt{res\_conv0\_x}\\ \ctexttt{res\_batchnorm0\_x}\\ \ctexttt{res\_relu0\_x}\\ \ctexttt{res\_conv1\_x}\\ \ctexttt{res\_batchnorm1\_x}\\ \ctexttt{shortcut\;+\;output}\\ \ctexttt{res\_relu1\_x}\end{tabular}} & \multicolumn{2}{c}{$256\times8\times8$}     & \multicolumn{2}{c}{\begin{tabular}[c]{@{}c@{}} \hspace{0.8cm}\ldelim[{\bracketscale}{1.0cm}[]\hspace{-0.8cm} conv $3\times3,\:256$ \rdelim]{\bracketscale}{1.0cm}[\normalfont $\times 19$] \\ conv $3\times3,\:256$  \end{tabular}}                                              \\ \midrule
\textit{\ctexttt{value head}}                                      & \textit{\ctexttt{policy head}}                                     & $\;1$             & \begin{tabular}[c]{@{}c@{}} $2272$ /\\ $5184$\end{tabular}         & \hspace{0.8cm}Table \ref{tab:value_head} & \begin{tabular}[c]{@{}c@{}} Table \ref{tab:policy_head}\\ Table \ref{tab:policy_map}\end{tabular} \\ \bottomrule
\end{tabular}
\label{tab:A0_arch}
\end{table}
\begin{table}[tp]
\centering
\label{tab:preact_arch}
\caption{\normalsize{\model{8-value-policy-map-preAct-relu+bn}: $19 \times 256$}}
\vspace{.2cm}
\begin{tabular}{cccccc}
\toprule
\multicolumn{2}{c}{\textbf{Layer Name}}                                                                  & \multicolumn{2}{c}{\textbf{Output Size}} & \multicolumn{2}{c}{\begin{tabular}[c]{@{}c@{}}\textbf{AlphaZero Resnet}\\ \textbf{39-layer (pre-activation)}\end{tabular}}                                  \\ \midrule 
\multicolumn{2}{c}{\begin{tabular}[c]{@{}c@{}}\ctexttt{conv0}\\ \ctexttt{batchnorm0}\\ \ctexttt{relu0}\end{tabular}}          & \multicolumn{2}{c}{$256\times8\times8$}     & \multicolumn{2}{c}{conv $3\times3,\:256$}                                              \\
\midrule
\multicolumn{2}{c}{\begin{tabular}[c]{@{}c@{}}\ctexttt{res\_batchnorm0\_x}   \\ \ctexttt{res\_conv0\_x} \\ \ctexttt{res\_batchnorm1\_x}\\ \ctexttt{res\_relu0\_x}\\ \ctexttt{res\_conv1\_x}\\ \ctexttt{res\_batchnorm2\_x}\\ \ctexttt{shortcut\;+\;output}\end{tabular}} & \multicolumn{2}{c}{$256\times8\times8$}     & \multicolumn{2}{c}{\begin{tabular}[c]{@{}c@{}} \hspace{0.8cm}\ldelim[{\bracketscale}{1.0cm}[]\hspace{-0.8cm} conv $3\times3,\:256$ \rdelim]{\bracketscale}{1.0cm}[\normalfont $\times 19$] \\ conv $3\times3,\:256$  \end{tabular}}                                              \\
\midrule
\multicolumn{2}{c}{\begin{tabular}[c]{@{}c@{}}\ctexttt{batchnorm1}\\ \ctexttt{relu1}\end{tabular}}          & \multicolumn{2}{c}{$256\times8\times8$}     & \multicolumn{2}{c}{ }                                              \\
\midrule
\textit{\ctexttt{value head}}                                      & \textit{\ctexttt{policy head}}                                     & $\;1$             & \begin{tabular}[c]{@{}c@{}} $2272$ /\\ $5184$\end{tabular}            & \hspace{0.8cm}Table \ref{tab:value_head} & \begin{tabular}[c]{@{}c@{}} Table \ref{tab:policy_head}\\ Table \ref{tab:policy_map}\end{tabular} \\ \bottomrule
\end{tabular}
\end{table}

\fi
\begin{table}[tp]
\centering
\caption{\model{RISEv2 mobile} / \textit{8-value-policy-map-mobile architecture}: $13 \times 256$}
\label{tab:risev2_mobile}
\vspace{.2cm}
\begin{tabular}{cccccc}
\toprule
\multicolumn{2}{c}{\textbf{Layer Name}} & \multicolumn{2}{c}{\textbf{Output Size}} & \multicolumn{2}{c}{\begin{tabular}[c]{@{}c@{}}\textbf{RISEv2 mobile}\\ \textbf{40-Layer}\end{tabular}} \\
\midrule
\multicolumn{2}{c}{\begin{tabular}[c]{@{}c@{}}\ctexttt{conv0}\\ \ctexttt{batchnorm0}\\ \ctexttt{relu0}\end{tabular}} & \multicolumn{2}{c}{$256\times8\times8$} & \multicolumn{2}{c}{conv $3\times3,\:256$} \\
\midrule
\multicolumn{2}{c}{\begin{tabular}[c]{@{}c@{}}\ctexttt{res\_conv0\_x}\\ \ctexttt{res\_batchnorm0\_x}\\ \ctexttt{res\_relu0\_x}\\ \ctexttt{res\_conv1\_x}\\ \ctexttt{res\_batchnorm1\_x}\\ \ctexttt{res\_relu1\_x}\\ \ctexttt{res\_conv2\_x}\\ \ctexttt{res\_batchnorm2\_x}\\ \ctexttt{shortcut\;+\;output}\end{tabular}} & \multicolumn{2}{c}{$256\times8\times8$} & \multicolumn{2}{c}{\begin{tabular}[c]{@{}c@{}} \hspace{0.5cm}\ldelim[{4}{.8cm}[]\hspace{-0.cm} (SE-Block, $r=2$)  \hspace{0.6cm} \rdelim]{4}{0.8cm}[\normalfont $\times 13$]\\ conv $1\times1, 128+64\,x$\\ dconv $3\times3, 128+64\,x$\\ conv $1\times1,\:256$\end{tabular}} \\
\midrule
\textit{\ctexttt{value head}} & \textit{\ctexttt{policy head}} & $\;1$ & \begin{tabular}[c]{@{}c@{}} $2272$ /\\ $5184$\end{tabular} &
\ifarxiv
\hspace{0.8cm}Table \ref{tab:value_head} & \begin{tabular}[c]{@{}c@{}}Table \ref{tab:policy_head}\\ Table \ref{tab:policy_map}\end{tabular}
\fi
\\
\bottomrule
\end{tabular}
\end{table}

\ifarxiv
\begin{table}[tp]
\centering
\caption{\normalsize{Value head for different architectures with $n$-channels}}
\label{tab:value_head}
\vspace{.2cm}
\begin{tabular}{ccc}
\toprule
\textbf{Layer Name}                                                             & \textbf{Output Size} & \textbf{Value Head N-channels}          
\\ \midrule
\begin{tabular}[c]{@{}c@{}}\ctexttt{conv0}\\ \ctexttt{batchnorm0}\\ \ctexttt{relu0}\end{tabular} & $n\times8\times8$ & conv $1\times1,\:n$ \\
\midrule
\begin{tabular}[c]{@{}c@{}}\ctexttt{flatten0}\\ \ctexttt{fully\_connected0}\\ \ctexttt{relu1}\end{tabular} & $256$ & fc,$\:256$ \\
\midrule
\begin{tabular}[c]{@{}c@{}}\ctexttt{fully\_connected1}\\ \ctexttt{tanh0}\end{tabular} & $1$ & fc,$\:1$ \\
\bottomrule
\end{tabular}
\end{table}
\begin{table}[]
\centering
\caption{\normalsize{Policy head type with $n$-channels}}
\label{tab:policy_head}
\vspace{.2cm}
\begin{tabular}{ccc}
\toprule
\textbf{Layer Name}                                                             & \textbf{Output Size} & \textbf{Policy Head N-channels}          
\\ \midrule
\begin{tabular}[c]{@{}c@{}}\ctexttt{conv0}\\ \ctexttt{batchnorm0}\\ \ctexttt{relu0}\end{tabular} & $n\times8\times8$ & conv $1\times1,\:n$ \\
\midrule
\begin{tabular}[c]{@{}c@{}}\ctexttt{flatten0}\\ \ctexttt{fully\_connected0}\\ \ctexttt{softmax0}\end{tabular} & $2272$ & fc,$\:2272$ \\
\bottomrule
\end{tabular}
\end{table}
\begin{table}[tp]
\centering
\caption{\normalsize{Policy head type policy-map}}
\label{tab:policy_map}
\vspace{.2cm}
\begin{tabular}{ccc}
\toprule
\textbf{Layer Name}                                                             & \textbf{Output Size} & \textbf{Policy Map}          
\\ \midrule
\begin{tabular}[c]{@{}c@{}}\ctexttt{conv0}\\ \ctexttt{batchnorm0}\\ \ctexttt{relu0}\end{tabular} & $256\times8\times8$ & conv $3\times3,\:256$ \\
\midrule
\begin{tabular}[c]{@{}c@{}}\ctexttt{conv1}\\ \ctexttt{flatten0}\\ \ctexttt{softmax0}\end{tabular} & $5184$ & conv $3\times3,\:81$ \\
\bottomrule
\end{tabular}
\end{table} 
\fi

Residual connections 
\citep{he2016deep} play an important role for training computer vision architectures effectively.
Later the original version has been revisited in ResneXt \citep{xie2017aggregated} making use of branching in form of group convolutions.


We train several different architectures on the same training set with the same optimizer settings.\footnote{Figure \ref{fig:models_overview} was generated with the tool
\url{https://github.com/HarisIqbal88/PlotNeuralNet}, accessed 2019-08-19
}

Specifically, model \textit{4-value-8-policy}, \textit{8-value-16-policy} and \textit{8-value-policy-map} essentially follow the original \textit{AlphaZero} network architecture
\ifarxiv
(Table \ref{tab:A0_arch})
\fi
but use different value and policy heads. Specifically, 
\textit{4-value-8-policy} means that four channels in the value head and eight channels in the policy head are used. \textit{8-value-policy-map} has a policy head 
\ifarxiv
according to Table~\ref{tab:policy_output} and~\ref{tab:policy_map}.
\fi

First we tried training the original \textit{AlphaGoZero} network architecture \model{1-value-2-policy} \citep{silverd2016masteringthe}, which has one channel for the value and two channels for the policy head.
This unfortunately, did not work for crazyhouse and led to massive gradient problems, especially for deeper networks.
The reason is that the policy for crazyhouse is much more complex than in the game of Go.
In Go you can only drop a single piece type or pass a move, but in crazyhouse you can do any regular chess move and additionally drop up to five different piece types.
When only relying on two channels, these layers turn into a critical bottleneck, and the network learns to encode semantic information on squares, which are rarely used in play.
Based on our analysis of the policy activation maps, similar to Figure~\ref{fig:policy_act_maps}, we observed that for these networks, usually squares on the queen-side hold information such as the piece type to drop. In cases where these squares are used in actual play, we encountered massive gradient problems during training.
We found that at least eight channels are necessary to achieve relatively stable gradients.

\model{RISEv2-mobile} / \model{8-value-policy-map-mobile} is a new network design which replaces the default residual block with the inverted residual block of MobileNet v2 \citep{sandler2018mobilenetv2} making use of group depthwise convolutions. Moreover, it follows the concept of the Pyramid-Architecture \citep{han2017deep}: due to our more compact input representation only about half activation maps are used after the first convolution layer (Figure~\ref{fig:first_act_maps}). 
Therefore, the number of channels for the $3\times3$ convolutional layer of the first block start with $128$ channels and is increased by $64$ for each residual block reaching $896$ channels in the last block. We call this block type an operating bottleneck block due to either reducing or expanding the number of channels.

It also uses Squeeze Excitation Layers (SE; \citealt{hu2018squeeze}) which enables the network to individually enhance channels activation maps and based the winning entry of the ImageNet classification challenge ILSVRC in 2017 \citep{russakovsky2015imagenet}.
For our network we use a ration $r$ of two and apply SE-Layer to the last five residual blocks. The name RISE originates from the Resnet architecture \citep{he2016deep, xie2017aggregated}, Inception model \citep{szegedy2016rethinking, szegedy2017inception} and SE-Layers \citep{hu2018squeeze}.

Model \model{8-value-policy-map-preAct-relu+bn} replaces the default residual block with a preactivation block \citep{he2016identity} and adds an additional batchnormalization layer to each residual block as suggested by \citet{han2017deep}.

\cite{dong2017eraserelu} and \cite{zhao_rethink_2017} discovered that the common 1:1 ratio between the number of convolutional layers and ReLU activations is suboptimal and that removing the final activation in a residual block or the first activation in the case of a pre-activation can result in improvements.
Model \model{8-value-policy-map-preAct-relu+bn} and \model{RISEv2-mobile}\,/\,\model{8-value-policy-map-mobile} follow a 2:1 ratio.

Furthermore, motivated by the empirical findings\cfootnote{\url{https://github.com/pudae/tensorflow-densenet/issues/1}, accessed 2019-07-30}\cfootnote{\url{https://github.com/keras-team/keras/issues/1802}, accessed 2019-07-30}\cfootnote{\url{https://www.reddit.com/r/MachineLearning/comments/67gonq/d_batch_normalization_before_or_after_relu/}, \mbox{accessed} 2019-07-30}
we also tried flipping the conventional order of Batchnorm-ReLU into ReLU-Batchnorm.
Here, we observed a faster convergence during training, but also witnessed NaN-values. 
The model continued to converge to NaN-values even when relying on checkpoints fall-backs of a healthy model state.

\ifarxiv
As can be seen in Table~\ref{tab:inference_speed}, our proposed model
\fi
\model{8-value-policy-map-mobile} is up to three times faster on CPU and 1.4 times faster on GPU. The reason why the model does not scale as efficiently on GPU like on CPU is because group convolution and Squeeze Excitation layers are not as suited for GPU computation because they cause memory fraction \citep{ma2018shufflenet, hu2018squeeze}.
\begin{landscape}
%
\newcommand{\mowidth}{1.35}
\hfill
\begin{figure}
\begin{minipage}{\mowidth\textwidth}
\subcaptionbox{Model architecture \model{4-value-8-policy}: 19 residual blocks followed by a value head with 4 channels and a policy head with 8 channels
}{
\includegraphics[width=\textwidth]{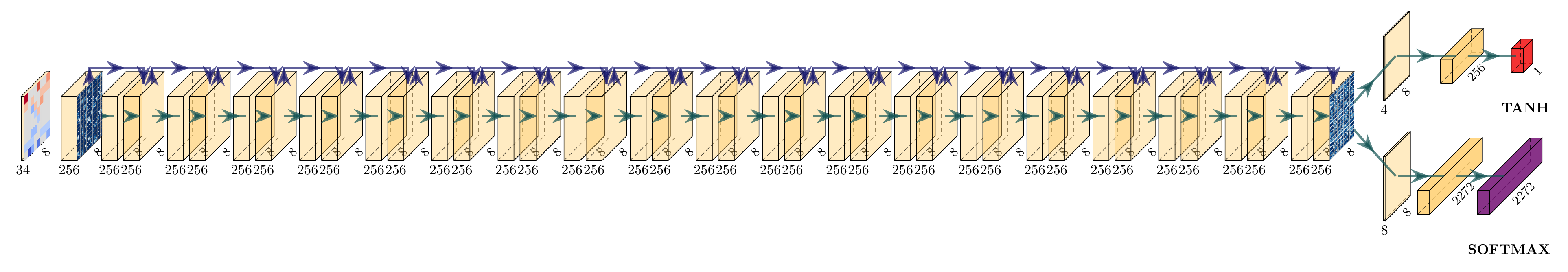}
}
\label{fig:lr_schedule}
\end{minipage}%
\hfill
\begin{minipage}{\mowidth\textwidth}
\subcaptionbox{Model \model{8-value-policy-map}: 19 residual blocks followed by a value head with 8 channels and a policy map representation
\ifarxiv
(see Table \ref{tab:policy_output})
\fi
}{
\includegraphics[width=\textwidth]{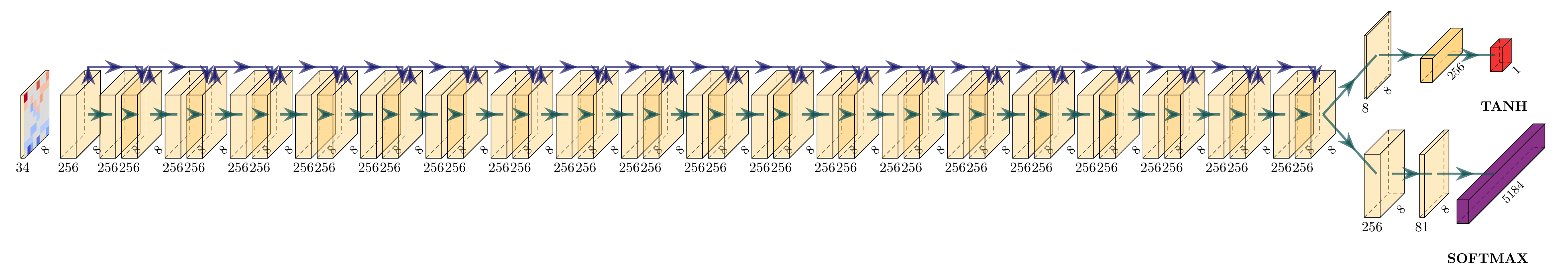}
}
\end{minipage}
\hfill
\begin{minipage}{\mowidth\textwidth}
\subcaptionbox{Model \model{RISEv2-mobile / 8-value-policy-map-mobile architecture}: 13 incrementally increasing operating bottleneck blocks}{
\includegraphics[width=\textwidth]{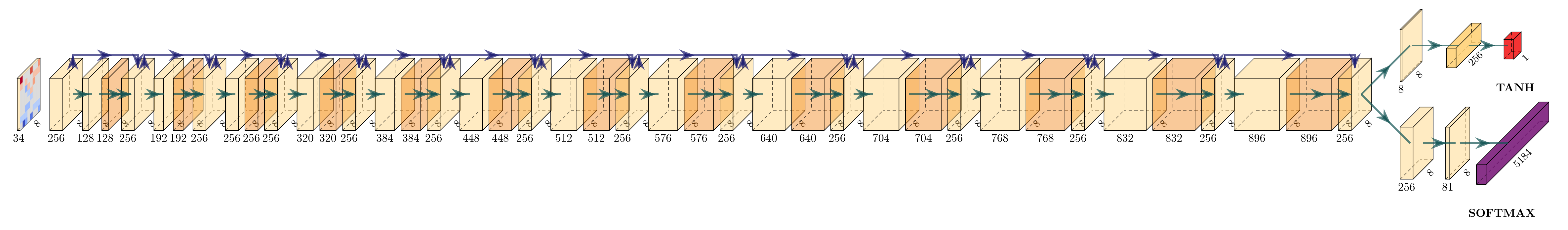}
}
\end{minipage}
\caption{Comparison of the activation map shapes for different architecture models. All models process the same input representation via a sequence of residual blocks followed by a value and policy head.}
\label{fig:models_overview}
\end{figure}
\addtocounter{subfigure}{-3}
\end{landscape}
%
\ifarxiv
\begin{figure}[tp]
\begin{center}
\includegraphics[width=\textwidth]{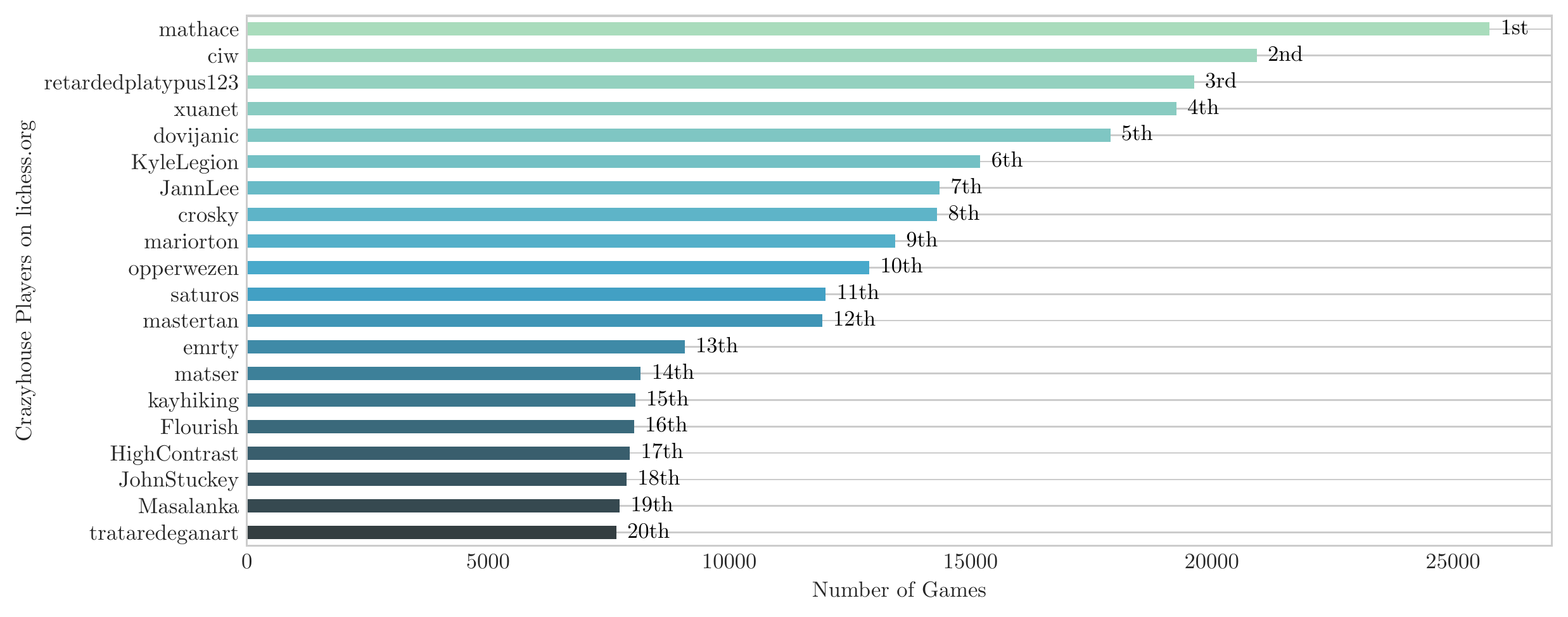}
\end{center}
\caption{Top 20 most active crazyhouse players with matches $\geq$ \num{2000} Elo for both players from January~2016 to June~2018.
The training data of \textit{CrazyAra} consists of \num{569537} games. These players took part in \num{46.03}\,\% of all games.}
\label{fig:data_players}
\end{figure}
\begin{figure}
\centering
\begin{minipage}{0.49\textwidth}
\subcaptionbox{Lichess data set}{
    \includegraphics[width=\textwidth]{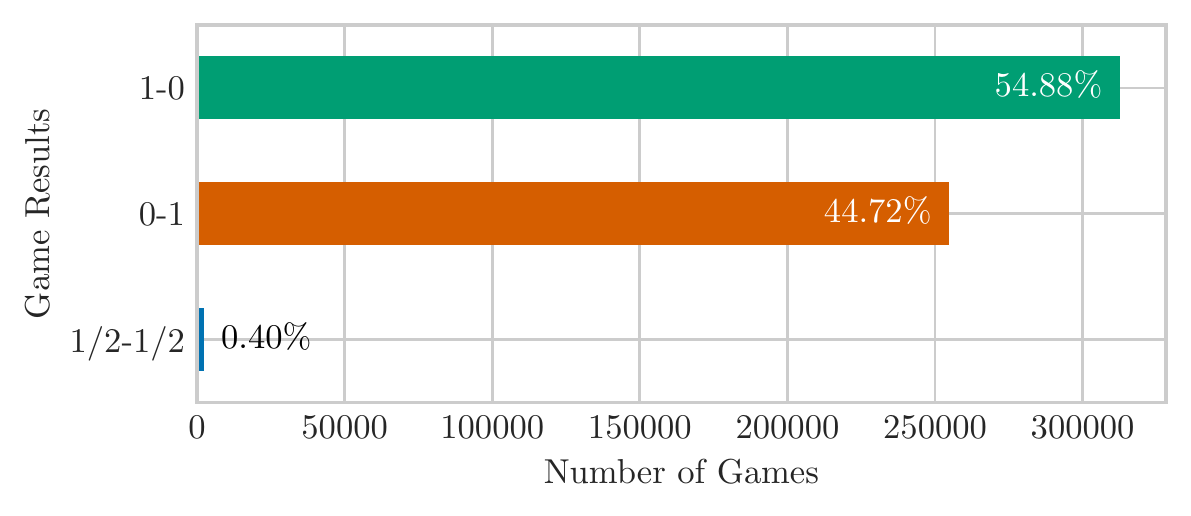}
    }
\end{minipage}
\begin{minipage}{0.49\textwidth}
\subcaptionbox{\alg{Stockfish} self play data set}{
  \includegraphics[width=\textwidth]{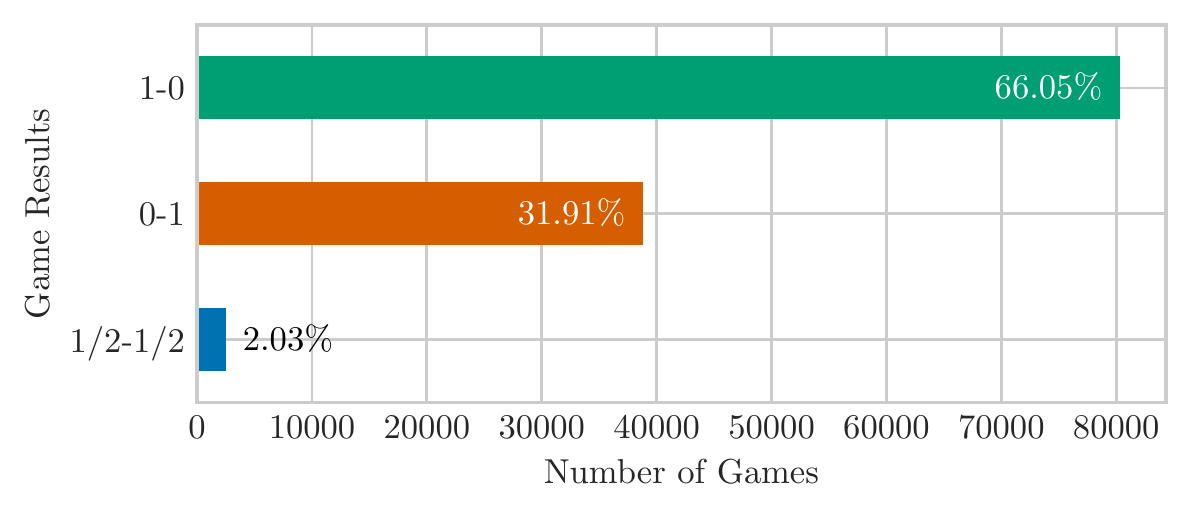}
}
\end{minipage}
\caption{Game outcomes of the lichess data set (\num{569537} games) and \alg{Stockfish} self play data set (\num{121571} games)}
\label{fig:game_outcomes}
\end{figure}
\addtocounter{subfigure}{-2}
\begin{figure}[tp]
\centering
\begin{minipage}{0.49\textwidth}
\subcaptionbox{Elo distribution for all selected games}{
  \includegraphics[width=\textwidth]{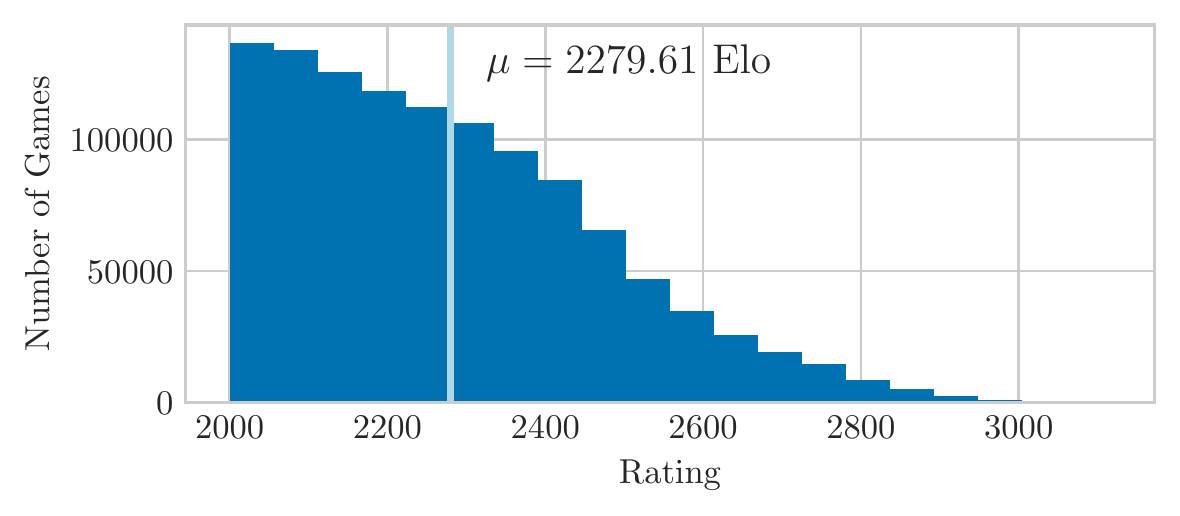}
}
\end{minipage}
\begin{minipage}{0.49\textwidth}
\subcaptionbox{Time control (\cstexttt{total\ move\ time[s]\ +\ increment[s]})}{
\includegraphics[width=\textwidth]{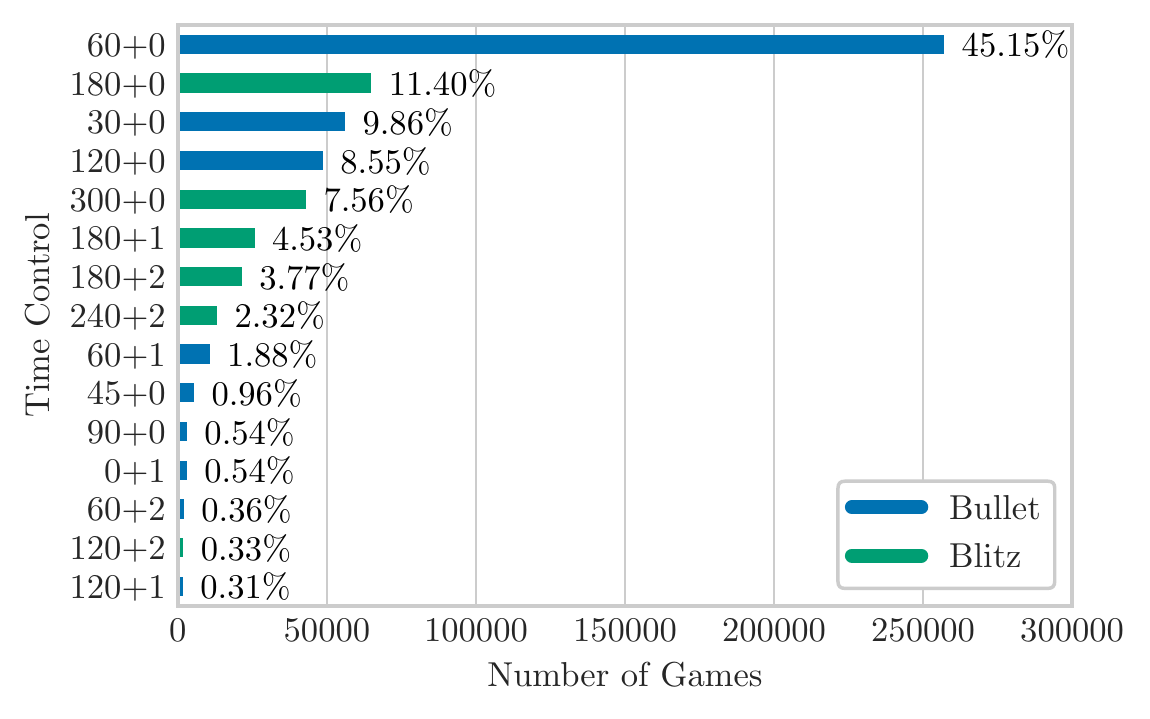}
}
\end{minipage}
\caption{Statistics of the lichess data set}
\label{fig:lichess_elo_tc}
\end{figure}
\addtocounter{subfigure}{-2}
\fi
\section{Training Data}
\label{sec:training_data}
Now that the architectures are in place, let us turn towards the data used for training.
As training data we mainly used \num{569537} human games\cfootnote{\url{https://database.lichess.org/}, accessed 2019-07-30} played by {\small\url{lichess.org}} users from January 2016 to June 2018 
in which both players had an Elo $\geq$ \num{2000}.
The majority of the games used for training have been played by a small group of active players: twenty players participated in $46.03\,\%$ of all
\ifarxiv
games (Figure~\ref{fig:data_players}).
\fi

In crazyhouse, the opening advantage for the first player is even more pronounced and the chance to draw is significantly reduced
\ifarxiv
compared to chess, as also shown in Figure~\ref{fig:game_outcomes}.
\fi
We used all games ending in checkmate, resignation, a draw or time forfeit except aborted games.
All games and moves in our data set are equally weighted in the loss function \eqref{eq:combined_loss}. 

The average Elo rating for a single game is \num{2279.61} and very short time controls are the
\ifarxiv
most popular (see Figure \ref{fig:lichess_elo_tc}).
\fi

One minute games make up $45.15\,\%$ of all games.
As a consequence the chances of blundering is increased and the quality of moves usually declines over a course of a game.
Some games are also won by winning on time in a lost position in which one person gives as many possible checks. 


Additionally, however, we also generated a data set based on \num{121571} \alg{Stockfish} self play games for training a version called \textit{CrazyAraFish}.
For each game position \alg{Stockfish} used one million $\pm$\num{100000} nodes with a hash size of \num{512}\,mb.
The opening book was based on a set of neural network
\ifarxiv
weights\cfootnote{\label{foot:RISEv1}The model used was based on our RISEv1 architecture\\\url{https://github.com/QueensGambit/CrazyAra/wiki/Model-architecture}, accessed 2019-08-19}
\fi
which was trained on the lichess.org database. Consequently the opening book correspondeds to the most popular human crazyhouse openings. The opening suite features \num{1000} unique opening positions where each opening had \num{5.84}\,$\pm$\,\num{2.77} plies. One ply describes a half-move in chess notation.


\section{Supervised Learning to play Crazyhouse}
\label{sec:supervised_training}
\begin{figure}
\centering
\begin{minipage}{.5\textwidth}
  \centering
  \subcaptionbox{Learning rate schedule}{
  \includegraphics[width=\linewidth]{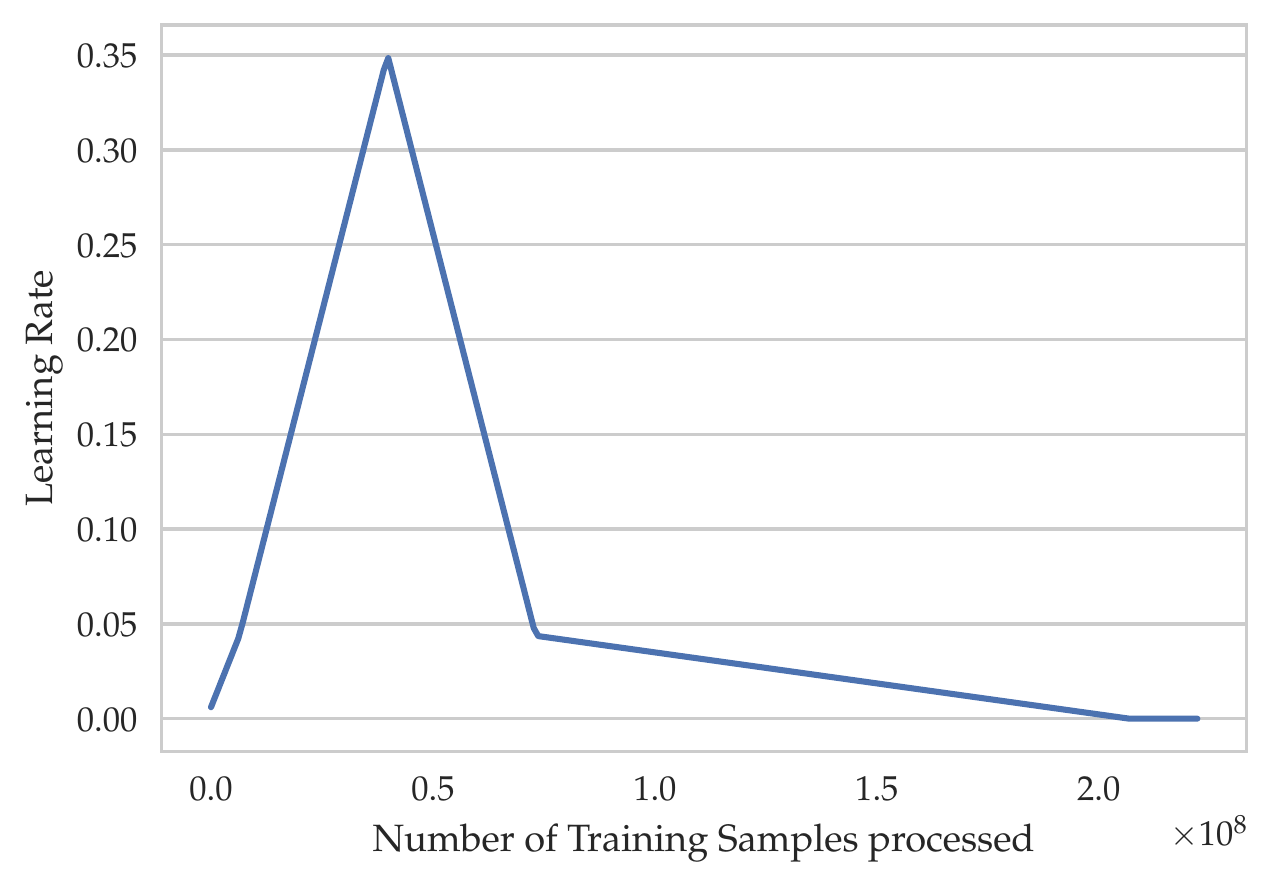}
  }
\label{fig:lr_schedule}
\end{minipage}%
\begin{minipage}{.5\textwidth}
  \centering
    \subcaptionbox{Momentum schedule}{
  \includegraphics[width=\linewidth]{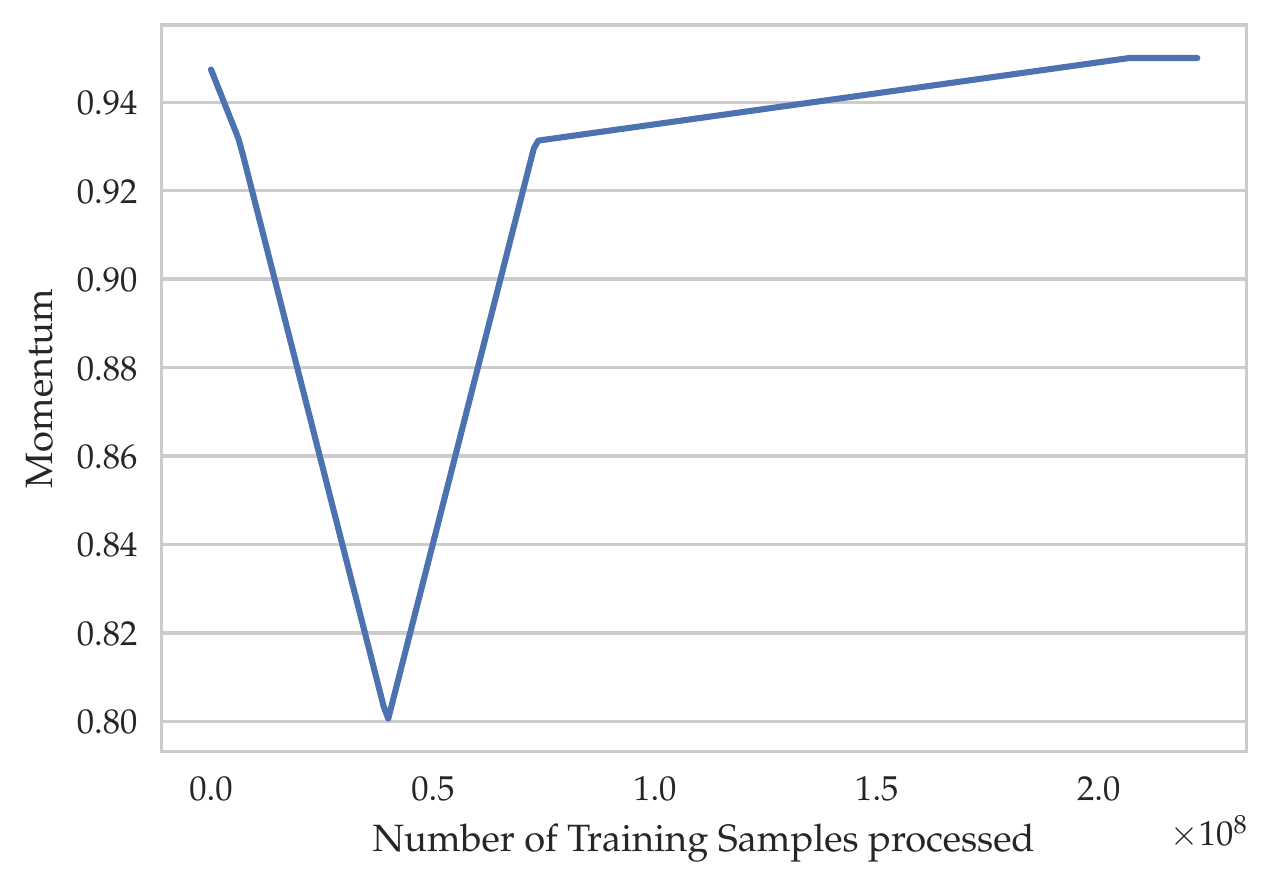}
  }
\end{minipage}
  \caption{Schedules used for modifying the parameters of Nesterov's stochastic gradient descent optimizer}
    \label{fig:schedules}
\end{figure}
\addtocounter{subfigure}{-2}
\begin{figure}
\centering
\begin{minipage}{0.49\textwidth}
\subcaptionbox{Policy Loss}{
  \includegraphics[width=\textwidth]{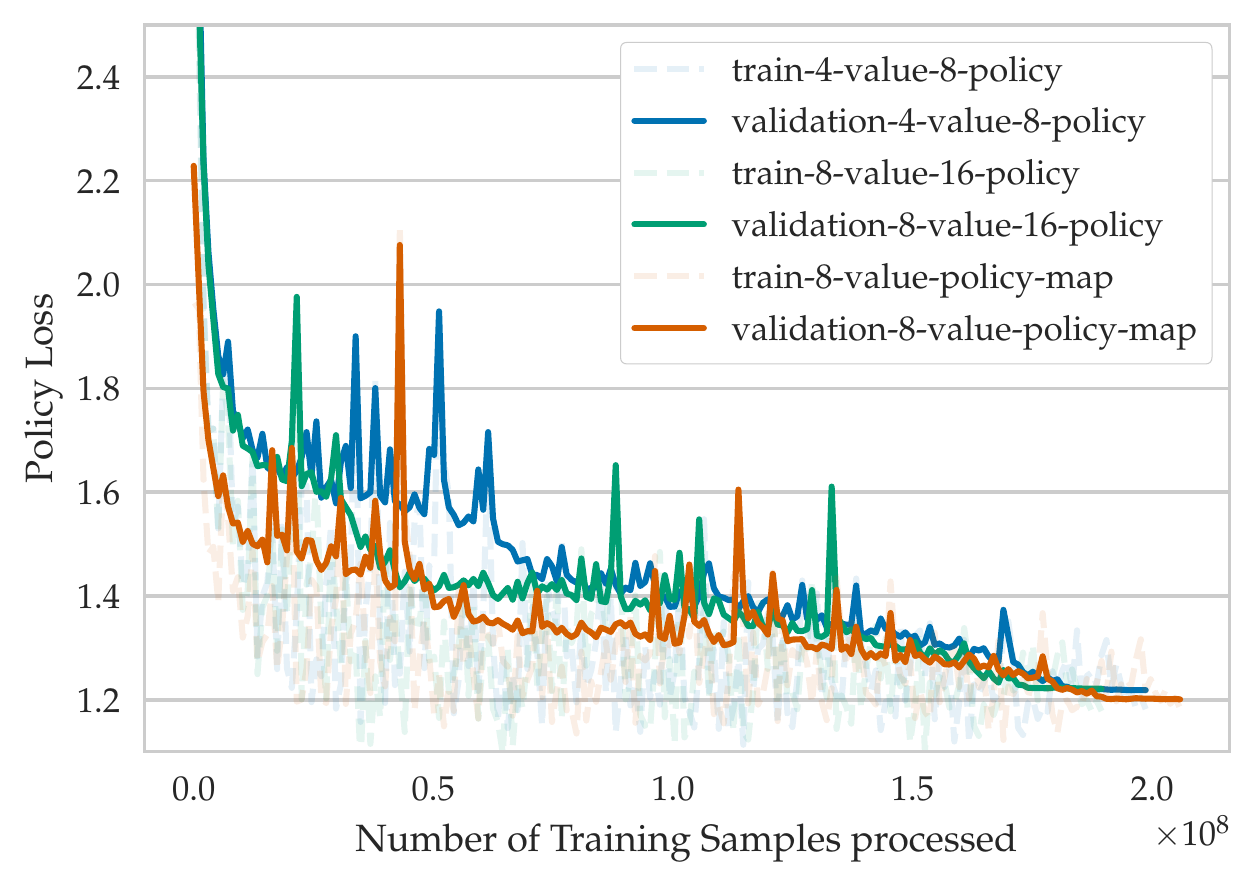}
  }
\end{minipage}
\begin{minipage}{0.49\textwidth}
    \subcaptionbox{Value Loss}{
  \includegraphics[width=\textwidth]{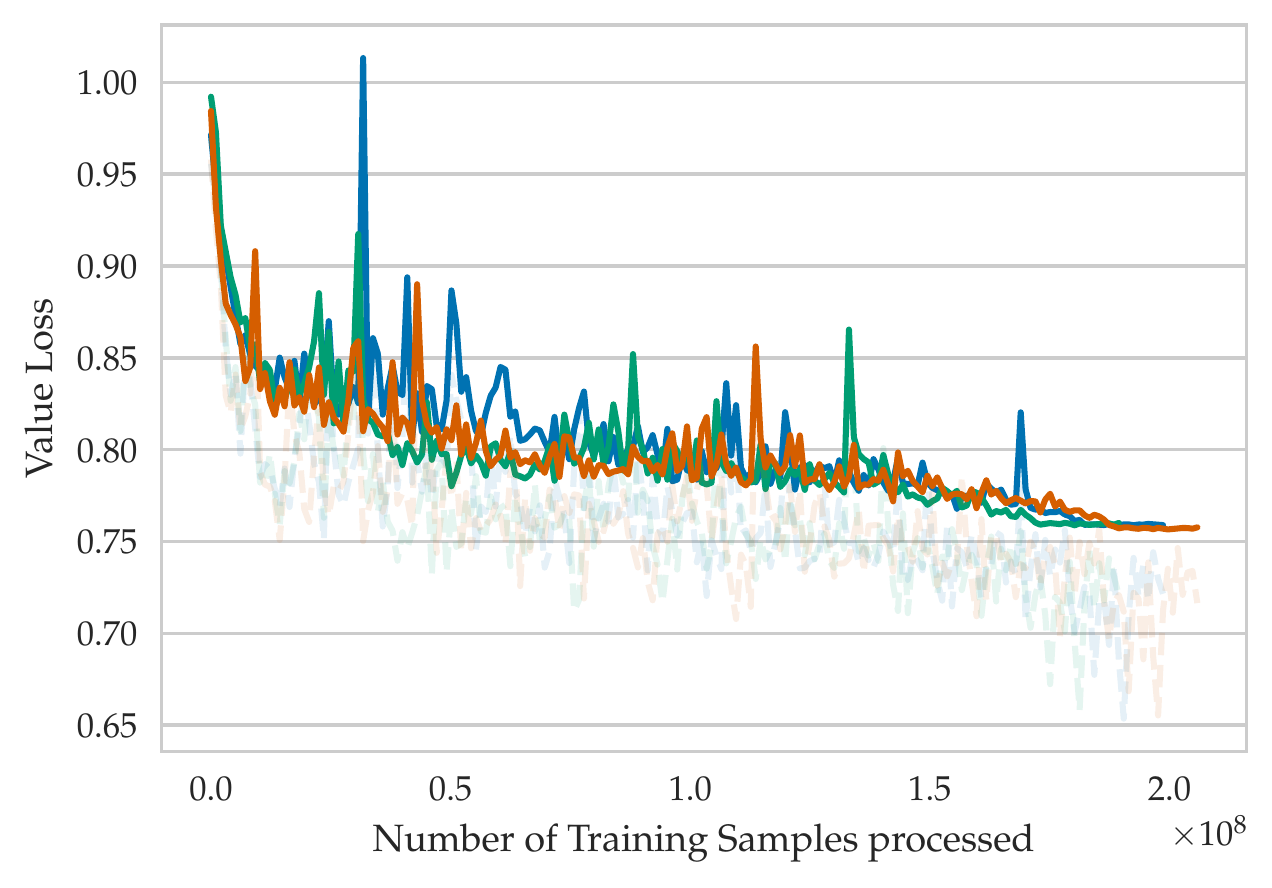}
  }
\end{minipage}
\begin{minipage}{0.49\textwidth}
\subcaptionbox{Policy Accuracy}{
  \includegraphics[width=\textwidth]{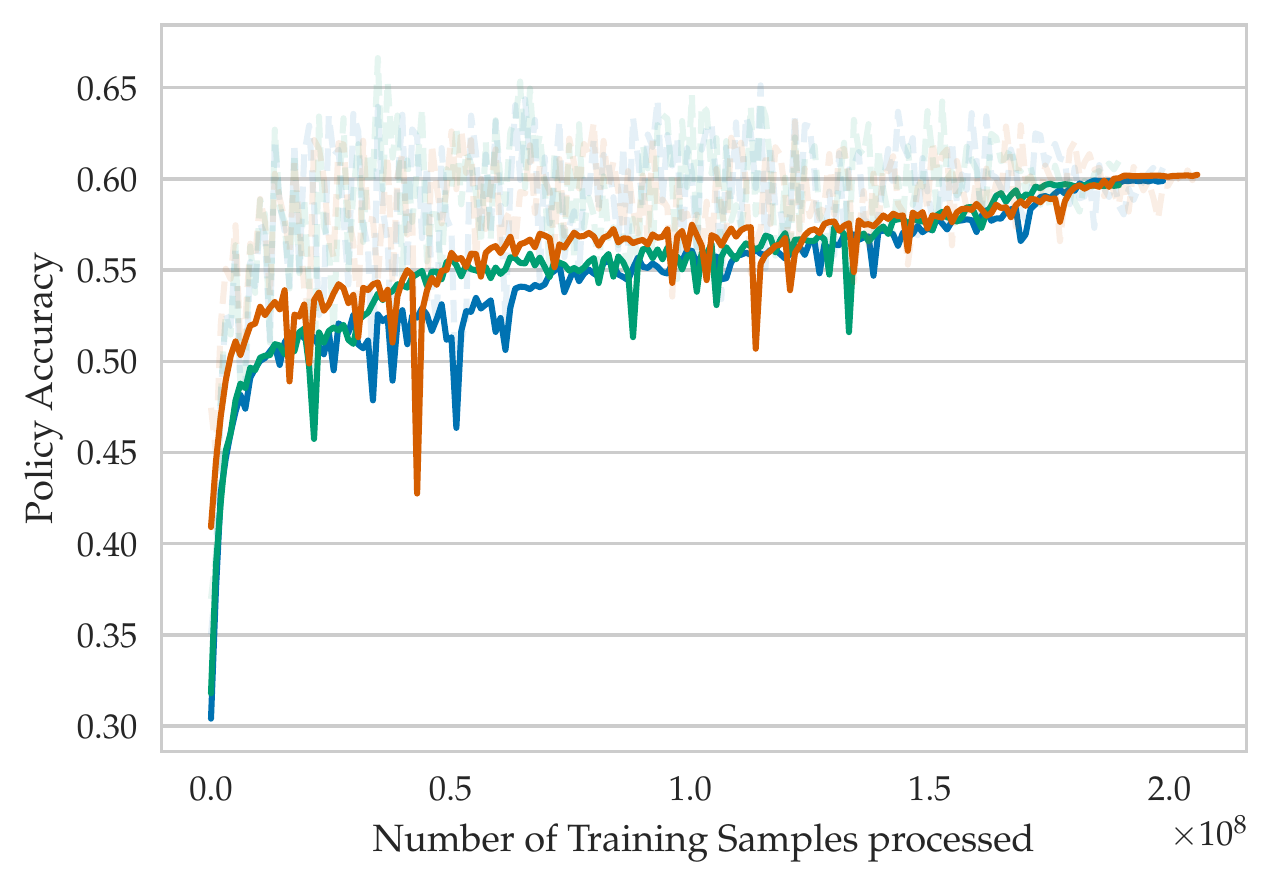}
}
\end{minipage}
\begin{minipage}{0.49\textwidth}
    \subcaptionbox{Value Accuracy Sign}{
  \includegraphics[width=\textwidth]{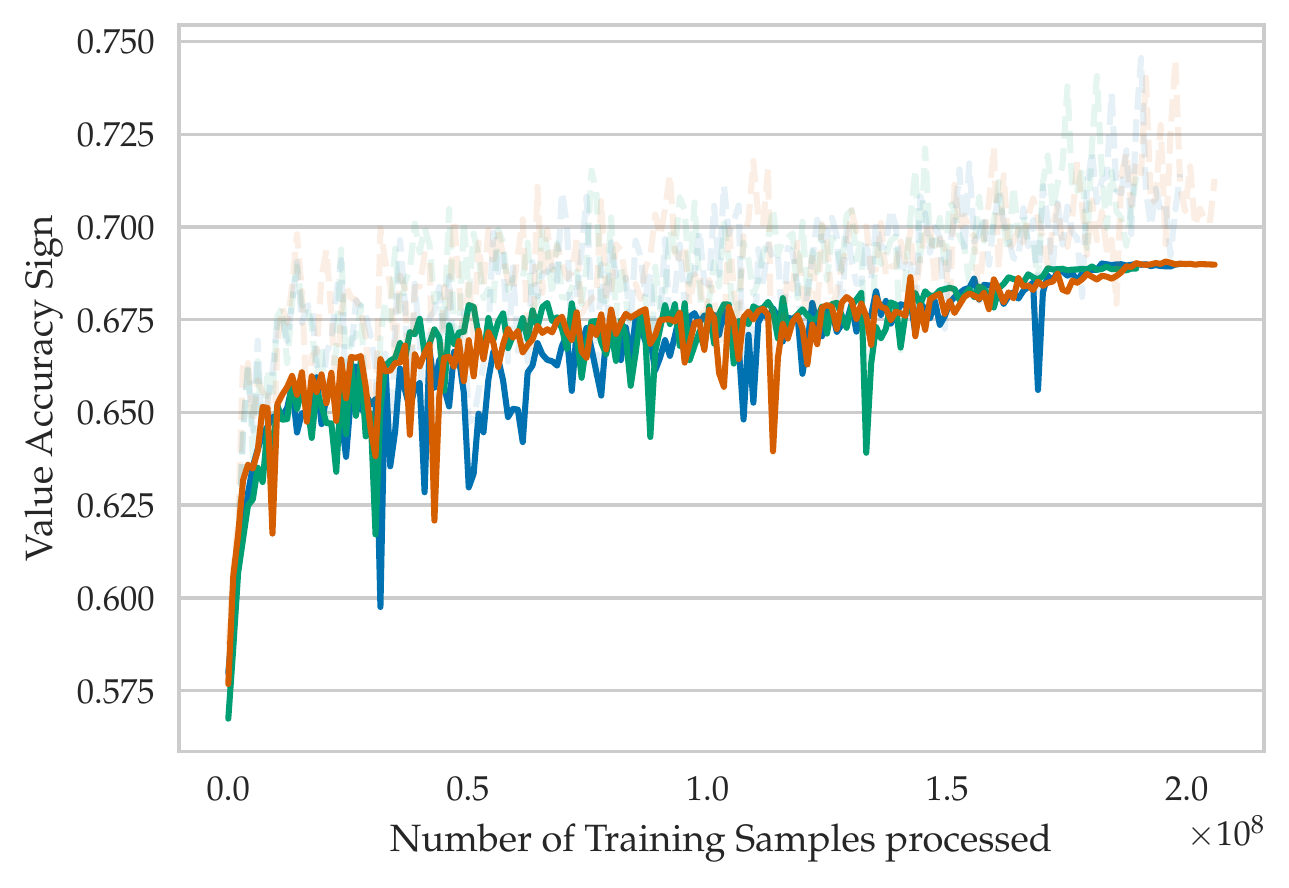}
}
\end{minipage}
\caption{Learning progress of training for seven epochs on the lichess.org crazyhouse data set for different model architectures. \textit{4-value-8-policy} means that four value channels and eight policy channels are used in the respective network heads. \textit{8-value-policy-map} means that eight value channels are used in the value head and the policy
\ifarxiv
is encoded as shown in Table \ref{tab:policy_output}.
\fi
}
\label{fig:learn_plots}
\end{figure}
\addtocounter{subfigure}{-4}
\begin{figure}
\centering
\begin{minipage}{0.49\textwidth}
\subcaptionbox{Validation Policy Loss}{
  \includegraphics[width=\textwidth]{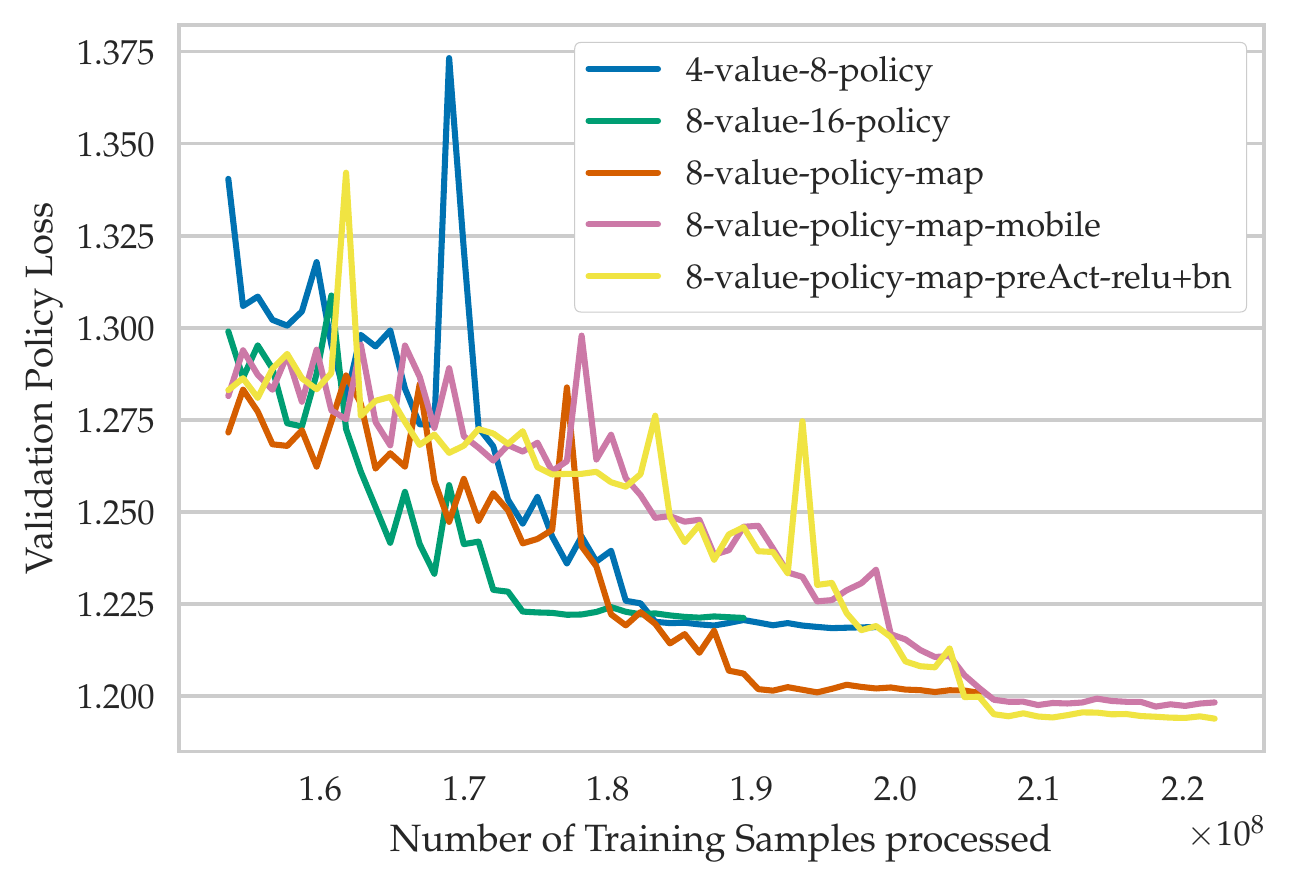}
}
\end{minipage}
\begin{minipage}{0.49\textwidth}
    \subcaptionbox{Validation Value Loss}{
  \includegraphics[width=\textwidth]{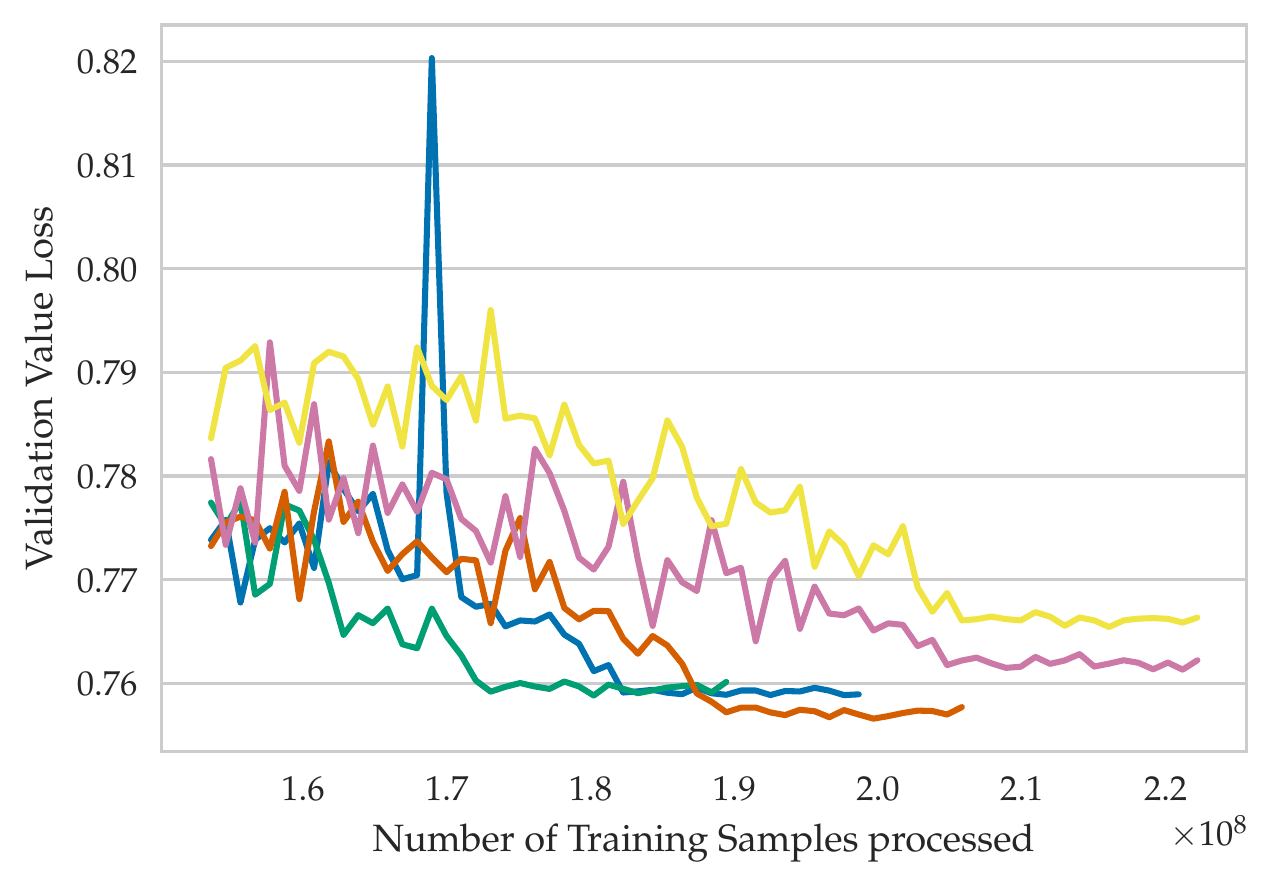}
}
\end{minipage}
\caption{Detailed view on the last iterations of training for different model architectures}
\label{fig:learn_plots}
\end{figure}
\addtocounter{subfigure}{-2}
\ifarxiv
\begin{table}[]
\centering
\caption{\normalsize{Average inference time for a batch of size eight for different network architectures and hardware. Time was measured using \num{300} measurements with \num{30} preceding warm-up iterations. Lower values are better.}}
\label{tab:inference_speed}
\vspace{0.2cm}
\resizebox{\textwidth}{!}{%
\begin{tabular}{lcccccc}
\toprule
\textbf{Hardware} & \textbf{Library} & \begin{tabular}[c]{@{}l@{}}\textbf{4-value-}\\ \textbf{8-policy}\end{tabular} & 
\begin{tabular}[c]{@{}l@{}}\textbf{8-value-}\\ \textbf{16-policy}\end{tabular} & \begin{tabular}[c]{@{}l@{}}\textbf{8-value-}\\ \textbf{policy-map}\end{tabular} & \begin{tabular}[c]{@{}l@{}}\textbf{8-value-}\\
\textbf{policy-map-}\\ \textbf{mobile}\end{tabular} & \begin{tabular}[c]{@{}l@{}}\textbf{8-value-}\\ \textbf{policy-map-}\\ \textbf{preAct}\\ \textbf{-relu+bn}\end{tabular} \\ 
\midrule
\begin{tabular}[c]{@{}l@{}}GeForce GTX 1080\\  Ti/PCIe/SSE2\end{tabular} & \begin{tabular}[c]{@{}c@{}}MXNet-cu10 1.4.1,\\  CUDA 10.0,\\ cuDNN v7.5.1.10\end{tabular} & 4.6000ms & 4.5986ms & 4.8215ms & \textbf{3.3500ms} & 4.9129ms \\ 
\begin{tabular}[c]{@{}l@{}}AMD® Ryzen 7 1700\\  eight-core processor × 16\end{tabular} & MXNet 1.4.1 & 263.5504ms & 271.8701ms & 282.6009ms & \textbf{117.5744ms} & 279.2975ms \\ 
\begin{tabular}[c]{@{}l@{}}Intel®Core™i5-8250U\\  CPU @ 1.60GHz×8\end{tabular} & MXNet 1.4.1 & 169.0708ms & 172.8639ms & 176.7654ms & \textbf{85.0383ms} & 176.1070ms \\ 
\begin{tabular}[c]{@{}l@{}}Intel®Core™i5-8250U\\  CPU @ 1.60GHz×8\end{tabular} & MXNet-mkl 1.4.1 & 93.0978ms & 94.7184ms & 97.1217ms & \textbf{33.7602ms} & 96.2892ms \\ 
\begin{tabular}[c]{@{}l@{}}Intel®Core™i5-8250U\\  CPU @ 1.60GHz×8\end{tabular} & \begin{tabular}[c]{@{}l@{}}MXNet-mkl 1.4.1,\\ graph optimization\end{tabular} & 82.2078ms & 83.2606ms & 86.5579ms & \textbf{28.3759ms} & 94.5224ms\\
\bottomrule
\end{tabular}
}
\end{table}
\fi
\begin{table}[]
\centering
\caption{Performance metrics for different models on the  lichess.org crazyhouse validation set}
\vspace{.2cm}
\label{tab:performance_comp}
\begin{tabular}{lccccc}
\toprule
\textbf{\begin{tabular}[c]{@{}l@{}}Evaluation Metrics on\\ the Validation Set\end{tabular}} & \textbf{\begin{tabular}[c]{@{}c@{}}4-value-\\ 8-policy\end{tabular}} & \textbf{\begin{tabular}[c]{@{}c@{}}8-value-\\ 16-policy\end{tabular}} & \textbf{\begin{tabular}[c]{@{}c@{}}8-value-\\ policy-map\end{tabular}} & \textbf{\begin{tabular}[c]{@{}c@{}}8-value-\\ policy-map-\\ mobile\end{tabular}} & \textbf{\begin{tabular}[c]{@{}c@{}}8-value-\\ policy-map-\\ preAct\\ -relu+bn\end{tabular}} \\
\midrule
Combined Loss & 1.2138 & 1.2166 & 1.1964 & 1.1925 & \textbf{1.1896} \\
Policy Loss & 1.2184 & 1.2212 & 1.2008 & 1.1968 & \textbf{1.1938} \\
Value Loss & 0.7596 & 0.7601 & \textbf{0.7577} & 0.7619 & 0.7663 \\
Policy Accuracy & 0.5986 & 0.5965 & 0.6023 & 0.6032 & \textbf{0.6042} \\
Value Accuracy Sign & 0.6894 & 0.6888 & \textbf{0.6899} & 0.6889 & 0.6868 \\
Mate-In-One-Accuracy & 0.954 & 0.942 & \textbf{0.955} & 0.945 & \textbf{0.955} \\
Mate-In-Top-5-Accuracy & \textbf{0.999} & 0.998 & \textbf{0.999} & 0.998 & \textbf{0.999} \\
Mate-In-One-Value-Loss & 0.0532 & \textbf{0.0474} & 0.0560 & 0.0567 & 0.0624 \\
\bottomrule
\end{tabular}
\end{table}


We trained the resulting models for seven epochs, using a one cycle learning rate schedule combined with a momentum schedule \citep{smith2019super} and updating the weights of the neural network by Stochastic Gradient Descent with Neterov's Momentum (NAG;  \citealt{botev2017nesterov}). As batch-size, we chose the highest value possible on our hardware, which is \num{1024}, and a weight-decay of $10^{-4}$ for regularization. 
The first iterations were treated as a warm-up period followed by a linear long cool-down period.
The maximum and minimum learning rate (Figure~\ref{fig:schedules}) were set to $0.35$ and $0.00001$ respectively and were determined using a learning rate range test \citep{smith2018disciplined}.
The maximum learning rate of $0.35$ is higher than the typical learning rates and acts as an additional source of regularization \citep{smith2019super}.
The momentum schedule was built based on the learning rate schedule (Figure~\ref{fig:schedules}) with a maximum value of $0.95$ and minimum value of $0.85$.
Linear schedules, in contrast to a step-wise learning rate reduction by a fixed factor, greatly reduced the number of training iterations needed and also yielded a higher generalization ability.
The advantages of linear schedules have also been verified on ImageNet by \cite{mishkin2016systematic}.


    
    

As a new metric, we define the ``\textit{value accuracy sign}'' metric, which determines if the network predicts the correct winner given a random game position.
Drawn games are neglected for this metric.
Generally, improving the performance on the value loss is harder because it has a lower influence on the combined loss.
Nonetheless, the gap between validation and train loss is still higher for  value prediction, which confirms that reducing its influence is necessary to avoid overfitting on this small data set.

Furthermore, the value loss decreased over the course of a game.
After training, we evaluated our models using a mate in one data set (Table~\ref{tab:performance_comp}), which was generated using \num{1000} positions from our test set. On average there are \num{115.817} legal moves, from which \num{1.623} lead to direct mate. As our best mate in one accuracy, we achieved $0.955$ with a value loss of $0.047$.
Because the first candidate moves sometimes lead to a mate in \#2 or mate in \#3 instead, we also provide the top five mate in one accuracy where we achieved a value of $0.999$.


Model \model{8-value-policy-map-preAct-relu+bn} and \model{RISEv2-mobile} / \model{8-value-policy-map-mobile}, which both use a 2:1 convolution-ReLU ratio, performed best regarding the combined loss, but worse for the value loss compared to other models.

For training on the \textit{Stockfish} self play data, we used transfer learning of our pre-existing model\,\footnoteref{foot:RISEv1} keeping all model parameters trainable; this model had an initial move prediction accuracy of \num{46.29}\%. We chose a maximum learning rate of $0.01$ and minimum learning rate $0.00001$. After convergence the move prediction increased to \num{56.6}\% on the validation set. We also achieved a significant lower value loss of $0.4407$ on this set.
This is primarily because the games by  \textit{Stockfish} do not contain as many back and forth blunders as human games and the prior win-probability for white is
\ifarxiv
higher (see Figure~\ref{fig:game_outcomes}).
\fi
\section{Configuration of the Monte-Carlo Tree Search}
\label{sec:mcts}

The initial prior policy $\mathbf{p}$ provided by the (deep) neural network $f_\theta$ is now improved using Monte Carlo Tree Search~(MCTS).
\subsection{Default parameter settings}
For the MCTS we used the most recent version of the PUCT algorithm due \cite{silver2017mastering}.
The current game position which is subject to search is the root node of the search tree.
If this position has already been expanded in the previous search then the respective subtree becomes the new search tree.
Next, the statistics of the root node are updated through a continuous sequence of rollouts.
A rollout or simulation is defined as the process of traversing the tree until a new unexplored node is encountered. Then, this node is expanded and evaluated by the (deep) neural network, returning the value prediction and distribution over all possible moves.
This value prediction is backpropagated through the tree for every node that has been visited along its path and the process restarts.
No random rollouts are used and all predictions are provided by a single shared neural network.

Specifically, a node is selected at each step $t$ by taking $a_t = \text{argmax}_a(\text{Q}(s_t, a) + U(s_t, a))$ where $U(s,a) = c_{\text{puct}} P(s,a) {\sqrt{\sum_b N_r(s, b)}}\slash{(1+ N(s,a))}$.
We choose $c_{\text{puct-init}}=2.5$ as our exploration constant which is scaled over the search by 
\begin{equation}
    \label{eq:scale_cpuct}
    c_{\text{puct}}(s) = \log{\frac{\sum_a{N(s,a)}  + c_{\text{puct-base}} + 1}{c_{\text{puct-base}}}} + c_{\text{puct-init}}
\end{equation}{}
with a $c_{\text{puct-base}}$ of \num{19652}.
We apply dirichlet noise with a factor of $25\%$ to the probability distribution $P(s,a)$ of the root node with $\alpha$ of $0.2$ to encourage exploration.
To avoid that all threads traverse the same path during a concurrent rollout we use a virtual loss of $3$ which temporarily reduces the chance of visiting the same current node.

We initialized the Q-values to $-1$ for all newly unvisited nodes. This value caused a lot of misunderstandings and confusion because in the initial \textit{AlphaZero} papers the network was described to return a value range output of $[-1, +1]$, but the initialisation of Q-Values to be zero assumed the values to be in the range of $[0, 1]$.
When treating unvisited nodes as draws, it often led to bad search artifacts and over-exploration, especially when explored positions have negative values because unvisited and low visited nodes have a higher value compared to more visited nodes.
As an intermediate solution when initializing unvisited nodes as draws, we introduced a pruning technique in which nodes are clipped for the rest of the search that (1) did not return a better value prediction than their parent node and (2) have a prior policy visit percentage of below $0.1\,\%$. 
This pruning was used in the matches with JannLee (Section \ref{sec:JannLee}) and caused a major problem in long time control settings:  a key move, which would have defended a certain mate threat, has been clipped.
\subsection{Changes to Monte-Carlo Tree Search}
\label{sec:mcts_changes}
However, to really master the game of crazyhouse at the level of a world champion, we also had to modify standard MTCS in several ways that we now discuss.

\begin{minipage}{\textwidth}
\begin{minipage}{0.49\textwidth}
      \includegraphics[width=\textwidth]{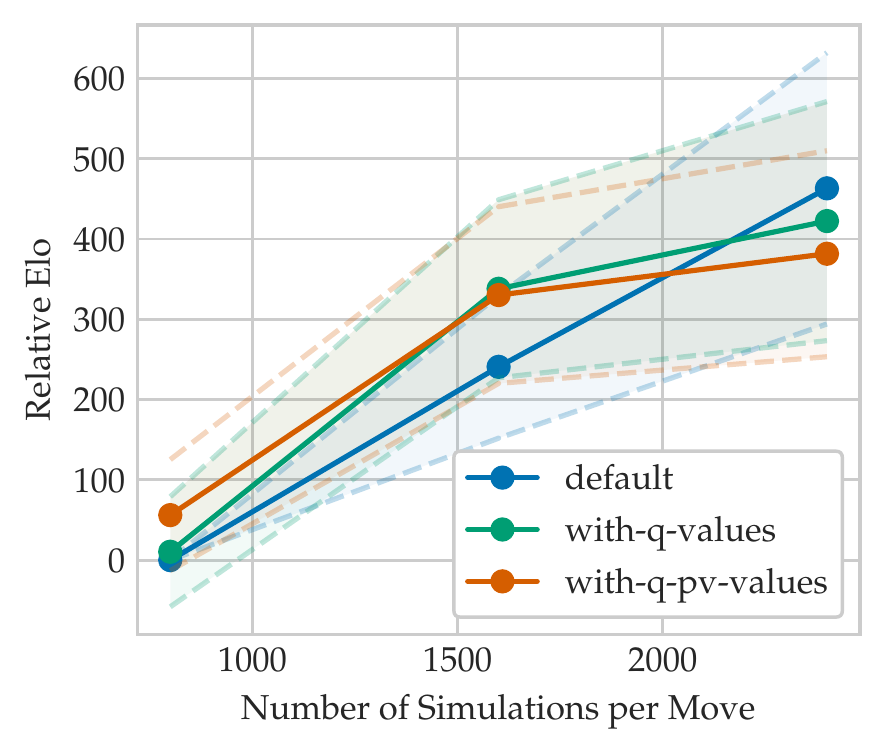}
    \captionof{figure}{Self Elo increase with respect to nodes} 
    \label{fig:elo_increase}
\end{minipage}
\ifarxiv
\hfill
\hspace{0.2cm}
\fi
\begin{minipage}{0.49\textwidth}
\centering
\ifarxiv
\fi
\captionof{table}{
Match results of different MCTS move selection types playing each setting against 800 simulations per move using only the most visited node. Matches were generated with with \textit{CrazyAra 0.4.1} using model \model{4-value-8-policy}. Games start from 50 unique CCVA opening positions with a temperature value of zero for all settings.}
\resizebox{\textwidth}{!}{%
\begin{tabular}{lccccc}
\toprule
\textbf{Type} & \textbf{Simulations} & \textbf{+} & \textbf{-} & \textbf{=} & \textbf{Elo Difference} \\ 
\midrule
default (visits only) & \num{800} & - & - & - & $0$ \\ 
default (visits only) & \num{1600} & $80$ & $20$ & $0$ & $240.82 \pm 88.88$ \\ 
default (visits only) & \num{2400} & $93$ & $6$ & $1$ & $\mathbf{463.16}\pm168.96$ \\ 
visits \& q-values & \num{800} & $51$ & $48$ & $1$ & $10.43\pm68.57$ \\ 
visits \& q-values & \num{1600} & $87$ & $12$ & $1$ & $\mathbf{338.04}\pm110.92$ \\ 
visits \& q-values & \num{2400} & $91$ & $8$ & $0$ & $422.38\pm148.92$ \\ 
visits \& q-pv-values & \num{800} & $57$ & $41$ & $2$ & $\mathbf{56.07}\pm69.14$ \\ 
visits \& q-pv-values & \num{1600} & $87$ & $13$ & $0$ & $330.23\pm110.06$ \\ 
visits \& q-pv-values & \num{2400} & $90$ & $10$ & $0$ & $381.70\pm128.31$ \\
\bottomrule
\label{tab:elo_increase}
\end{tabular}
\normalsize

}
\end{minipage}
%
\end{minipage}

\subsubsection{Integration of Q-Values for final move selection}
For selecting the final move after search, the vanilla version uses an equation only based on the number of visits for each direct child node
\begin{equation}
    \pi ( a | s_0 ) = \frac{N(s_0 , a )^{\frac{1}{\tau}}}{\sum_b N(s_0 , b)^{\frac{1}{\tau}}},
\end{equation}
where $\tau$ is a temperature parameter which controls the exploration. In tournament play against other engines, $\tau$ is set to $0$ which results in choosing the move with the highest number of visits.

We investigated taking information about the Q-values into account, which do not require additional search time and are updated for every rollout. 
Using Q-values for move selection is motivated by the fact that the most frequently visited node is not necessarily the best one, but the quality for each move is in principle described by its Q-value. Usually there is a strong correlation between the most visited move and the move with the highest Q-value. However, in cases when a strong counter reply was only discovered late during search, the Q-value converges quickly, but the respective node still remains at most visits for several samples. 

\cite{silver2018general} also acknowledged that deviating from the most visited node for move selection can yield better results:
\textit{``When we forced AlphaZero to play with greater diversity (by softmax sampling with a temperature of 10.0 among moves for which the value was no more than 1\,\% away
from the best move for the first 30 plies) the winning rate increased from 5.8\,\% to 14\,\%.''}
When naively picking the node with the highest Q-value or directly combining Q-values with number of visits, we encounter the problem that Q-values of nodes with low visit counts can be poorly calibrated and can exhibit an overestimation of its actual value.
Therefore we apply the following procedure. First, we re-scale the Q-values into the range $[0,1]$ and set all Q-values with a visit count $< \text{Q}_\text{thresh} \text{max}_a(N(s_0, a))$ to $0$.
We set $\text{Q}_\text{thresh}$ to $0.33$ and denote these updated Q-values as $\text{Q'}(s_0, a)$.

The Q-values are then integrated as a linear combination combination of visits and Q-values:
\begin{equation}
    \label{eq:q_move_select}
    \pi ( a | s_0 ) = (1 - \text{Q}_\text{factor}) \frac{N(s_0 , a )}{\sum_b N(s_0 , b)} + \text{Q}_\text{factor} \: \text{Q'}(s_0, a).
\end{equation}
\subsubsection{Q-Values with principal variation}
The Q-values can be further adjusted by updating them taking the information of the Principal Variation (PV) for each move candidate into account:
\begin{equation}
    \text{Q}(s_0, a) = \min(\text{Q}(s_0, a), \text{Q}(s_t, x)),
\end{equation}
where $t=5$ and $x$ is the selected move at each depth along the rollout.
The PV-line describes the suggested optimal line of play for both players and is constructed by iteratively choosing the next move according to Equation~\ref{eq:q_move_select} until $s_t$ or a terminal node $s_T$ has been reached.
%
As can be seen in Figure~\ref{fig:elo_increase} and Table~\ref{tab:elo_increase}, the relative increase in Elo\cfootnote{Matches are available in the supplementary materials. The starting opening positions are available at\\\url{https://sites.google.com/site/zhassociation/download}: \texttt{ccva-50start.pgn}, accessed 2019-07-30} is more drastic compared to the scalability of \alg{AlphaZero} for classical chess and more similar to the Elo increase of \alg{AlphaZero} for shogi \citep{silver2017mastering}. We think this is due to the lower chances for draws and higher move complexity of crazyhouse which statistically increases the chance to blunder.
This makes crazyhouse an excellent testing environment for both MCTS and Alpha Beta Search because effects of changes in the search are reinforced while the number of draws and average game length is vastly reduced.

Move selection which also makes use of Q-values outperformed the vanilla version for a node count $< \num{1600}$.
To improve its behaviour also for higher node counts, we applied a dynamic adaption of $\text{Q}_\text{thresh}$, similar to Equation~\ref{eq:scale_cpuct}, and keep $\text{Q}_\text{factor}$ fixed at $0.7$:
\begin{equation}
    \label{eq:scaling-q-thresh}
    \text{Q}_\text{thresh}(s) = \text{Q}_\text{thresh-max} - \exp{ \left( -\frac{\sum_a{N(s,a)}}{\text{Q}_\text{thresh-base}} \right)} \left(\text{Q}_\text{thresh-max} - \text{Q}_\text{thresh-init}\right),
\end{equation}
where $\text{Q}_\text{thresh-init}$ is \num{0.5}, $\text{Q}_\text{thresh-max}$ is \num{0.9} and $\text{Q}_\text{thresh-base}$ is \num{1965}.

\subsubsection{Centi-pawn conversion}
To achieve a better comparability with other crazyhouse engines, we convert our final Q-value of the suggested move after search to the centi-pawn (cp) metric with
\begin{equation}
    \text{cp} = - \frac{v}{\abs{v}} \cdot \log\frac{1-\abs{v}}{\log{\lambda}},
\end{equation}
where $\lambda$ is \num{1.2}.

\subsubsection{Time dependent search}
We also integrate a basic time management system, which does not search moves based on a fixed number of nodes, but on a variable move time $t_\text{current}$. This makes the engine applicable on different hardware and better suited for playing other engines.
There are two main time settings for engine games. In basic one, a predefined constant time is assigned for a given number of moves (e.\,g., \num{40}).
In sudden death mode a total game time is given to both players for all moves. Optionally, the time can be increased by a certain amount on each move, also called increment, in order to reduce the risk of losing on time in long games.

Our time management uses in principle a fixed move time, which depends on the expected game length, remaining time and increment.
We allocate a constant move time and add \num{70}\,\% of the increment time per move.
For sudden death games, we assume a game length of \num{50} and switch to proportional based system at move \num{40}. In the proportional system, we allocate \num{5}\,\% of the available move time and therefore assume that the game will last \num{20} moves longer.
This formula models the empirical distribution of the expected number of moves to the end of the game as presented by \cite{vuvckovic2009time}.


Moreover, we stop the search immediately if there is only a single legal move available (e.\,g. the king is in check with only one escape square) or prematurely at half the allocated time for \textit{easier} positions in order to save time for the rest of the game.
We consider a position to be \textit{easy} if the first candidate move of the prior policy has a likelihood greater than $90\,\%$ and remains the move with the highest Q-value.
We extend the search-time by half of the additional preassigned time for \textit{critical} positions.
A position is considered \textit{critical} if the the Q-value of the current candidate move is $0.1$ smaller than the Q-value of the last played move of the previous state.
The concept of prematurely stopping MCTS search for the game of Go has been investigated by \cite{hendrik2016}.


Last, for games with human players, we also adjust the allocated time per move
\begin{equation}
    t_\text{current} = t_\text{current} + t_\text{factor} t_\text{current},
\end{equation}
where $t_\text{factor} \sim [-0.1, +0.1]$ to increase playing variety.

\subsubsection{Transposition table}
\label{sec:trans_table}
Furthermore, we introduce a transposition table, which stores a pointer to all unique nodes in the tree.
Since the number of nodes is several magnitudes lower than for alpha beta engines, its memory size is negligible.
Transposition tables are used in most modern chess engines as a look-up table to store and reuse the value evaluation for already explored nodes. In the case of MCTS, we can reuse already existing policy prediction vectors as well as value evaluations. Additionally, one might copy the move generation instead of recomputing it.
Transpositions occur frequently during search in chess games and can increase the evaluated nodes per second by a factor of two or more on certain position.
Because our input representation depends on the movement counter as well as the no-progress counter, we only take transpositions into account where these counters share the same value.

\subsubsection{U-Value exploration factor}
\label{sec:u_value_explore}
We also notice that for certain positions, if a node was found with a Q-value $\gg 0$, then the node exploration of unvisited nodes is sharply reduced.
This is because all nodes are initialized with a Q-value of $-1$ and represent losing positions. We make the division factor for calculating the U-values parameterizable:
\begin{equation}
    U(s,a) = c_{\text{puct}} P(s,a) \frac{\sqrt{\sum_b N(s, b)}}{u_{\text{divisor}}+ N(s,a)}.
\end{equation}{}
This increases the chance of exploring unvisited nodes at least once, according to the principle 
\begin{quote}
\textit{``When you see a good move, look for a better one''}--- Emanuel Lasker.
\end{quote}
A $u_{\text{divisor}} < 1$ increases the need of exploring unvisited nodes and can help to reduce the chance of missing key moves, but comes at the cost of losing search depth.
To avoid over-exploration at nodes with a low visits count, we reduce  $u_{\text{divisor}}$ over time, similar to Equation~\ref{eq:scaling-q-thresh}:

\begin{equation}
    \label{eq:scaling-q-thresh}
    u_{\text{divisor}}(s) = u_\text{min} - \exp{ \left( -\frac{\sum_a{N(s,a)}}{u_\text{base}} \right)} \left(u_\text{min} - u_\text{init}\right),
\end{equation}
where $u_\text{min}$ is \num{0.25}, $u_\text{init}$ is \num{1} and $u_\text{base}$ is \num{1965}.

\subsubsection{Integration of domain knowledge}
\label{sec:domain_knowledge}
Checks are usually important moves in crazyhouse and it can have detrimental effects if these are missed during search.
To ensure that checks have been sufficiently explored, we add the option to enhance the prior probability for all checking moves $P_{check\text{'}}(s,a) < check_\text{tresh}$ by
\begin{equation}
    P_{check\text{'}}(s,a) = P_{check\text{'}}(s,a) + check_\text{factor} \max_a({P(s,a)}),
\end{equation}
where we set $check_\text{tresh}$ to $0.1$, $check_\text{factor}$ to $0.5$ and renormalize the distribution afterwards.
This modification has the following motivations: the preassigned order for checking moves should be preserved, but checking moves with a low probability are preferred over low confidence non-checking moves.
The top-candidate non-checking moves should remain as the move with the highest prior probability.

Including this step might not be beneficial as soon as our network $f_\theta$ reaches a certain level of play, but it provides guarantees and was found to greatly increase the efficiency in positions where a forced mate is possible or in which the opponent is threatening a forced mating sequence.

Further, we disable any exploration for a particular node as soon as a move was found which leads to a winning terminal node.
A direct checkmate is a case which is known to be the best move for all available moves, so additional exploration is unneeded an can only distort the value evaluation.
\section{Discussion}
\label{sec:discussion}
Before moving on to our empirical evaluation, let us discuss the pros and cons of the techniques used in \alg{CrazyAra} compared to alternatives as well as provide an illustrative example for our MCTS approach.

\subsection{The pros and cons of MCTS for crazyhouse}
As mentioned, alpha-beta engines are strongest at open tactical positions. This holds particularly true for finding forced sequences such as a long series of checks.
For example, it is common for \textit{Stockfish} to find a forced mate of 20 or up to 40 half-moves in under five seconds in crazyhouse games given sufficient computing power.

In contrast, MCTS shows the opposite behavior and shares similar strength and weakness when compared to human play. It exhibits a significantly lower number of node evaluation and is generally inferior in solving tactical positions quickly if the tactic does not follow its current pattern recognition.
On the other hand, it is often better at executing long term strategies and sacrifices because its search is guided by a non-linear policy and is able to explore promising paths more deeply.
Alpha-beta engines commonly purely rely on handcrafted linear value evaluations and there is no flow of information in a proposed principal evaluation.
The non-linear, more costly value evaluation function can also allow it to vary between small nuances in similar positions.
Additionally, (deep) neural networks are able to implicitly build an opening book based on supervised training data or self-play, whereas traditional alpha-beta engines need to search a repeating position from scratch or have to store evaluations in a look-up table which is linearly increasing in size.

In crazyhouse the importance of tactics is increased compared to classical chess and generally when a certain tactic has been missed, the game is essentially decided in engine vs engine games.
Stronger strategic skills result in long grinding games, building up a small advantage move by move, and usually take longer to execute.

MCTS is naturally parallelizable and can make use of every single node evaluation while minimax-search needs to explore a full depth to update its evaluation and is harder to parallelize. Also the computational effort increases exponentially for each depth in minimax-serach.
Besides that, MCTS returns a distribution over all legal moves which can be used for sampling like in reinforcement learning. 
As a downside, this version of MCTS highly depends on the probability distribution $a \sim P(s, a)$ of the network $f_\theta$ which can have certain blind spots of missing critical moves.
Minimax search with a alpha-beta pruning is known to have the problem of the horizon effect where a miss-leading value evaluation is given because certain tactics have not been resolved. To counteract this, minimax-based algorithms employ quiescence search to explore certain moves at greater depth \citep{QuiescenceSearch}.
For MCTS search taking the average over all future value evaluation for all expansions of a node can be misleading if there is only a single winning line and in the worst case to divergence.



%
%

\subsection{Exemplary MCTS search}
\begin{figure}
\centering
\centering
\begin{minipage}{.49\textwidth}
\centering
\input{chessboards/tactic_pos.tex}
\label{fig:tactical_board}
\end{minipage}%
\hfill
\begin{minipage}{0.49\textwidth}
  \centering
  \subcaptionbox{Sorted policy distribution of the first ten candidate moves}{
    \includegraphics[width=\linewidth]{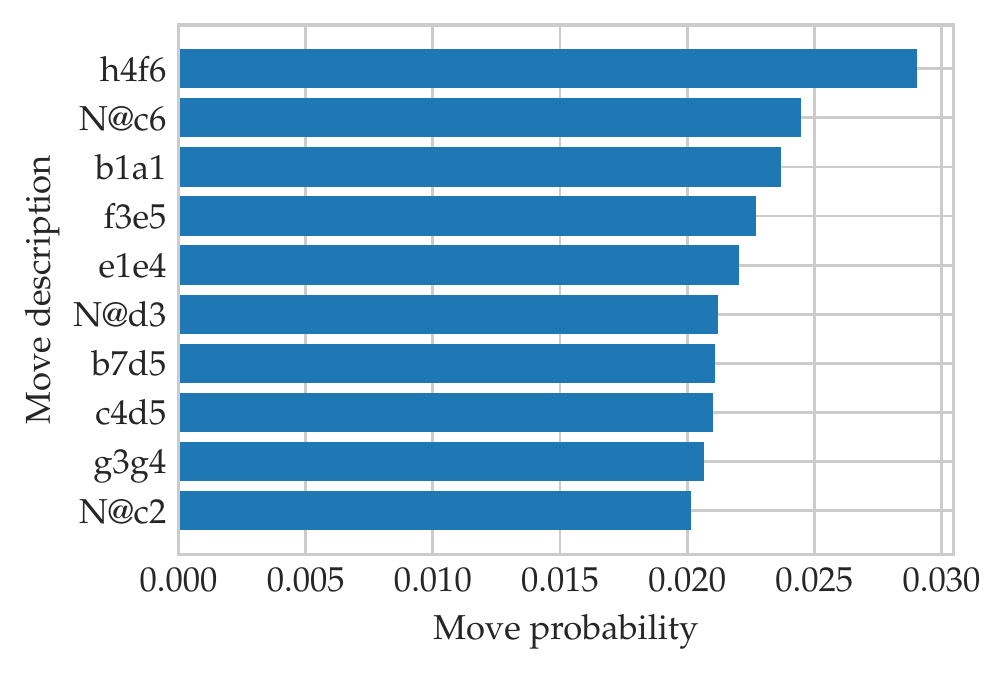}
  }
  \label{fig:candidate_moves}
\end{minipage}
\begin{minipage}{0.49\textwidth}
\centering
\subcaptionbox{Simulation progression}{
  \includegraphics[width=\textwidth]{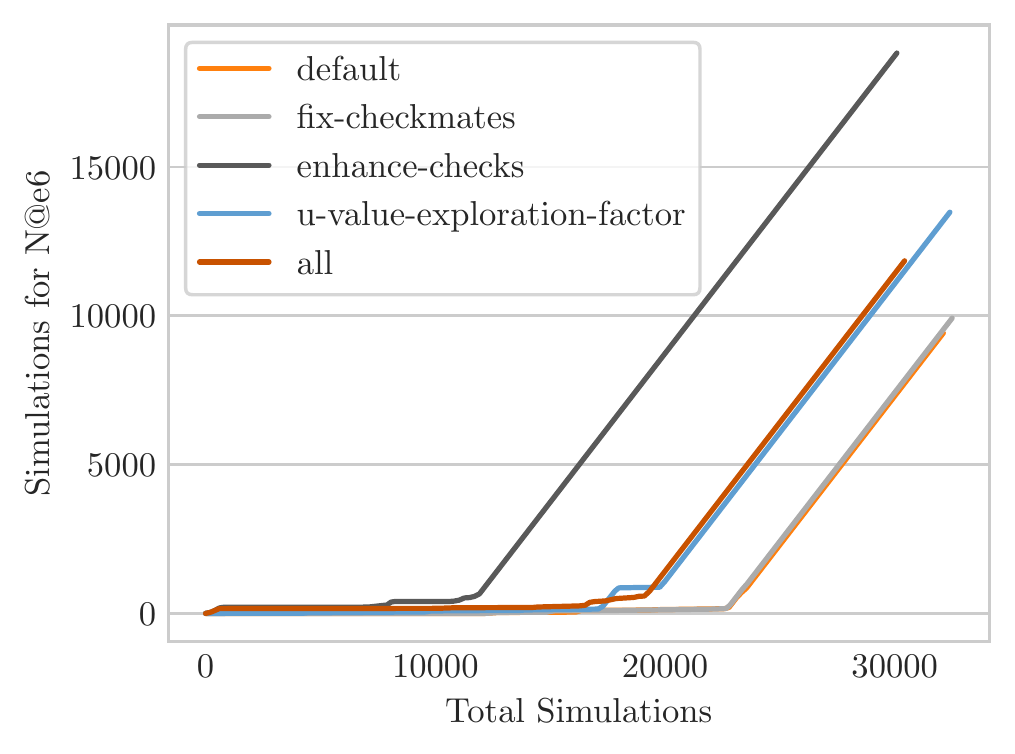}
}
\end{minipage}
\hfill
\begin{minipage}{0.49\textwidth}
\centering
\subcaptionbox{Q-value progression}{
  \includegraphics[width=\textwidth]{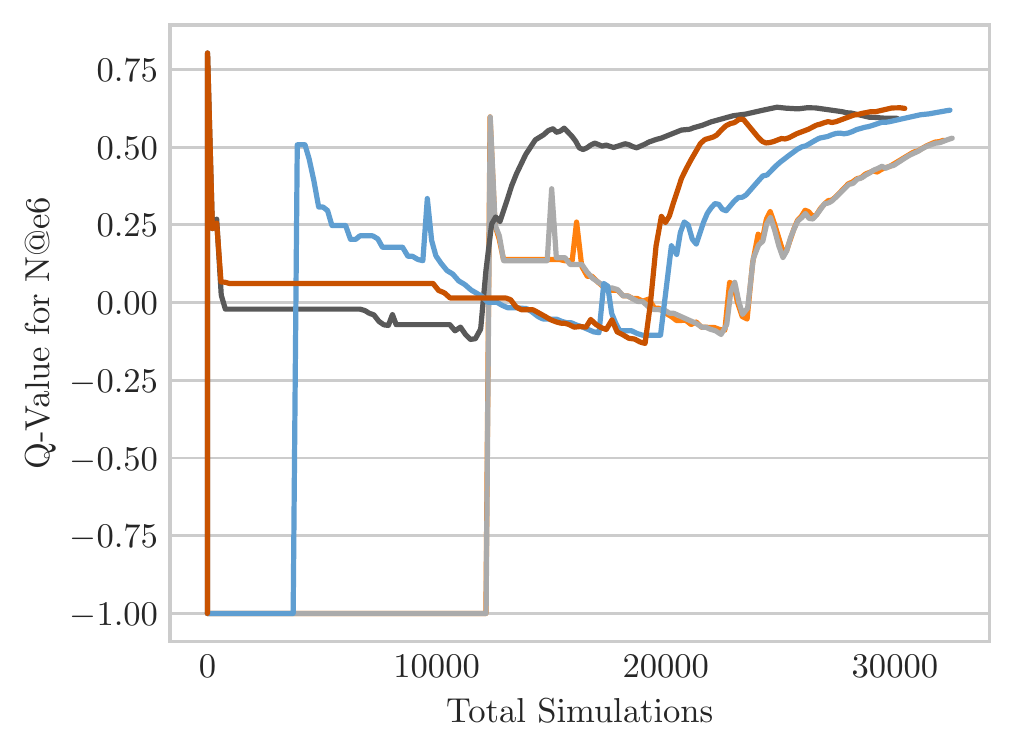}
}
\end{minipage}
%
\caption{Search tree statistics over time for the Move \ctexttt{N@e6} using different MCTS modifications}
\label{fig:mcts_progression}
\end{figure}

Figure \ref{fig:mcts_progression} shows a possible board position in a crazyhouse game\cfootnote{The fen for this position is: \cstexttt{3k2r1/pBpr1p1p/Pp3p1B/3p4/2PPn2B/5NPp/q4PpP/1R1QR1K1/NNbp w - - 1 23}} in which $P(s,a)$ misses the correct move in its first ten candidate moves.
The white player has a high material advantage, but black is threatening direct mate by \ctexttt{Qxf2\#} as well as \ctexttt{Qxb1} followed by \ctexttt{R@h1\#}.
White has to find the following sequence of moves
\ctexttt{24.\;N@e6!!\;fxe6!\ 25.\;Bxf6!\;Ke8\ 26.\;P@f7!!\;Kxf7\ 27.\;N@g5!\;Ke8\ 28.\;Nxh3!}
to defend both mate threats and to keep a high winning advantage advantage. Intermediate captures such as \ctexttt{Rxe4} or \ctexttt{N@c6} or a different move ordering are also losing.
There are \num{73} moves available and $P(s,a)$ of the used model assigns a low softmax-activation of \num{2.69e-05} for the move \ctexttt{N@e6} resulting as the $53^\text{\,rd}$ candidate move.
Figure~\ref{fig:mcts_progression} shows the progression of the number of visits and Q-value for the move \ctexttt{N@e6} using our introduced MCTS adaptations.
To disable the effect of randomness and fluctuations due to thread scheduling, we make the search fully deterministic for this experiment: we set the number of threads to one with a batch-size of eight and replace the Dirichlet noise distribution by a constant distribution which uniformly increases the prior probability for all moves in the root node.

The convergence for the node visits behaves linearly as soon as the move \ctexttt{N@e6} emerged as the most promising move.
If the move remains unexplored its corresponding Q-value stays at a value of $-1$. After the move is visited for only a few samples the event is associated with a spike. These high initial Q-value degrade over time because simulations chose the wrong follow-up lines and consequently back-propagate a miss-leading value evaluation. In cases where the move \ctexttt{N@e6} remains unvisited for a period, the Q-value behaves flat.
As soon as all of white only moves have been found in response to different attempts by black, the Q-value quickly recovers and converges to a value of approximately $+0.6$.

When comparing our changes to MCTS, we notice the following:
in the \textit{default} MCTS version the move \ctexttt{N@e6} remains unvisited for more than \num{10000} simulations due to a very low prior probability.
In the case when exploration is disabled as soon as a mate in one has been found (\textit{fix-checkmates}) it requires slightly fewer samples.
If the prior probability for all checking moves is uniformly increased (\textit{enhance-checks}, Section~\ref{sec:domain_knowledge}) the convergence is overall the fastest.
For the parameterized \textit{u-value-exploration-factor}, the initial visit and convergence of the Q-value is pulled forward.
The last setting (\textit{all}) combines all of our pre-mentioned search adaptions. The initial node exploration occurs similarly to \textit{enhance-checks}, and after the  Q-value degraded the node remains unexplored for ~\num{8000} simulations.
However, after the correct lines for white have been found for the first few depths, the Q-value convergences as quickly as \textit{enhance-checks}.

\section{Experimental Evaluation}
\label{sec:results}
Our intention here is to evaluate the playing strength of \alg{CrazyAra}. To this end, we let \alg{CrazyAra} play matches against other crazyhouse engines as well as the human player Justin Tan, the 2017 crazyhouse world champion. 

\subsection{Matches with human professional players} 
\label{sec:JannLee}


Over the course of its development, \alg{CrazyAra} was hosted several times on lichess.org playing among others the strongest human crazyhouse players from whom it has learnt to play the game.
On December 21st 2018 at 18:00 GMT \textit{CrazyAra 0.3.1} played multiple world crazyhouse champion Justin Tan, also known as \textit{LM JannLee} and won four of five informal games\cfootnote{All games can be found in the supplementary section. The match has been streamed live and commented by Justin Tan.} in the
\ifarxiv
time control of 5\,min + 15\,s (Table \ref{tab:jannlee_results}).
\fi
For the settings, we used a temperature value $\tau$ of $0.07$ for the first four moves, thinking during opponents turn called ``Ponder'' was disabled and we achieved a NPS of \num{250}.
Our recommended changes to MCTS in Sections~\ref{sec:u_value_explore} and \ref{sec:domain_knowledge} had not been integrated in this version.
\ifarxiv
\begin{table}[]
    \centering
    \caption{\normalsize{Match results between Justin Tan and \textit{CrazyAra 0.3.1}}}
    \label{tab:jannlee_results}
    \vspace{0.2cm}
\begin{tabular}{lllccc}
\toprule
\textbf{White} & \textbf{Black} & \textbf{Opening} & \textbf{Time Control} & \textbf{Ply Count} & \textbf{Result} \\
\midrule
CrazyAra 0.3.0 & LM JannLee & C00 French Defense & 300+15 & 107 & 1-0 \\
LM JannLee & CrazyAra 0.3.0 & B00 Nimzowitsch Defense & 300+15 & 73 & 1-0 \\
CrazyAra 0.3.0 & LM JannLee & C20 King's Pawn Game & 300+15 & 51 & 1-0 \\
LM JannLee & CrazyAra 0.3.0 & C00 French Defense & 300+15 & 46 & 0-1 \\
CrazyAra 0.3.0 & LM JannLee & B02 Alekhine Defense & 300+15 & 127 & 1-0 \\
\bottomrule
\end{tabular}
\end{table}

\fi

\subsection{Strength evaluation with other crazyhouse engines}
\label{sec:strength_eval}
We also evaluate the playing strength of \textit{CrazyAra 0.6.0} using the network architecture \model{8-value-policy-map-mobile} (see Table \ref{tab:risev2_mobile}) 
on an Intel® Core™ i5-8250U CPU @ 1.60GHz × 8, Ubuntu 18.04.2 LTS against all participants of the second CCVA Computer Championships \citep{mosca_2nd_2017} 
or their respective updated version.

All engines including \textit{CrazyAra} were run on the same hardware in a time control of 15\,min + 5\,s per move.
The number of threads was set to $8$ and the hash size to \num{1024}\,mb, if the engine provides an option for it.
If only a Windows executable is available for an engine, we made it compatible with the help of Wine \citep{julliard_winehq_1994}, which reduces the original NPS of an engine by about $10\,\%$.
``Ponder'' was turned off and we also allowed other engines to use opening books or position learning.
As the deep learning framework for \textit{CrazyAra}, we used MXNet 1.4.1 \citep{chen2015mxnet} with the Intel-MKL backend \citep{wang2014intel} and also enabled the network subgraph optimization by Intel.
Arithmetic, vectorized operations in the MCTS search were formulated in the Blaze 3.6 library \citep{iglberger2012expression, iglberger2012high} and move generation routines were integrated from \alg{multi-variant Stockfish.}\cfootnote{\url{https://github.com/ddugovic/Stockfish}, accessed 2019-07-30}

To get a reference point, DeepMind's \textit{AlphaZero} ``\textit{used 44 processor cores and four of Google’s first-generation TPUs generating 
about \num{60000} positions per second in chess compared to Stockfish’s roughly 60 million}'' yielding a ratio of \num{1}:\num{1000}.
\textit{AlphaZero} started to outperform the search efficiency of \textit{Stockfish~8} in their setup after \num{30000} node evaluations \citep{silver2017mastering}.
We achieved \num{330} nodes per second (NPS) for \textit{CrazyAra 0.6.0} on the aforementioned CPU, resulting in approximate \num{7260} nodes per move in this time control compared to an average NPS between 1 million nodes for most alpha-beta engines and \num{4.6} million for multi-variant \textit{Stockfish 10}, resulting in a ratio between \num{1}:\num{3000} and \num{1}:\num{14000}.
Despite this large gap in number of evaluated nodes, \textit{CrazyAra} often achieved a higher depth compared to other engines.
This is partly due to 
the forcing nature of crazyhouse in which a majority of the moves are losing outright and filtered out by the neural network.
However, we use the term depth for \CrazyAra as the length of the explored principal variation after the search, and
alpha-beta engines usually explore the full tree or much larger parts of it.
The depth and centipawn evaluation of the current board position is denoted as \ctexttt{\{<centipawn>/<depth>\ <time spent>\}} after each move where a positive centipawn value describes an advantage in the view of the respective engine.

We played ten matches with each engine starting from
\ifarxiv
the following
\fi
five common crazyhouse opening
\ifarxiv
positions, see also Figure \ref{fig:openings_cpu}.
\fi
Each position was played twice, one in which \CrazyAra played the white and one in which it played the black pieces. The matches were played without a resign threshold and always ended in a mate position or a draw. 
We enabled all of our proposed changes for the MCTS search (Section~\ref{sec:mcts_changes})
and used a temperature value of zero to make the moves for \CrazyAra relatively deterministic and to give all other engines the same chances.

The results of the matches (see Table \ref{tab:strength_eval_cpu}) demonstrate that
\alg{CrazyAra 0.6.0} clearly won against twelve of the thirteen participants,\cfootnote{\alg{PyChess'} rating of rating of \num{1566.25} corresponds to version 0.12.4, \alg{Immortal's} rating of \num{2997.33} corresponds to version 3.04, \alg{Stockfish's} rating of \num{3946.06}  corresponds to version 2017-09-23 using a single thread instead of eight} with either ten or nine wins out of ten matches.
All matches with the respective engine evaluations and their depth on each move are available in the supplementary section.\cfootnote{For more information about the engines and their corresponding authors please refer to the \textit{Crazyhouse Alpha List} maintained by Simon Guenther \url{http://rwbc-chess.de/chronology.htm}, accessed 2019-07-30}
\ifarxiv
\ifarxiv
\def\boardwidth{0.199}
\else
\def\boardwidth{0.195}
\fi
\begin{figure}
\centering
\begin{minipage}{\boardwidth\textwidth}
\resizebox{\textwidth}{!}{
\chessboard[setfen=rnbqkbnr/pppp1ppp/8/4p3/4P3/8/PPPP1PPP/RNBQKBNR/ w KQkq - 2 2,]
}
\caption*{\ccsmall{C20 King's Pawn Game}}
\end{minipage}
\begin{minipage}{\boardwidth\textwidth}
\resizebox{\textwidth}{!}{
\chessboard[setfen=rnbqkbnr/pppp1ppp/4p3/8/4P3/8/PPPP1PPP/RNBQKBNR/ w KQkq - 2 2,]
}
\caption*{\ccsmall{C00 French Defense}}
\end{minipage}%
\begin{minipage}{\boardwidth\textwidth}
\resizebox{\textwidth}{!}{
\chessboard[setfen=rnbqkb1r/pppppppp/5n2/8/4P3/8/PPPP1PPP/RNBQKBNR/ w KQkq - 2 2,]
}
\caption*{\ccsmall{B02 Alekhine Defense}}
\end{minipage}%
\begin{minipage}{\boardwidth\textwidth}
\resizebox{\textwidth}{!}{
\chessboard[setfen=rnbqkbnr/ppp1pppp/3p4/8/4P3/8/PPPP1PPP/RNBQKBNR/ w KQkq - 2 2,]
}
\caption*{\ccsmall{B07 Pirc Defense}}
\end{minipage}%
\begin{minipage}{\boardwidth\textwidth}
\vspace{0.37cm}
\resizebox{\textwidth}{!}{
\chessboard[setfen=rnbqkbnr/ppp1pppp/8/3p4/3P4/8/PPP1PPPP/RNBQKBNR/ w KQkq - 2 2,]
}
\caption*{\ccsmall{D00 Queen's Pawn\\Game}}
\end{minipage}%
\caption{Chosen opening positions validating the playing strength of \alg{CrazyAra 0.6.0} on CPU}
\label{fig:openings_cpu}
\end{figure}
\begin{figure}
\centering
\begin{minipage}{\boardwidth\textwidth}
\resizebox{\textwidth}{!}{
\chessboard[setfen=r1bqkbnr/pppp1ppp/2n5/1B2p3/4P3/5N2/PPPP1PPP/RNBQK2R b KQkq - 3 3,]
}
\caption*{\ccsmall{C60 Ruy Lopez}}
\end{minipage}
\begin{minipage}{\boardwidth\textwidth}
\resizebox{\textwidth}{!}{
\chessboard[setfen=r1bq1rk1/ppppbppp/2n2n2/4p1B1/2B1P3/3P1N2/PPP2PPP/RN1Q1RK1 w - - 3 7,]
}
\caption*{\ccsmall{C50 Giuoco Piano}}
\end{minipage}%
\begin{minipage}{\boardwidth\textwidth}
\resizebox{\textwidth}{!}{
\chessboard[
pgfstyle={text},
text=\;\qquad\WhiteKnightOnWhite,
markfield={h8},
text=\;\qquad\BlackPawnOnWhite, 
markfield={h7},
setfen=r1bqkb1r/ppp2ppp/2n2p2/3p4/3P4/8/PPP2PPP/RNBQKBNR/Np w KQkq - 2 5,]
}
\caption*{\ccsmall{B02 Crosky Gambit 
}}
\end{minipage}%
\begin{minipage}{\boardwidth\textwidth}
\vspace{0.37cm}
\resizebox{\textwidth}{!}{
\chessboard[setfen=rn1qkbnr/ppp1pppp/8/3p1b2/3P4/4P3/PPP2PPP/RNBQKBNR w KQkq - 1 3,]
}
\caption*{\ccsmall{D00 Queen's Pawn\\Game} 
}
\end{minipage}%
\begin{minipage}{\boardwidth\textwidth}
\resizebox{\textwidth}{!}{
\chessboard[setfen=rnbqkb1r/ppp1pppp/5n2/3p4/3P4/6P1/PPP1PPBP/RNBQK1NR b KQkq - 2 3,]
}
\caption*{\ccsmall{A00 Hungarian Opening}}
\end{minipage}%
\caption{Chosen opening positions between \alg{CrazyAraFish 0.6.0} on GPU and \alg{Stockfish 10}}
\label{fig:openings_gpu}
\end{figure}
\vspace{-0.37cm}
\fi
\begin{table}[]
\centering
\caption{Match results of \alg{CrazyAra 0.6.0} on CPU playing twelve different crazyhouse engines. \textbf{Wapc} and \textbf{Lapc} means ``win average ply count'' and ``loss average ply count'' and describe the average game length. Engine denoted with \textdagger $\,$ use Wine for emulation.} 
\label{tab:strength_eval_cpu}
\vspace{.2cm}
\begin{tabular}{lcccccccc}
\toprule
\textbf{Engine Name} & \textbf{Version}         & \textbf{Elo Rating}             & \textbf{NPS (Million)}       & \textbf{Wapc}     & \textbf{Lapc}    & \textbf{+}  & \textbf{=} & \textbf{-}   \\
\midrule %
CrazyAra    & 0.6.0           & -                            & \num{0.00033} & $62\pm18$  & $74\pm24$  & \num{118} & \num{0} & \num{12}  \\ 
PyChess     & 1.1             & \textgreater{}\,\num{1566.25}* & \num{0.012}   &  -         & $41\pm11$  & \num{0}   & \num{0} & \num{10}  \\ 
KKFChess \textdagger& 2.6.7b  & \num{1849.50}                & \num{2.9}     & -          & $57\pm10$  & \num{0}   & \num{0} & \num{10}  \\ 
TSCP ZH     & 1.1             & \num{1888.47}                & \num{0.5}     & -          & $54\pm12$  & \num{0}   & \num{0} & \num{10}  \\ 
Pulsar      & 2009-9b         & \num{1982.07}                & ?             & -          & $61\pm13$  & \num{0}   & \num{0} & \num{10}  \\ 
Feuerstein \textdagger & 0.4.6.1 & \num{2205.74}             & \num{0.1}     & -          & $57\pm8$   & \num{0}   & \num{0} & \num{10}  \\ 
Nebiyu \textdagger & 1.45a    & \num{2244.39}                & \num{1.5}     & -          & $42\pm8$   & \num{0}   & \num{0} & \num{10}  \\ 
SjaakII     & 1.4.1           & \num{2245.56}                & \num{0.425}   & -          & $61\pm10$  & \num{0}   & \num{0} & \num{10}  \\ 
Sjeng       & 11.2            & \num{2300.00}                & \num{0.7}     & -          & $66\pm15$  & \num{0}   & \num{0} & \num{10}  \\ 
CrazyWa     & 1.0.1           & \num{2500.00}                & \num{1.4}     & -          & $73\pm10$  & \num{0}   & \num{0} & \num{10}  \\ 
Sunsetter   & 9               & \num{2703.39}                & \num{1.5}     & -          & $74\pm19$  & \num{0}   & \num{0} & \num{10}  \\ 
TjChess     & 1.37            & \num{2732.58}                & \num{1.37}    & $53\pm0$   & $85\pm17$  & \num{1}   & \num{0} & \num{9}   \\ 
Imortal \textdagger & 4.3     & \textgreater{}\,\num{2997.33}* & 0.9           & $134\pm0$  & $77\pm9$   & \num{1}   & \num{0} & \num{9}   \\ 
Stockfish   & 10 (2018-11-29) & \textgreater{}\,\num{3946.06}* & 4.6           & $70\pm16$  & -          & \num{10}  & \num{0} & \num{0}  \\
\bottomrule
\end{tabular}
\end{table}

However, \CrazyAra lost all games to \alg{Stockfish}. Despite this, it was able to generate $+2.62$, $+6.13$, $+6.44$ centipawn positions in three separate games as white according to the evaluation of \alg{Stockfish}.
To reduce the effect of the opening advantage for the first player in crazyhouse and also the fact that \CrazyAra can make use of an implicitly built opening-book, we choose five opening positions for the final evaluation, which are more balanced and less popular in human
\ifarxiv
play (see Figure \ref{fig:openings_gpu}).
\fi


The games between \alg{CrazyAraFish 0.6.0} and \alg{Stockfish 10} were generated on an AMD® Ryzen 7, 1700 eight-core processor × 16 for both engines and also a GTX1080ti for \alg{CrazyAraFish 0.6.0}.
\alg{Stockfish} achieves \num{6.7} million NPS on our setup.
The hash size for \alg{Stockfish} was set to 1024\,mb.\cfootnote{We also tried choosing a higher hash size for \alg{Stockfish} e.\,g., 4096\,mb, but found it to be unstable resulting in game crashes for the \alg{Stockfish} executable \textit{x86\_64-modern 2018-11-29}. These crashes have been reported to the corresponding developers.}
We used a batch-size of eight and two threads for traversing the search tree in the case of \alg{CrazyAraFish}.
In this setting, \alg{CrazyAraFish 0.6.0} achieved a NPS of \num{1400} resulting in a node ratio of about \num{1}:\num{4700}. In positions where many transpositions and terminal nodes were visited, the NPS increased to \num{4000}. 
The matches were played in a long time control format of 30\,min + 30\,s. Here,
\alg{CrazyAraFish 0.6.0} won three games and drew one game out of ten games (see Table~\ref{tab:strength_eval_cafish060}).
\begin{table}[]
\centering
\caption{Match results of \alg{CrazyAraFish 0.6.0} playing \alg{Stockfish 10} in a time control of 30\,min + 30\,s} 
\label{tab:strength_eval_cafish060}
\vspace{.2cm}
\begin{tabular}{lcccccccc}
\toprule
\textbf{Engine Name} & \textbf{Version} & \textbf{Elo Rating} & \textbf{NPS (Million)} & \textbf{Wapc} & \textbf{Lapc} & \textbf{+} & \textbf{=} & \textbf{-} \\
\midrule
CrazyAraFish & 0.6.0 & - & 0.0014 & $118\pm22$ & $98\pm34$ & 3 & 1 & 6 \\
Stockfish & 10 (2018-11-29) & \textgreater{}\,3,946.06 & 6.7 & $98\pm34$ & $118\pm22$ & 6 & 1 & 3 \\
\bottomrule
\end{tabular}
\end{table}
\ifarxiv
In the following Figures~\ref{fig:cafish_sf_game4},~\ref{fig:cafish_sf_game8},~\ref{fig:cafish_sf_game9} and \ref{fig:cafish_sf_game10} the four most interesting games of \alg{CrazyAraFish} are shown. All other games can be found in the supplementary materials.
\newpage
\begin{figure}
\begin{minipage}{\textwidth}
\centering
\subcaptionbox{Evaluation progression for both engines}{
\includegraphics[width=\textwidth]{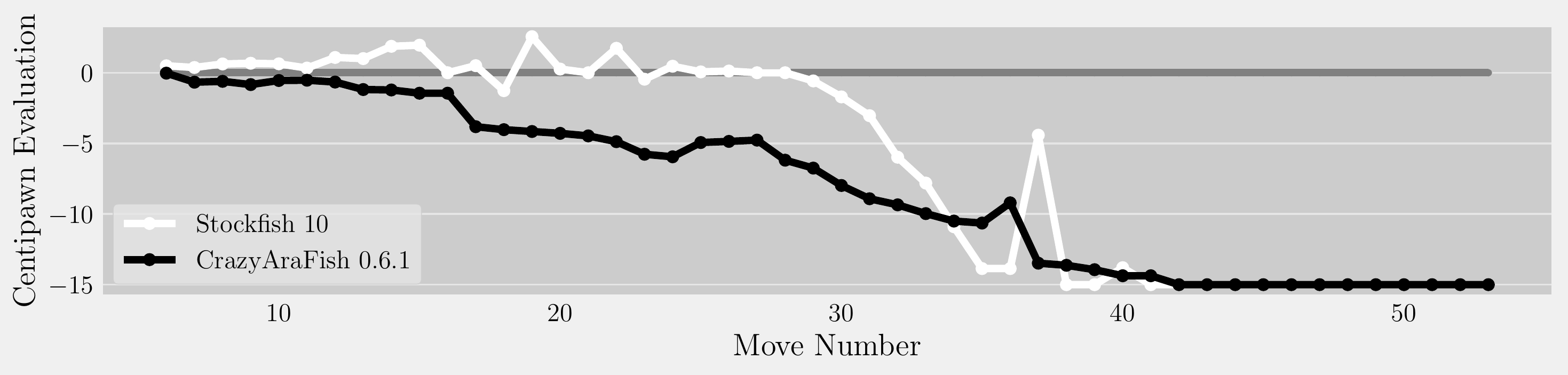}
}
\label{fig:eval_game1}
\end{minipage}
\begin{minipage}{\textwidth}
\vspace{\floatsep}
\crule
\begin{csmall}
[Event "CrazyAraFish-0.6.0-vs-SF10"]\newline
[Site "Darmstadt, GER"]\newline
[Date "2019.07.26"]\newline
[Round "2"]\newline
[White "stockfish-x86\_64-modern  2018-11-29"]\newline
[Black "CrazyAraFish-0.6.0"]\newline
[Result "0-1"]\newline
[PlyCount "108"]\newline
[TimeControl "1800+30"]\newline
[Variant "crazyhouse"]\newline


1. e4 \{book\} e5 \{book\} 2. Nf3 \{book\} Nc6 \{book\} 3. Bc4 \{book\} Bc5 \{book\}
4. O-O \{book\} Nf6 \{book\} 5. d3 \{book\} O-O \{book\} 6. Bg5 \{book\} Be7 \{book\}
7. Be3 \{+0.49/28 118s\} d6 \{+0.03/35 62s\} 8. a3 \{+0.37/31 164s\}
Kh8 \{+0.67/38 62s\} 9. Kh1 \{+0.62/27 23s\} Bg4 \{+0.61/37 62s\}
10. Nbd2 \{+0.67/29 54s\} Nd4 \{+0.84/49 63s\} 11. Rg1 \{+0.63/30 230s\}
c6 \{+0.55/31 63s\} 12. c3 \{+0.33/31 88s\} Ne6 \{+0.53/35 63s\} 13. h3 \{+1.08/25 16s\}
Bxf3 \{+0.66/38 63s\} 14. Nxf3 \{+1.00/28 102s\} N@g4 \{+1.19/39 64s\}
15. hxg4 \{+1.87/25 27s\} Nxg4 \{+1.22/33 64s\} 16. B@f5 \{+1.95/30 349s\}
Nxe3 \{+1.45/31 64s\} 17. fxe3 \{0.00/30 100s\} d5 \{+1.45/53 64s\}
18. Bb3 \{+0.50/25 123s\} Ng5 \{+3.82/33 65s\} 19. exd5 \{-1.27/24 51s\}
Nxf3 \{+4.03/39 65s\} 20. Qxf3 \{+2.55/25 25s\} N@h4 \{+4.16/37 65s\}
21. Qg4 \{+0.26/29 343s\} Nxf5 \{+4.29/35 66s\} 22. Qxf5 \{0.00/25 28s\}
cxd5 \{+4.47/34 66s\} 23. P@h6 \{+1.73/25 72s\} gxh6 \{+4.88/38 66s\}
24. Qxe5+ \{-0.46/27 236s\} B@f6 \{+5.77/33 67s\} 25. Qh5 \{+0.46/25 23s\}
P@g7 \{+5.95/37 67s\} 26. P@f2 \{+0.06/27 78s\} B@e6 \{+4.94/35 67s\}
27. N@f4 \{+0.12/29 47s\} Bd6 \{+4.86/38 67s\} 28. Qe2 \{0.00/28 39s\}
Bxf4 \{+4.77/30 68s\} 29. exf4 \{0.00/25 13s\} P@h3 \{+6.19/34 68s\}
30. gxh3 \{-0.58/25 50s\} Bxh3 \{+6.75/31 69s\} 31. P@g2 \{-1.71/28 97s\}
Bxg2+ \{+7.98/33 69s\} 32. Rxg2 \{-3.04/21 23s\} P@h3 \{+8.92/36 69s\}
33. Rg3 \{-5.98/23 58s\} Re8 \{+9.35/36 70s\} 34. B@e6 \{-7.81/22 32s\}
Rxe6 \{+9.97/34 71s\} 35. Qxe6 \{-10.90/21 30s\} B@g2+ \{+10.50/27 71s\}
36. Kh2 \{-13.86/21 30s\} N@f3+ \{+10.64/31 72s\} 37. Rxf3 \{-13.86/1 0s\}
Bxf3 \{+9.21/43 72s\} 38. N@e3 \{-4.42/22 27s\} R@g2+ \{+13.48/28 73s\}
39. Nxg2 \{-20.48/24 60s\} hxg2 \{+13.63/23 74s\} 40. B@e2 \{-21.62/23 33s\}
Bxe2 \{+13.94/23 75s\} 41. Qxe2 \{-13.80/19 7.6s\} B@f1 \{+14.37/22 48s\}
42. Qf3 \{-18.02/21 39s\} Qd7 \{+14.37/21 48s\} 43. f5 \{-32.92/18 43s\}
N@e5 \{+17.28/13 47s\} 44. Qg3 \{-30.51/17 30s\} P@h4 \{+20.17/9 46s\}
45. Qxg2 \{-M16/28 30s\} Bxg2 \{+22.24/18 45s\} 46. Kxg2 \{-M14/30 9.6s\}
Qxf5 \{+28.03/11 44s\} 47. R@b8+ \{-M12/30 15s\} Rxb8 \{+35.36/11 22s\}
48. R@g8+ \{-M10/29 16s\} Rxg8 \{+35.81/17 44s\} 49. N@g6+ \{-M8/46 21s\}
hxg6 \{+34.93/15 43s\} 50. N@g1 \{-M6/51 16s\} P@f3+ \{+37.14/7 42s\}
51. Kf1 \{-M6/85 17s\} P@g2+ \{+37.45/5 42s\} 52. Ke1 \{-M4/1 0.001s\}
Qxd3 \{+55.96/3 41s\} 53. N@d2 \{-M4/67 13s\} R@f1+ \{+85.21/3 41s\}
54. Nxf1 \{-M2/1 0.001s\} gxf1=Q\# \{+M1/1 60s, Black mates\} 0-1
\end{csmall}
\crule
\end{minipage}
\caption{Game 4 / 10}
\label{fig:cafish_sf_game4}
\end{figure}
\addtocounter{subfigure}{-1}
\newpage
\begin{figure}
\begin{minipage}{\textwidth}
\subcaptionbox{Evaluation progression for both engines}{ 
\includegraphics[width=\textwidth]{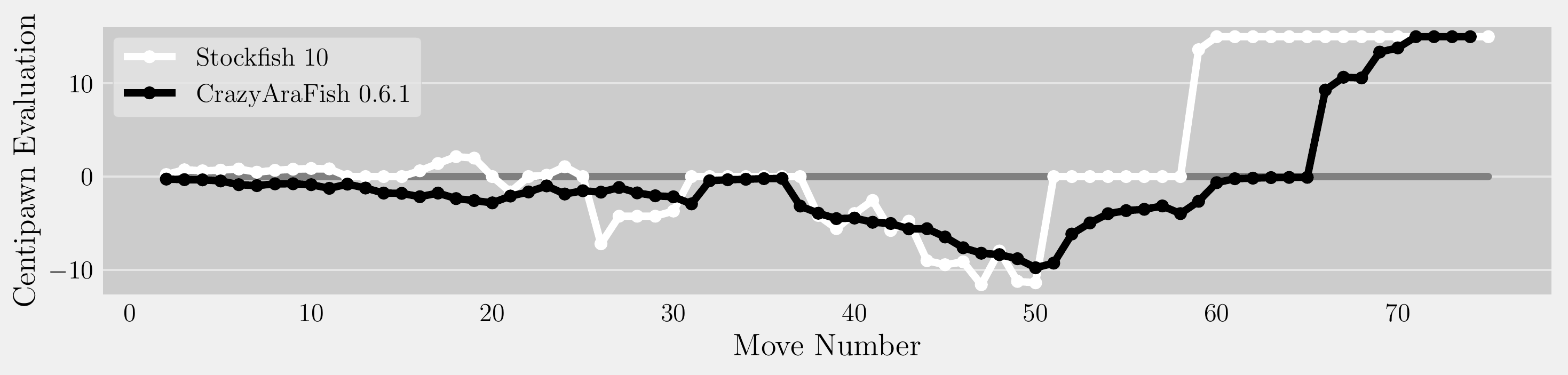}
}
\label{fig:eval_game1}
\end{minipage}
\begin{minipage}{\textwidth}
\vspace{\floatsep}
\crule
\begin{csmall}
[Event "CrazyAraFish-0.6.0-vs-SF10"]\newline
[Site "Darmstadt, GER"]\newline
[Date "2019.07.27"]\newline
[Round "4"]\newline
[White "stockfish-x86\_64-modern  2018-11-29"]\newline
[Black "CrazyAraFish-0.6.0"]\newline
[Result "1-0"]\newline
[PlyCount "151"]\newline
[TimeControl "1800+30"]\newline
[Variant "crazyhouse"]\newline
\\
1. e3 \{book\} d5 \{book\} 2. d4 \{book\} Bf5 \{book\} 3. Bd3 \{+0.22/31 173s\}
e6 \{+0.27/24 58s\} 4. Bxf5 \{+0.73/27 31s\} exf5 \{+0.33/23 59s\}
5. Ne2 \{+0.62/30 276s\} Nf6 \{+0.35/25 59s\} 6. Nbc3 \{+0.69/26 25s\}
Nc6 \{+0.47/23 59s\} 7. B@h4 \{+0.81/29 190s\} Be7 \{+0.87/29 60s\}
8. O-O \{+0.47/28 169s\} O-O \{+0.97/41 60s\} 9. Bd2 \{+0.67/28 40s\}
B@d6 \{+0.78/32 60s\} 10. Ng3 \{+0.79/25 19s\} Ne4 \{+0.77/28 60s\}
11. Bxe7 \{+0.87/27 45s\} Nxe7 \{+0.87/32 60s\} 12. B@e5 \{+0.83/29 157s\}
B@h6 \{+1.26/26 61s\} 13. Ncxe4 \{0.00/29 112s\} dxe4 \{+0.81/45 61s\}
14. Qh5 \{0.00/29 52s\} Bxe5 \{+1.23/47 61s\} 15. dxe5 \{0.00/29 44s\}
B@f3 \{+1.77/49 61s\} 16. gxf3 \{0.00/29 24s\} exf3 \{+1.80/47 61s\}
17. Bc3 \{+0.62/24 203s\} Nd5 \{+2.16/40 62s\} 18. B@h3 \{+1.40/23 28s\}
Nxc3 \{+1.77/40 62s\} 19. bxc3 \{+2.14/23 14s\} N@e2+ \{+2.36/38 62s\}
20. Nxe2 \{+1.98/25 59s\} fxe2 \{+2.57/38 62s\} 21. Qxe2 \{0.00/29 65s\}
P@g4 \{+2.82/35 63s\} 22. B@f4 \{-1.65/28 323s\} Bxf4 \{+2.09/34 64s\}
23. exf4 \{0.00/22 15s\} N@f3+ \{+1.65/42 63s\} 24. Qxf3 \{+0.15/23 19s\}
gxf3 \{+1.00/34 64s\} 25. B@h1 \{+1.06/26 53s\} B@g4 \{+1.87/45 64s\}
26. Bxg4 \{0.00/26 156s\} fxg4 \{+1.52/44 64s\} 27. B@f6 \{-7.21/22 117s\}
B@g3 \{+1.67/65 65s\} 28. hxg3 \{-4.25/25 21s\} B@h2+ \{+1.17/63 65s\}
29. Kxh2 \{-4.25/1 0.001s\} Q@h3+ \{+1.75/61 65s\} 30. Kg1 \{-4.25/1 0s\}
Qxf6 \{+2.06/55 66s\} 31. N@h2 \{-3.70/25 14s\} Qxf4 \{+2.17/52 66s\}
32. N@e7+ \{0.00/26 43s\} Kh8 \{+2.93/21 1.0s\} 33. P@f6 \{0.00/26 37s\}
B@h6 \{+0.45/44 106s\} 34. fxg7+ \{0.00/27 49s\} Bxg7 \{+0.34/42 35s\}
35. P@f6 \{0.00/29 18s\} Qxe5 \{+0.29/40 72s\} 36. B@d4 \{0.00/29 23s\}
Qxf6 \{+0.22/38 73s\} 37. Bxf6 \{0.00/30 97s\} Bxf6 \{+0.20/36 73s\}
38. B@h4 \{0.00/30 45s\} Bxe7 \{+3.16/42 75s\} 39. Bxf3 \{-4.17/23 34s\}
B@g7 \{+3.93/32 75s\} 40. P@f6 \{-5.58/26 98s\} Bexf6 \{+4.51/40 76s\}
41. Bxf6 \{-3.96/24 22s\} Bxf6 \{+4.45/38 49s\} 42. Q@f4 \{-2.56/23 47s\}
B@e5 \{+4.90/36 48s\} 43. Qxg4 \{-5.78/22 29s\} Qxg4 \{+5.03/31 47s\}
44. Nxg4 \{-4.78/21 26s\} P@h2+ \{+5.61/28 46s\} 45. Nxh2 \{-9.02/19 49s\}
P@e2 \{+5.59/32 46s\} 46. Q@g2 \{-9.45/20 30s\} P@h3 \{+6.48/26 45s\}
47. Qxh3 \{-9.15/19 8.5s\} N@g5 \{+7.63/27 44s\} 48. Qg4 \{-11.59/22 52s\}
Nxf3+ \{+8.22/27 43s\} 49. Qxf3 \{-7.98/20 7.0s\} exf1=R+ \{+8.37/30 42s\}
50. Rxf1 \{-11.24/25 53s\} P@h3 \{+8.81/23 42s\} 51. B@h1 \{-11.39/23 30s\}
B@c6 \{+9.78/25 41s\} 52. Qxf6+ \{0.00/25 12s\} Bxf6 \{+9.28/22 41s\}
53. B@g5 \{0.00/28 32s\} P@g7 \{+6.16/39 60s\} 54. Bxf6 \{0.00/28 19s\}
Q@g6 \{+4.97/33 39s\} 55. Bxg7+ \{0.00/29 20s\} Qxg7 \{+3.98/31 38s\}
56. P@f6 \{0.00/30 36s\} Qxf6 \{+3.65/36 38s\} 57. B@e7 \{0.00/29 43s\}
Q@g5 \{+3.51/39 37s\} 58. Bxf6+ \{0.00/30 25s\} Qxf6 \{+3.15/37 37s\}
59. Q@f5 \{0.00/30 24s\} B@e5 \{+3.97/25 37s\} 60. P@h6 \{+13.61/27 58s\}
P@g7 \{+2.65/38 54s\} 61. hxg7+ \{+16.60/22 15s\} Qxg7 \{+0.65/19 18s\}
62. P@h6 \{+19.05/24 41s\} Qxh6 \{+0.23/17 36s\} 63. Qxe5+ \{+20.50/21 10s\}
P@g7 \{+0.17/15 35s\} 64. P@f6 \{+28.84/22 23s\} gxf6 \{+0.11/13 35s\}
65. B@g5 \{+37.64/21 17s\} B@g7 \{+0.09/11 35s\} 66. Bxh6 \{+47.40/23 13s\}
Bxh6 \{+0.08/9 35s\} 67. N@g5 \{+M25/20 20s\} R@g6 \{-9.28/28 52s\}
68. P@g7+ \{+M19/28 14s\} Rxg7 \{-10.65/20 33s\} 69. Qxf6 \{+M15/31 17s\}
B@g6 \{-10.58/18 33s\} 70. N@e7 \{+M13/38 20s\} Bxg5 \{-13.35/12 33s\}
71. Qxg7+ \{+M11/48 22s\} Kxg7 \{-13.80/10 0.037s\} 72. N@h5+ \{+M9/50 58s\}
Bxh5 \{-19.59/8 34s\} 73. Nf5+ \{+M7/54 30s\} Kf6 \{-23.11/6 34s\}
74. Q@e7+ \{+M5/59 18s\} Kg6 \{-30.89/4 34s\} 75. R@g7+ \{+M3/65 25s\}
Kxf5 \{-M1/2 0.022s\} 76. Rxg5\# \{+M1/127 0.017s, White mates\} 1-0
\end{csmall}
\crule
\end{minipage}
\caption{Game 8 / 10}
\label{fig:cafish_sf_game8}
\end{figure}
\addtocounter{subfigure}{-1}
\newpage
\begin{figure}
\begin{minipage}{\textwidth}
\centering
\subcaptionbox{Evaluation progression for both engines\label{fig:eval_game9}}{
\includegraphics[width=\textwidth]{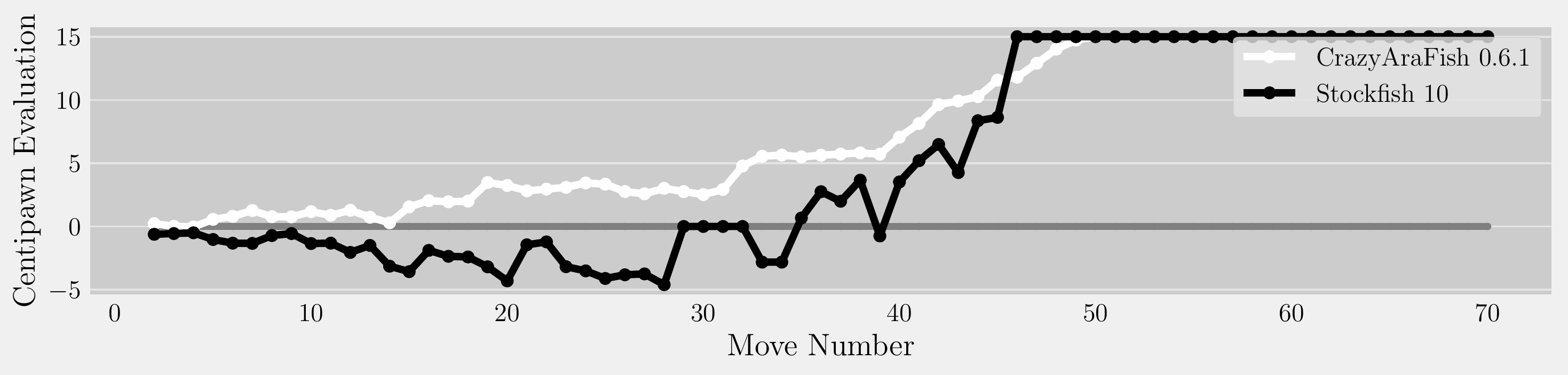}
}
\end{minipage}
\begin{minipage}{\textwidth}
\vspace{\floatsep}
\crule
\begin{csmall}
[Event "CrazyAraFish-0.6.0-vs-SF10"]\newline
[Site "Darmstadt, GER"]\newline
[Date "2019.07.27"]\newline
[Round "5"]\newline
[White "CrazyAraFish-0.6.0"]\newline
[Black "stockfish-x86\_64-modern  2018-11-29"]\newline
[Result "1-0"]\newline
[PlyCount "143"]\newline
[TimeControl "1800+30"]\newline
[Variant "crazyhouse"]\newline


1. g3 \{book\} d5 \{book\} 2. d4 \{book\} Nf6 \{book\} 3. Bg2 \{book\} Bf5 \{+0.62/30 139s\}
4. Nf3 \{+0.22/38 59s\} e6 \{+0.56/28 47s\} 5. Nc3 \{+0.01/37 60s\} Be7 \{+0.50/29 33s\}
6. Nh4 \{-0.04/29 60s\} Be4 \{+1.03/28 58s\} 7. Nxe4 \{+0.56/23 60s\}
Nxe4 \{+1.32/25 18s\} 8. O-O \{+0.80/21 60s\} O-O \{+1.34/26 57s\}
9. B@e5 \{+1.26/32 61s\} f6 \{+0.72/29 254s\} 10. Bef4 \{+0.77/47 61s\}
f5 \{+0.56/27 43s\} 11. f3 \{+0.76/35 61s\} Nd6 \{+1.35/29 58s\}
12. Bxd6 \{+1.17/27 61s\} Qxd6 \{+1.32/28 32s\} 13. N@f4 \{+0.88/43 61s\}
Nc6 \{+2.05/27 63s\} 14. Kh1 \{+1.28/32 62s\} B@e8 \{+1.50/29 239s\}
15. Be3 \{+0.73/41 62s\} N@c4 \{+3.14/24 24s\} 16. Qc1 \{+0.29/42 62s\}
Nxe3 \{+3.57/27 63s\} 17. Qxe3 \{+1.55/48 62s\} Nxd4 \{+1.89/28 89s\}
18. Qxd4 \{+2.04/34 63s\} B@b6 \{+2.36/28 73s\} 19. Qxg7+ \{+1.95/41 63s\}
Kxg7 \{+2.41/1 0.001s\} 20. N@h5+ \{+2.00/39 63s\} Bxh5 \{+3.19/24 24s\}
21. Nxh5+ \{+3.47/50 64s\} Kf7 \{+4.30/26 38s\} 22. P@f2 \{+3.24/52 64s\}
Q@g5 \{+1.45/29 353s\} 23. N@g7 \{+2.82/68 64s\} P@e5 \{+1.22/22 51s\}
24. B@c1 \{+2.95/39 64s\} Qxg7 \{+3.18/28 53s\} 25. Nxg7 \{+3.09/40 65s\}
Kxg7 \{+3.51/26 15s\} 26. Q@h6+ \{+3.44/52 65s\} Kg8 \{+4.11/28 31s\}
27. Ng6 \{+3.34/45 65s\} N@d7 \{+3.82/32 145s\} 28. Nxe7+ \{+2.76/61 66s\}
Qxe7 \{+3.75/30 27s\} 29. Bg5 \{+2.58/62 66s\} N@f6 \{+4.60/26 40s\}
30. B@g6 \{+3.01/39 67s\} hxg6 \{0.00/32 91s\} 31. Qxg6+ \{+2.76/59 67s\}
B@g7 \{0.00/32 24s\} 32. P@h6 \{+2.52/63 67s\} Rf7 \{0.00/33 38s\}
33. c4 \{+2.92/53 68s\} d4 \{0.00/29 228s\} 34. Qxg7+ \{+4.78/39 70s\}
Rxg7 \{+2.82/1 0s\} 35. hxg7 \{+5.55/45 69s\} Kxg7 \{+2.82/23 42s\}
36. B@h6+ \{+5.64/54 70s\} Kh8 \{-0.67/26 40s\} 37. Bxf6+ \{+5.50/51 70s\}
Qxf6 \{-2.75/28 80s\} 38. Bg5 \{+5.63/50 71s\} Qg7 \{-1.99/26 35s\}
39. R@e7 \{+5.72/49 72s\} P@f7 \{-3.66/29 161s\} 40. N@h6 \{+5.82/47 73s\}
N@d6 \{+0.75/26 61s\} 41. Rxd7 \{+5.71/40 47s\} Q@e8 \{-3.52/27 67s\}
42. Rxd6 \{+7.06/36 48s\} cxd6 \{-5.20/25 36s\} 43. N@f6 \{+8.14/35 46s\}
Qef8 \{-6.48/24 30s\} 44. N@d7 \{+9.63/40 45s\} Qd8 \{-4.26/26 13s\}
45. Nxb6 \{+9.92/43 44s\} axb6 \{-8.36/28 48s\} 46. B@h4 \{+10.26/29 43s\}
B@e7 \{-8.62/25 30s\} 47. Nh5 \{+11.57/29 43s\} Bxg5 \{-16.58/25 30s\}
48. Nxg7 \{+11.80/29 42s\} B@g6 \{-17.85/26 30s\} 49. Nxe6 \{+12.90/37 42s\}
fxe6 \{-17.90/24 14s\} 50. Q@f7 \{+14.02/29 41s\} R@g7 \{-21.50/26 39s\}
51. Qxg6 \{+14.73/24 40s\} Rxg6 \{-25.64/21 36s\} 52. Nf7+ \{+15.14/21 40s\}
Kh7 \{-26.56/21 30s\} 53. Nxd8 \{+15.01/13 40s\} Rxd8 \{-30.14/18 30s\}
54. Bxg5 \{+16.85/15 39s\} Rxg5 \{-24.47/19 10s\} 55. B@f6 \{+17.60/18 38s\}
Q@g6 \{-33.20/16 50s\} 56. Bxd8 \{+18.55/18 38s\} N@h6 \{-32.17/18 26s\}
57. B@f6 \{+20.36/19 38s\} B@g7 \{-37.25/16 34s\} 58. Bxg5 \{+22.83/22 37s\}
N@h5 \{-38.50/18 30s\} 59. Bxh6 \{+23.79/12 37s\} Bxh6 \{-44.75/16 30s\}
60. N@e7 \{+24.61/12 36s\} B@f7 \{-M16/31 30s\} 61. Nxg6 \{+25.65/15 36s\}
Bxg6 \{-M14/34 11s\} 62. Q@e7+ \{+24.38/11 36s\} N@f7 \{-M12/38 15s\}
63. Q@e8 \{+27.64/13 36s\} N@h8 \{-M12/38 21s\} 64. R@g8 \{+29.68/11 35s\}
N@f8 \{-M10/49 16s\} 65. Rxf8 \{+31.01/9 35s\} Bxf8 \{-M10/54 15s\}
66. Q7xf8 \{+37.97/7 35s\} Nxg3+ \{-M10/61 15s\} 67. fxg3 \{+42.21/5 34s\}
R@g1+ \{-M8/71 18s\} 68. Rxg1 \{+43.94/3 34s\} P@g7 \{-M6/97 19s\}
69. Qg8+ \{+34.15/7 34s\} Kh6 \{-M6/1 0.001s\} 70. R@h4+ \{+42.23/5 34s\}
Bh5 \{-M4/1 0.001s\} 71. R@h7+ \{+81.41/3 34s\} Kg6 \{-M2/1 0.002s\}
72. Qxg7\# \{+M1/1 50s, White mates\} 1-0
\end{csmall}
\noindent\makebox[\linewidth]{\rule{\textwidth}{0.4pt}}
\end{minipage}
\caption{Game 9 / 10}
\label{fig:cafish_sf_game9}
\end{figure}
\addtocounter{subfigure}{-1}
\newpage
\begin{figure}
\begin{minipage}{\textwidth}
\begin{center}
\subcaptionbox{Evaluation progression for both engines}{
\includegraphics[width=\textwidth]{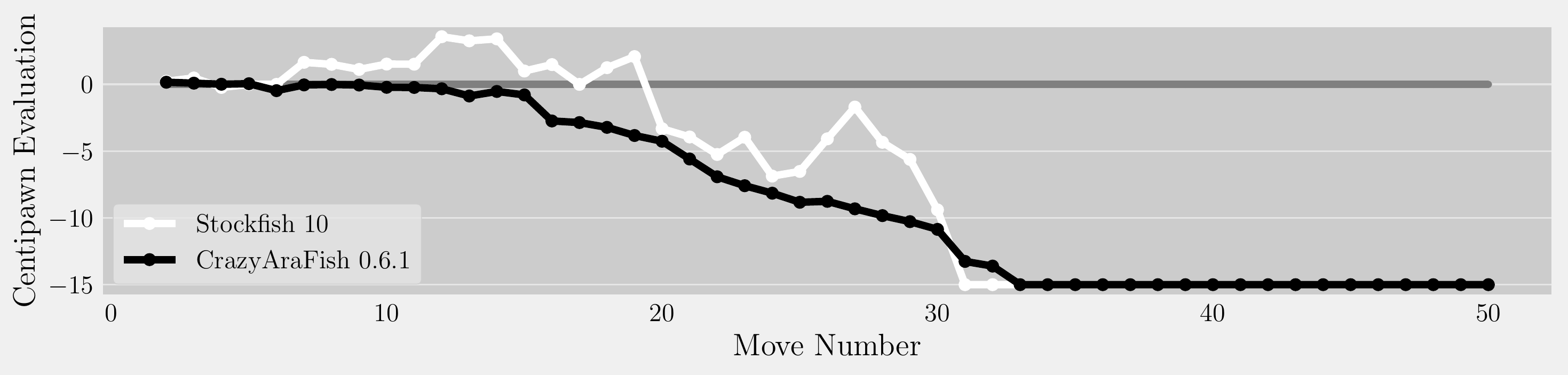}
}
\end{center}
\label{fig:eval_game1}
\end{minipage}
\begin{minipage}{\textwidth}
\vspace{\floatsep}
\crule
\begin{csmall}
[Event "CrazyAraFish-0.6.0-vs-SF10"]\newline
[Site "Darmstadt, GER"]\newline
[Date "2019.07.27"]\newline
[Round "5"]\newline
[White "stockfish-x86\_64-modern  2018-11-29"]\newline
[Black "CrazyAraFish-0.6.0"]\newline
[Result "0-1"]\newline
[PlyCount "102"]\newline
[TimeControl "1800+30"]\newline
[Variant "crazyhouse"]\newline

1. g3 \{book\} d5 \{book\} 2. d4 \{book\} Nf6 \{book\} 3. Bg2 \{book\} Bf5 \{-0.16/37 58s\}
4. Nc3 \{+0.23/30 174s\} e6 \{-0.09/26 59s\} 5. Nf3 \{+0.48/27 33s\}
Nc6 \{-0.01/26 59s\} 6. O-O \{-0.23/28 120s\} Be7 \{-0.05/28 59s\}
7. Ne5 \{0.00/31 79s\} O-O \{+0.47/32 60s\} 8. h3 \{0.00/33 409s\} h6 \{+0.04/36 60s\}
9. Nxc6 \{+1.65/27 55s\} bxc6 \{+0.01/43 60s\} 10. N@e5 \{+1.50/26 41s\}
Qe8 \{+0.05/42 60s\} 11. Bf4 \{+1.12/31 129s\} Kh8 \{+0.22/48 60s\}
12. Qd2 \{+1.52/25 25s\} Kh7 \{+0.23/50 61s\} 13. Qc1 \{+1.51/24 111s\}
N@g6 \{+0.33/38 61s\} 14. Nxg6 \{+3.57/26 21s\} fxg6 \{+0.87/35 61s\}
15. Bxc7 \{+3.26/27 69s\} N@g5 \{+0.53/50 61s\} 16. N@f4 \{+3.41/26 101s\}
c5 \{+0.79/41 61s\} 17. Be5 \{+1.00/27 337s\} cxd4 \{+2.74/40 62s\}
18. Bxd4 \{+1.48/25 11s\} P@g4 \{+2.86/50 62s\} 19. hxg4 \{0.00/25 137s\}
Nxg4 \{+3.22/45 62s\} 20. P@f3 \{+1.25/26 31s\} e5 \{+3.83/43 62s\}
21. fxg4 \{+2.08/21 10s\} Bxg4 \{+4.26/48 63s\} 22. Bxe5 \{-3.32/24 258s\}
P@f3 \{+5.59/30 63s\} 23. exf3 \{-3.94/23 128s\} Bxf3 \{+6.92/19 63s\}
24. P@f7 \{-5.25/24 97s\} Qxf7 \{+7.59/35 64s\} 25. P@h5 \{-3.96/20 17s\}
Bxh5 \{+8.15/38 64s\} 26. f3 \{-6.87/23 62s\} P@h3 \{+8.83/29 65s\}
27. Nxh3 \{-6.52/20 12s\} Nxh3+ \{+8.76/31 65s\} 28. Bxh3 \{-4.08/20 18s\}
N@g5 \{+9.32/33 65s\} 29. P@g2 \{-1.70/19 10s\} Nxh3+ \{+9.83/35 66s\}
30. gxh3 \{-4.34/21 37s\} P@g2 \{+10.28/31 66s\} 31. Rf2 \{-5.62/22 68s\}
B@d6 \{+10.86/31 67s\} 32. N@f6+ \{-9.40/18 39s\} Bxf6 \{+13.26/15 68s\}
33. Bxd6 \{-15.35/17 30s\} Bxc3 \{+13.61/21 67s\} 34. Bf4 \{-17.50/17 30s\}
P@d2 \{+16.76/21 68s\} 35. Qd1 \{-22.15/18 30s\} Bxb2 \{+17.33/21 69s\}
36. Qxd2 \{-24.37/16 30s\} N@c4 \{+19.66/18 69s\} 37. Qe2 \{-24.40/18 30s\}
N@d4 \{+20.09/19 70s\} 38. Qd3 \{-31.53/17 30s\} Bxa1 \{+19.74/20 70s\}
39. Rxg2 \{-34.87/17 30s\} Nxf3+ \{+21.68/21 71s\} 40. Kh1 \{-53.50/18 27s\}
R@e1+ \{+23.10/15 72s\} 41. B@f1 \{-44.17/18 26s\} P@e4 \{+24.32/16 47s\}
42. N@h2 \{-47.50/18 37s\} exd3 \{+26.75/17 46s\} 43. P@e7 \{-M18/21 30s\}
Nxh2 \{+29.91/23 45s\} 44. exf8=N+ \{-M16/26 18s\} Rxf8 \{+30.27/13 45s\}
45. R@h8+ \{-M14/29 22s\} Kxh8 \{+31.60/9 44s\} 46. P@e7 \{-M12/37 21s\}
Rxf1+ \{+29.56/11 43s\} 47. N@g1 \{-M10/53 13s\} Rxg1+ \{+29.11/9 42s\}
48. Rxg1 \{-M8/57 18s\} N@f2+ \{+31.25/7 42s\} 49. Kg2 \{-M6/67 18s\}
Bf3+ \{+49.58/3 41s\} 50. Kxh2 \{-M4/75 26s\} N@g4+ \{+83.63/3 41s\}
51. hxg4 \{-M2/1 0.001s\} R@h3\# \{+M1/1 60s, Black mates\} 0-1
\end{csmall}
\crule
\end{minipage}
\caption{Game 10 / 10}
\label{fig:cafish_sf_game10}
\end{figure}
\addtocounter{subfigure}{-1}
\fi
\FloatBarrier

\section{Conclusion}
\label{sec:conclusion}
In this work we have developed a crazyhouse chess program, called \alg{CrazyAra}, based on a combination of deep neural networks and tree search, that plays at the level of the strongest human players. Despite the highly tactical game-style of crazyhouse and a relatively small and low quality data set,
it is possible to achieve a remarkable performance when only using supervised learning. We demonstrated that MCTS is fairly successful at low samples, when powered by a (deep) neural network, and is able to drastically increase the quality of moves over time for the crazyhouse variant. Most importantly,
we demonstrated that the scheme proposed by \cite{silver2017mastering} can be and has to be improved and accelerated in different areas for crazyhouse: this includes achieving a better training performance
and the prediction of higher quality moves with the same number of MCTS simulations. 
Indeed, several optimizations are achievable in future work. 
A faster generation of rollouts e.\,g., by using low precision inference like float16 or int8 and potential future improvements in network design and MCTS will boost performance.
Additionally, applying reinforcement learning can help to increase the playing strength of \alg{CrazyAra} further. 
\paragraph*{Acknowledgments}
The authors thank the main \textit{Stockfish} developers of crazyhouse, Fabian Fichter, Daniel Dugovic, Niklas Fiekas for
valuable discussions. They also thank other crazyhouse-engine programmers including Bajusz Tamás, Harm Geert Muller and Ferdinand Mosca for providing their latest chess engine executable.
The authors are grateful to the lichess.org crazyhouse community \textit{LM~JannLee} (Justin Tan), \textit{IM~gsvc}, \textit{TheFinnisher}, \textit{IM~opperwezen} (IM~Vincent Rothuis), \textit{FM WinnerOleg} (FM~Oleg Papayan) and \textit{okei} among others for playing \textit{CrazyAra} on lichess.org and providing valuable feedback.
They thank github users \textit{@noelben} for the help in creating a time-management system and \textit{@Matuiss2} for improving the coding style and frequently testing the engine.
They appreciate the constructive feedback from Karl Stelzner when writing the paper. 
In particular, the authors are thankful to the users \textit{crazyhorse}, \textit{ObiWanBenoni}, \textit{Pichau} and \textit{varvarakh} for helping in generating the \textit{Stockfish} self-play data set.
Finally, the authors thank Thibault Duplessis and the lichess.org developer team for creating and maintaining lichess.org, providing a BOT API and the lichess.org database.

\ifarxiv
\bibliographystyle{apalike}
\fi
\bibliography{library}
\end{document}